\def\preprintversion{1}

\documentclass{article} %
\newif\ifpreprint
\ifdefined\preprintversion
  \preprinttrue
\else
  \preprintfalse
\fi
\usepackage{iclr2027_conference,times}
\usepackage[T1]{fontenc}

\usepackage{amsmath,amsfonts,bm}

\def\eqref#1{equation~\ref{#1}}

\def\1{\bm{1}}

\DeclareMathAlphabet{\mathsfit}{\encodingdefault}{\sfdefault}{m}{sl}
\SetMathAlphabet{\mathsfit}{bold}{\encodingdefault}{\sfdefault}{bx}{n}

\usepackage{hyperref}
\usepackage{url}
\usepackage{xurl}
\usepackage{iftex}
\ifPDFTeX\else
  \usepackage{newunicodechar}
  \newunicodechar{’}{\textquoteright}
\fi
\usepackage{xcolor}
\usepackage[normalem]{ulem}

\usepackage{graphicx}
\usepackage{wrapfig}
\usepackage{booktabs}
\usepackage{multirow}
\usepackage{subcaption}
\usepackage{titletoc} 
\usepackage{tabularx}
\usepackage{array}
\usepackage{needspace}
\usepackage{float}
\usepackage{placeins}
\usepackage{etoolbox}
\hypersetup{
    hidelinks,
    pdftitle={Massive Activations in Hybrid Linear Attention Large Language Models: Pre-Attention Spikes and Inter-Spike Plateaus},
    pdfauthor={Anonymous authors},
    pdfkeywords={massive activations, hybrid linear attention, pre-attention spikes, inter-spike plateaus}
}
\colorlet{revisioncolor}{black}
\newcommand{\rev}[1]{\textcolor{revisioncolor}{#1}}
\newsavebox{\revisiontablebox}
\newcommand{\revisionfit}[1]{%
  \begingroup\sbox{\revisiontablebox}{#1}%
  \ifdim\wd\revisiontablebox>\linewidth
    \resizebox{\linewidth}{!}{\usebox{\revisiontablebox}}%
  \else\usebox{\revisiontablebox}\fi\endgroup}

\makeatletter
\let\rebuttalOriginalBibitem\@lbibitem
\def\@lbibitem[#1]#2{%
  \ifstrequal{#2}{bondarenko2023quantizable}{\color{revisioncolor}}{\color{black}}%
  \rebuttalOriginalBibitem[#1]{#2}}
\makeatother

\definecolor{maOrganization}{HTML}{0072CE}
\definecolor{maEmergence}{HTML}{C65D00}
\definecolor{maMechanisms}{HTML}{8E44AD}
\definecolor{maFunction}{HTML}{008578}

\title{Massive Activations in Hybrid Linear Attention Large Language Models: Pre-Attention Spikes and Inter-Spike Plateaus}

\ifpreprint
  \iclrfinalcopy
  \makeatletter
  \patchcmd{\@maketitle}{\lhead{Under review as a conference paper}}{\lhead{Preprint}}{}%
    {\PackageError{preprint}{Could not set the preprint running header}{Check the ICLR style's title definition.}}
  \makeatother
\hypersetup{pdfauthor={Zunhai Su; Bohan Sun; Xialie Zhuang; Shuibai Zhang; He Xiao; Hui Shen; Jing Xiong; Hengyuan Zhang; Zhongzhu Zhou; Tiantian Zhang; Ngai Wong; Chuan-Wei Kuo}}
\author{%
  \begin{minipage}[t]{\dimexpr\textwidth-2\tabcolsep\relax}
  \centering
  Zunhai Su\textsuperscript{1,2}\thanks{Zunhai Su and Bohan Sun contributed equally to this work.}\quad
  Bohan Sun\textsuperscript{1,3}\footnotemark[1]\quad
  Xialie Zhuang\textsuperscript{1}\thanks{Xialie Zhuang and Shuibai Zhang jointly led this project.}\quad
  Shuibai Zhang\textsuperscript{1}\footnotemark[2]\\[2pt]
  He Xiao\textsuperscript{2}\quad
  Hui Shen\textsuperscript{4}\quad
  Jing Xiong\textsuperscript{2}\quad
  Hengyuan Zhang\textsuperscript{2}\\[2pt]
  Zhongzhu Zhou\textsuperscript{5}\quad
  Tiantian Zhang\textsuperscript{6}\quad
  Ngai Wong\textsuperscript{2}\thanks{Ngai Wong and Chuan-Wei Kuo are the corresponding authors.}\quad
  Chuan-Wei Kuo\textsuperscript{1}\footnotemark[3]\\[6pt]
  {\normalfont\small
  \textsuperscript{1}StartLux\quad
  \textsuperscript{2}The University of Hong Kong\quad
  \textsuperscript{3}University of Chinese Academy of Sciences\\[2pt]
  \textsuperscript{4}Independent Researcher\quad
  \textsuperscript{5}University of Sydney\quad
  \textsuperscript{6}Columbia University\endgraf}
  \end{minipage}%
}

\else
  \author{Anonymous authors}
\fi

\begin{document}

\maketitle

\begin{abstract}
Hybrid linear attention large language models (HLA LLMs) interleave linear and full attention to balance efficiency and capacity, yet how hybridization reshapes internal activation dynamics remains poorly understood.
We present the first systematic study of massive activations (MAs) in layer-interleaved HLA LLMs, examining their \textcolor{maOrganization}{\emph{architectural organization}}, \textcolor{maEmergence}{\emph{training-time emergence}}, \textcolor{maMechanisms}{\emph{underlying mechanisms}}, and \textcolor{maFunction}{\emph{functional significance}}.
Across five linear attention architectures, six hybridization configurations, and five input domains, we identify two architecture-aligned morphologies: \emph{pre-attention spikes} (PAS) immediately before full attention and \emph{inter-spike plateaus} (ISP) persisting through intervening linear attention layers.
Denser full attention increasingly connects PAS through ISP, approaching the persistent MAs of conventional Transformers.
This organization also recurs across 12 public checkpoints spanning 1.2B--397B parameters, covering linear attention and state-space hybrids.
Controlled pretraining of Gated DeltaNet (GDN) hybrids up to 1.3B reveals early emergence and consolidation of both morphologies, alongside asymmetric gating effects.
Specifically, full attention output gates strongly attenuate MA magnitudes without eliminating their organization, whereas removing GDN output gates yields modest amplification.
Mechanistically, we develop a shared systematic-outlier account: PAS follows a localized write--sink--cancel process, while ISP is consistent with delayed cancellation.
Functionally, our interventions show that deleting only the four largest-magnitude PAS coordinates at each full attention input reduces mean downstream accuracy by 21.9\%--63.6\% relative to normal inference.
Moreover, reference-conditioned spike-to-plateau connection consistently improves mean real-world retrieval accuracy, yielding relative gains of 1.1\%--12.6\% without retraining.
\ifpreprint
Our code is available at \url{https://github.com/StartLuxLabs/Massive-Activations-HLA}.
\else
Anonymized code is provided in the supplementary material.
\fi
\end{abstract}

\section{Introduction}
\label{sec:introduction}
Softmax attention enables Transformers to model expressive, content-dependent interactions across the entire context~\citep{vaswani2017attention,team2025longcat,team2026longcat,team2026longcat1,team2025introducing}.
However, its computational cost grows quadratically with context length, making long-context modeling expensive~\citep{tay2022efficient,su2026oscar,su2025rotatekv,su2025kvsink}.
Linear attention reduces this cost through recurrent computation, but its fixed-size state restricts modeling capacity, particularly for recall-intensive tasks~\citep{yang2023gated,yang2025gated}.
Hybrid linear attention (HLA) LLMs address this trade-off by interleaving linear and full attention layers.
This design is increasingly common in modern LLMs, including Qwen3-Next, Qwen3.5, Kimi Linear, Kimi K3, Zamba, and Nemotron-H~\citep{yang2025qwen3,team2025kimi,team2026kimi,blakeman2025nemotron,glorioso2024zamba2}.

\begin{figure*}[t]
    \centering
    \vspace{-4mm}
    \includegraphics[width=\textwidth]{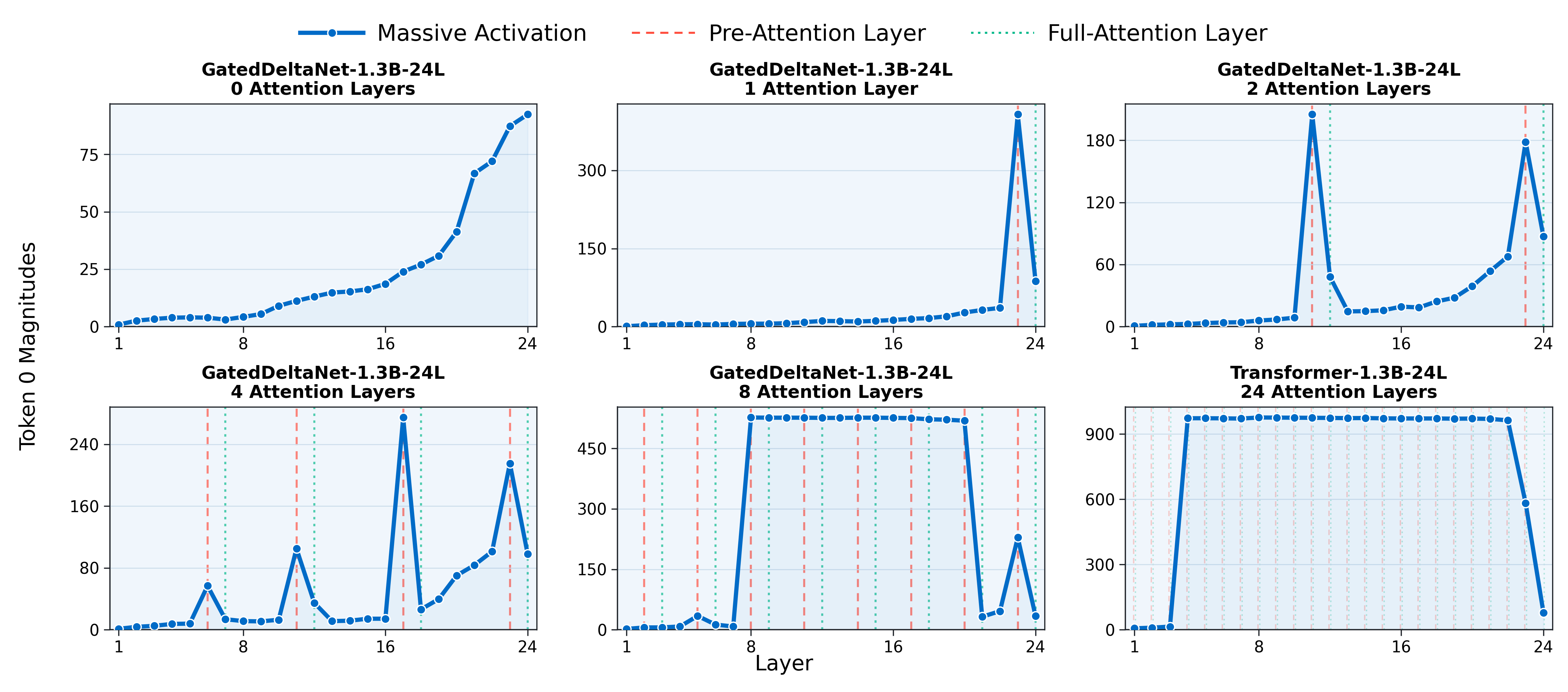}
    \vspace{-8mm}
    \caption{\textbf{MA morphology across GDN hybrid configurations.}
    MAs form \emph{pre-attention spikes} (PAS) immediately before full attention layers.
    Denser full attention promotes persistence between successive PAS, forming \emph{inter-spike plateaus} (ISP).
    Models are from the M-A-P Hybrid Linear Attention Research suite~\citep{wang2025systematic}.}
    \label{fig:overview}
\end{figure*}

Despite their growing adoption, how HLA architectures reshape internal activation dynamics remains poorly understood.
Of particular interest are \emph{massive activations} (MAs), a sparse set of hidden-state entries that exceed typical values by several orders of magnitude and concentrate at specific token positions~\citep{sun2024massive}.
Prior work attributes both their emergence and their close coupling with attention sinks to structural properties of softmax-based Transformers~\citep{su2026attention,xiao2024efficient,sun2026spike,an2025systematic,su2025kvsink}.
These findings establish MAs as a distinctive signature of Transformer dynamics and an informative probe of layerwise computation.
Yet their behavior beyond conventional full attention LLMs remains largely unexplored, raising a central question: \textbf{\textit{How does hybridization shape MA dynamics in HLA LLMs, and what do these dynamics reveal about their internal computation?}}

In this work, we systematically investigate MAs in layer-interleaved HLA LLMs.
Through extensive experiments across five linear attention architectures, six hybridization configurations, and five input domains (\S\ref{sec:architecture-dependence}--\S\ref{sec:ratio-dependence}), we uncover a hybridization-dependent organization distinct from that of full attention LLMs.
As illustrated in Figure~\ref{fig:overview}, MAs intensify immediately before full attention layers, forming \emph{pre-attention spikes} (PAS).
Denser full attention promotes persistence between PAS, producing \emph{inter-spike plateaus} (ISP).
At the full attention limit, these forms merge into the characteristic Transformer morphology, with relatively stable MAs across most intermediate layers.
This organization recurs across 12 public checkpoints from Kimi Linear, Qwen3.5, Nemotron-H, and Zamba2, spanning 1.2B--397B parameters and covering both linear attention and state-space hybrids, including base and instruction-tuned variants (\S\ref{sec:large-scale-generalization}).

Beyond inference-time characterization, we trace the emergence of PAS and ISP through controlled pretraining of Gated DeltaNet (GDN) hybrids and examine the effects of output gating (\S\ref{sec:controlled-pretraining}).
We further apply systematic-outlier analysis to their cross-layer evolution, linking PAS and ISP to the stable MA morphology of full attention models (\S\ref{sec:spike-lifecycle}--\S\ref{sec:plateau-lifecycle}).
Finally, targeted interventions demonstrate the functional importance of PAS and the potential of spike-to-plateau connection to improve real-world retrieval without retraining (\S\ref{sec:functional-interventions}).
Our main contributions are:
\begin{itemize}
    \item \textbf{Architecture-aligned MA morphology.}
    We present the first systematic study of MAs in layer-interleaved HLA LLMs, identifying PAS and ISP as two architecture-aligned morphologies that recur across architectures, hybridization configurations, input domains, and pretrained hybrids spanning 1.2B--397B parameters.

    \item \textbf{Controlled emergence and gating.}
    Pretraining GDN hybrids up to 1.3B shows that PAS and ISP emerge early and consolidate during optimization.
    Full attention output gating strongly attenuates their magnitudes but preserves their layerwise organization, whereas removing GDN output gates yields modest amplification.

    \item \textbf{A shared lifecycle explanation.}
    We develop a shared systematic-outlier account organized by MA cancellation timing.
    PAS follows a localized \emph{write--sink--cancel} process, while the extended persistence of ISP is consistent with delayed cancellation; this account connects both morphologies to the stable MAs observed at the full attention limit.

    \item \textbf{Functional importance and real-world retrieval improvement.}
    Through interventions across six frozen GDN hybrids and eleven benchmarks, we establish the functional importance of PAS and show that reference-conditioned spike-to-plateau connection consistently improves mean real-world retrieval performance without retraining.
\end{itemize}

\section{Preliminaries on Hybrid Linear Attention Models}
\label{sec:preliminaries}

\begingroup
\paragraph{Hybrid Residual Blocks.}
For a length-$T$ input with hidden width $d$, an $L$-block pre-normalized HLA model updates the residual stream $\mathbf{X}^{(0)}\in\mathbb{R}^{T\times d}$ as
\begin{align}
    \mathbf{H}^{(\ell)}
    &=
    \mathbf{X}^{(\ell-1)}
    +
    \operatorname{Mixer}^{(\ell)}
    \left(
        \operatorname{Norm}^{(\ell)}_{\mathrm{mix}}
        \left(\mathbf{X}^{(\ell-1)}\right)
    \right), \\
    \mathbf{X}^{(\ell)}
    &=
    \mathbf{H}^{(\ell)}
    +
    \operatorname{FFN}^{(\ell)}
    \left(
        \operatorname{Norm}^{(\ell)}_{\mathrm{ffn}}
        \left(\mathbf{H}^{(\ell)}\right)
    \right),
\end{align}
where $\mathbf{X}^{(\ell)}$ is the residual stream after block $\ell$. The mixer is full attention for $\ell\in\mathcal{I}_{\mathrm{FA}}\subseteq\{1,\ldots,L\}$ and linear attention otherwise. Letting $L_{\mathrm{FA}}=|\mathcal{I}_{\mathrm{FA}}|$, we define the \emph{hybridization ratio} as $\rho=L/L_{\mathrm{FA}}$: a $\rho{:}1$ configuration contains one full attention layer per $\rho$ sequence-mixing layers. Larger $\rho$ corresponds to sparser full attention, whereas $\rho=1$ recovers a full attention model.

\paragraph{Full Attention Layers.}
Omitting multi-head notation, full attention computes
\begin{equation}
[\mathbf{Q},\mathbf{K},\mathbf{V}]
=
\mathbf{X}[\mathbf{W}_{Q},\mathbf{W}_{K},\mathbf{W}_{V}],
\qquad
\operatorname{FA}(\mathbf{X})
=
\operatorname{softmax}\!\left(
\frac{\mathbf{Q}\mathbf{K}^{\top}}{\sqrt{d_h}}+\mathbf{M}
\right)\mathbf{V}\mathbf{W}_{O},
\label{eq:full-attention}
\end{equation}
where $d_h$ is the head dimension and $\mathbf{M}$ the causal mask. Full attention enables direct, content-dependent interactions with preceding tokens but incurs quadratic computation in $T$.

\paragraph{Linear Attention Layers.}
Linear attention replaces pairwise token interactions with a fixed-size recurrent state. A canonical recurrent formulation is
\begin{equation}
[\mathbf{q}_t,\mathbf{k}_t,\mathbf{v}_t]
=
\mathbf{x}_t[\mathbf{W}_Q,\mathbf{W}_K,\mathbf{W}_V],
\qquad
\mathbf{S}_t
=
\mathbf{F}_t\mathbf{S}_{t-1}
+
\mathbf{k}_t^\top\mathbf{v}_t,
\qquad
\mathbf{y}_t
=
\mathbf{q}_t\mathbf{S}_t\mathbf{W}_O,
\label{eq:linear-attention}
\end{equation}
where $\mathbf{S}_t$ is the recurrent state and $\mathbf{F}_t$ governs state retention. Variants differ in their state-transition and update rules, but all maintain states whose size is independent of sequence length, enabling linear-time sequence processing without a growing KV cache.

By interleaving full and linear attention across depth, HLA LLMs combine complementary sequence-mixing mechanisms whose interaction may reorganize internal computation across layers.
How this interaction shapes activation dynamics remains poorly understood, motivating our study of the layerwise organization of MAs.
A broader discussion of linear and hybrid linear attention, together with related work on MAs in LLMs, is provided in Appendix~\ref{app:related_work}.
\endgroup

\section{Massive Activation Dynamics under Hybridization}
\label{sec:empirical-characterization}

\subsection{Experimental Setup and Sink-Guided Tracking}
\label{sec:experimental-setup}
\begingroup

\paragraph{Analysis Roadmap.}
We first characterize PAS across linear attention architectures (\S\ref{sec:architecture-dependence}) and ISP across hybridization ratios (\S\ref{sec:ratio-dependence}).
Next, we examine whether this organization recurs in large-scale linear attention and state-space hybrids (\S\ref{sec:large-scale-generalization}).
Finally, controlled pretraining traces the emergence of both morphologies and isolates their response to output gating (\S\ref{sec:controlled-pretraining}).

\ifpreprint\Needspace{6\baselineskip}\fi
\paragraph{Models and Inputs.}
\label{sec:evaluation-inputs}
We use the M-A-P suite~\citep{wang2025systematic}: RetNet, HGRN, Gated Linear Attention (GLA), DeltaNet, and Gated DeltaNet (GDN), at 340M and 1.3B under a unified pretraining recipe.\footnote{\href{https://huggingface.co/collections/m-a-p/hybrid-linear-attention-research}{M-A-P Hybrid Linear Attention Research model collection on Hugging Face.}}
These backbones span fixed-decay, input-dependent gating, and delta-rule updates~\citep{sun2023retentive,qin2023hierarchically,yang2023gated,yang2024parallelizing,yang2025gated}.
We use \textit{``Summer is warm. Winter is cold.''} as a running example.
For cross-domain comparisons, we evaluate 100 inputs each from general prose in WikiText-103, scientific writing in Scientific Papers, mathematical reasoning in GSM8K, Python code in CodeSearchNet, and multilingual text in FLORES-200~\citep{merity2016pointer,cohan2018discourse,cobbe2021training,husain2019codesearchnet,costa2022no}.
Checkpoint and pretraining details appear in Appendix~\ref{app:map-model-suite}.

\paragraph{Sink-Guided Tracking of MAs.}
Prior work traces MAs through layerwise magnitude ranking, with highly activated tokens closely corresponding to attention sinks in full attention models~\citep{sun2024massive,an2025systematic}.
Under hybridization, however, sink tokens exhibit lower and less stable magnitude rankings, while the maximally activated token switches more frequently across layers.
Magnitude ranking alone can therefore conflate trajectories of different tokens.
We complement it with an attention-derived anchor that preserves token identity across depth.

For each input $x$, let $A_{x,q,t}^{(\ell,h)}$ denote the attention probability assigned by query token $q$ to source token $t$ in head $h$ of full attention layer $\ell$.
We select the dominant \emph{consensus sink} as
\begin{equation}
    t_x^\star
    =
    \arg\max_{1\leq t<T}
    \frac{1}{|\mathcal{I}_{\mathrm{FA}}|H|\mathcal{Q}_t|}
    \sum_{\ell\in\mathcal{I}_{\mathrm{FA}}}
    \sum_{h=1}^{H}
    \sum_{q\in\mathcal{Q}_t}
    A_{x,q,t}^{(\ell,h)},
    \qquad
    \mathcal{Q}_t=\{t+1,\ldots,T\},
    \label{eq:main-consensus-sink}
\end{equation}
where $H$ is the number of attention heads.
Averaging across layers and heads reduces local fluctuations, while normalization by $|\mathcal{Q}_t|$ accounts for differences in the number of eligible queries.
Holding this token fixed, we trace its maximum absolute activation as $m_{x,t_x^\star}^{(\ell)}=\|\mathbf{X}_{x,t_x^\star,:}^{(\ell)}\|_\infty=\max_j|X_{x,t_x^\star,j}^{(\ell)}|$.
This preserves token identity while allowing the maximally activated feature to change across depth.
The attention distribution determines only which token is tracked; whether and how strongly that token exhibits an MA remains determined by its activation magnitude.
Further details and diagnostic evidence are provided in Appendix~\ref{sec:identification}.

\endgroup

\subsection{Pre-Attention Spikes across Linear Attention Architectures}
\label{sec:architecture-dependence}
\begingroup

\paragraph{Recurring Pre-Attention Spikes.}
To isolate the effect of the linear attention mechanism, we fix model scale and hybridization ratio at 1.3B and $12{:}1$.
For qualitative comparison, we visualize first-token trajectories for the running example, using the most prevalent sink position as a representative anchor.
Despite differing recurrent updates, all five backbones exhibit pronounced activation maxima immediately before full attention (Figure~\ref{fig:la-architecture-pas}).
We term these architecture-aligned local maxima \emph{pre-attention spikes} (PAS).
Beyond this example, Appendix~\ref{app:map-token-dynamics} provides token-wise analyses, while Appendix~\ref{app:map-domain-dynamics} documents recurrence across five domains and both model scales.

\begin{figure*}[t]
\centering
\vspace{-3mm}
\includegraphics[width=\textwidth]{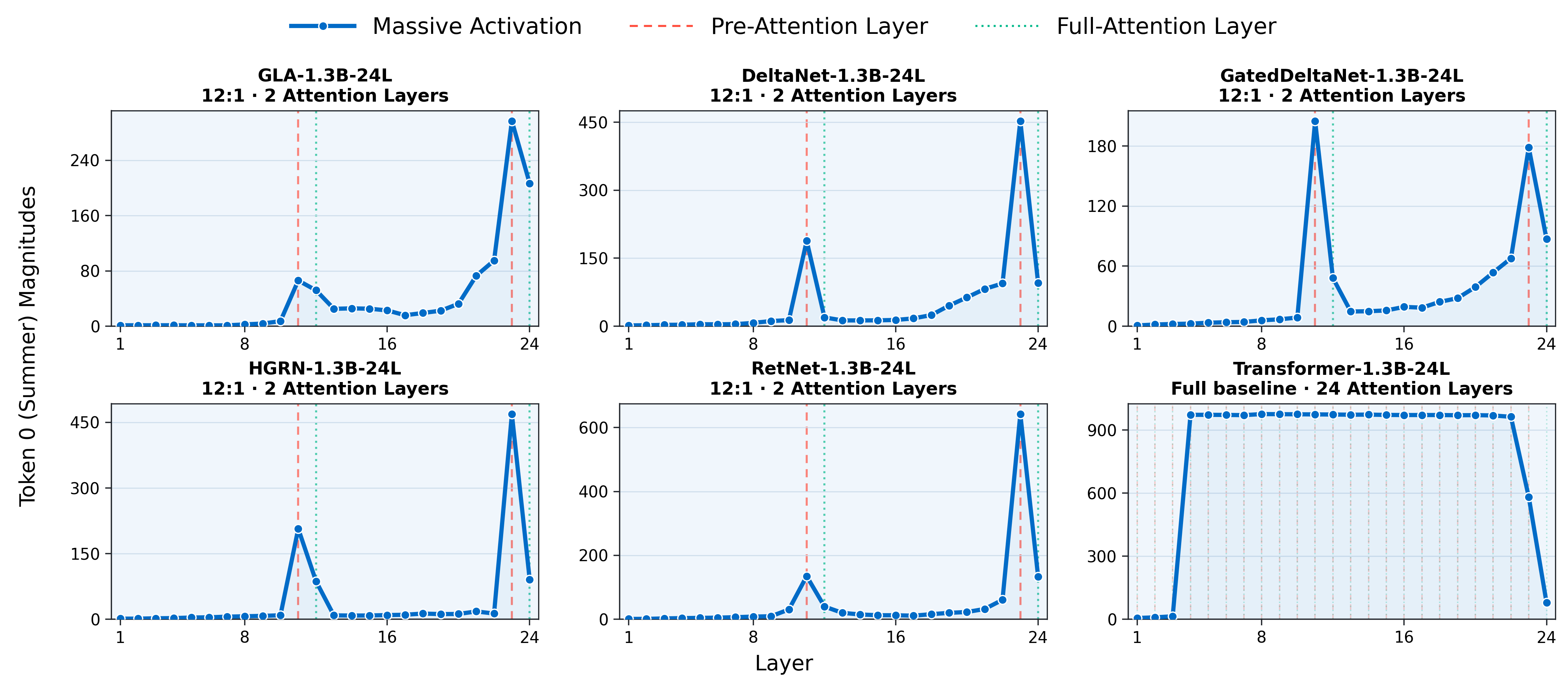}
\vspace{-8mm}
\caption{\textbf{PAS across linear attention architectures.}
First-token MA trajectories for 1.3B M-A-P hybrids at $12{:}1$, with a full attention Transformer for reference.
All five hybrid backbones exhibit activation maxima immediately before full attention.}
\label{fig:la-architecture-pas}
\end{figure*}

\paragraph{Pre-Attention Localization.}
We quantify alignment using the input-specific consensus sink $t_x^\star$ from \S\ref{sec:experimental-setup}.
For each full attention layer $f$, let $\mathcal{B}_f$ denote the contiguous preceding linear attention block, including $f-1$.
For an evaluation set $\mathcal{D}$, the \emph{sink--spike alignment rate} is
\begin{equation}
\operatorname{Align}(\mathcal{D})
=
\frac{1}{|\mathcal{D}|\,|\mathcal{I}_{\mathrm{FA}}|}
\sum_{x\in\mathcal{D}}
\sum_{f\in\mathcal{I}_{\mathrm{FA}}}
\mathbf{1}
\left[
f-1
\in
\operatorname*{arg\,max}_{\ell\in\mathcal{B}_f}
m_{x,t_x^\star}^{(\ell)}
\right],
\label{eq:pas-alignment}
\end{equation}
where $\mathbf{1}[\cdot]$ is the indicator function.
This measures the fraction of input--attention-layer pairs with blockwise maxima immediately before full attention.
Across 500 inputs (100 per domain), domain-macro-averaged alignment reaches $99.4$--$100\%$ for all ten backbone--scale pairs at 340M and 1.3B (Table~\ref{tab:pas-alignment}).
Table~\ref{tab:pas-alignment-domains} in Appendix~\ref{sec:identification} provides domain-specific results.
We assess peak strength through prominence, comparing consensus sinks with first-token and non-sink anchors; a random-layer baseline tests localization.
Paired bootstrap analyses support stronger localization and prominence for consensus sinks than non-sink controls (Appendix~\ref{app:map-statistical-controls}).

\begin{table}[t]
\centering
\begin{minipage}[t]{0.38\linewidth}
\ifpreprint\vspace{0pt}\else\vspace{-6mm}\fi
\caption{\textbf{Sink--spike alignment (\%).}
$12{:}1$ hybrids; macro-average over five domains with 100 inputs each.}
\vspace{-3mm}
\label{tab:pas-alignment}
\centering
\small
\setlength{\tabcolsep}{8pt}
\begin{tabular}{@{}lrr@{}}
\toprule
Backbone & 1.3B & 340M \\
\midrule
RetNet   & 99.9  & 100.0 \\
HGRN     & 100.0 & 100.0 \\
GLA      & 100.0 & 99.6  \\
DeltaNet & 100.0 & 99.4  \\
GDN      & 100.0 & 100.0 \\
\bottomrule
\end{tabular}
\vspace{-2mm}
\end{minipage}
\hfill
\begin{minipage}[t]{0.58\linewidth}
\ifpreprint\vspace{0pt}\else\vspace{-6mm}\fi
\caption{\textbf{Inter-spike retention (\%).}
Each entry reports $1.3\mathrm{B}/340\mathrm{M}$, averaged over the same five domains.}
\vspace{-3mm}
\label{tab:isp-retention}
\centering
\small
\setlength{\tabcolsep}{3pt}
\begin{tabular}{@{}lccc@{}}
\toprule
Backbone
& $\mathbf{12{:}1}$
& $\mathbf{6{:}1}$
& $\mathbf{3{:}1}$ \\
\midrule
RetNet   & 18.8 / 36.8 & 60.0 / 69.8 & 85.4 / 87.2 \\
HGRN     & 7.2 / 5.0   & 40.2 / 49.6 & 92.5 / 81.2 \\
GLA      & 27.3 / 31.9 & 51.7 / 61.6 & 88.0 / 88.1 \\
DeltaNet & 23.1 / 46.5 & 44.6 / 82.8 & 84.4 / 99.8 \\
GDN      & 18.4 / 39.7 & 26.6 / 45.2 & 77.8 / 86.7 \\
\bottomrule
\end{tabular}
\vspace{-2mm}
\end{minipage}
\end{table}

\endgroup

\subsection{From Spikes to Plateaus across Hybridization Ratios}
\label{sec:ratio-dependence}
\begingroup

To isolate hybridization effects, we fix the backbone and model scale while varying $\rho\in\{24,12,6,3\}$.
Pure linear attention and full attention provide limiting reference configurations.

\paragraph{From Localized Spikes to Sustained Plateaus.}
Figure~\ref{fig:overview} illustrates this progression in GDN.
PAS persists across hybrid configurations, but inter-spike activity changes with attention density.
Sparse full attention produces sharply localized peaks separated by activation troughs.
Denser full attention elevates inter-spike activations, connecting successive PAS into sustained \emph{inter-spike plateaus} (ISP).
The onset is backbone-dependent: prominent ISP appears at $6{:}1$ in DeltaNet but primarily at $3{:}1$ in GDN (Appendix~\ref{app:map-ma-dynamics}).
Thus, attention density organizes the progression, with backbone-dependent transition points.
At the full attention limit, spikes and plateaus merge into stable MAs across most intermediate layers, linking hybrid morphologies to those of conventional Transformers.

\paragraph{Inter-Spike Persistence.}
To quantify persistence, let
$\mathcal{I}_{\mathrm{FA}}=\{f_1<\cdots<f_K\}$
denote the full attention layers, and let $p_i=f_i-1$
denote their input positions.
We define
$\mathcal{J}_i=\{\ell:p_i<\ell<p_{i+1},\,
\ell\notin\mathcal{I}_{\mathrm{FA}}\}$
as the intervening linear-attention layers, excluding
the output of the full-attention block at $f_i$.
Using the fixed consensus sink from
\S\ref{sec:experimental-setup}, we define the
\emph{inter-spike retention score} as
\begin{equation}
\operatorname{ISR}(\mathcal{D})
=
\frac{1}{|\mathcal{D}|}
\sum_{x\in\mathcal{D}}
\frac{1}{K-1}
\sum_{i=1}^{K-1}
\frac{1}{|\mathcal{J}_i|}
\sum_{\ell\in\mathcal{J}_i}
\min\!\left(
1,\,
\frac{
m_{x,t_x^\star}^{(\ell)}
}{
\min\!\left(
m_{x,t_x^\star}^{(p_i)},
m_{x,t_x^\star}^{(p_{i+1})}
\right)
}
\right).
\label{eq:inter-spike-retention}
\end{equation}
ISR normalizes activation at the intervening linear-attention
layers by the weaker of the two PAS endpoints,
$p_i$ and $p_{i+1}$; values near zero indicate decay
and near one sustained persistence.
Clipping at one prevents above-reference activations
from dominating the score.
The metric requires at least two PAS and a nonempty
$\mathcal{J}_i$ for every adjacent pair; it is inapplicable
to single-spike configurations.
On the same 500 inputs, ISR increases with full attention
density across all five backbones and both scales
(Table~\ref{tab:isp-retention}).
Paired 95\% bootstrap confidence intervals for all
20 adjacent-ratio differences lie above zero.
To distinguish persistence from endpoint normalization
effects, Appendix~\ref{app:map-statistical-controls}
reports absolute inter-spike activation.

\endgroup

\subsection{Recurrence in Large-Scale Pretrained Hybrid Models}
\label{sec:large-scale-generalization}
\begingroup

We evaluate 12 public checkpoints from Kimi Linear, Qwen3.5, Nemotron-H, and Zamba2, spanning 1.2B--397B parameters~\citep{team2025kimi,yang2025qwen3,blakeman2025nemotron,glorioso2024zamba2}.
These families cover linear attention and state-space mixers, periodic and nonuniform attention schedules, and base and instruction-tuned variants.
Kimi Linear combines Kimi Delta Attention with periodic global MLA layers; Qwen3.5 uses three GDN layers followed by one output-gated full attention layer ($\rho=4$ under our definition).
Nemotron-H combines Mamba-2 with sparse, partly nonuniform self-attention, whereas Zamba2 periodically inserts shared Transformer blocks into a Mamba-2 backbone.
These designs test whether PAS--ISP organization extends beyond a particular recurrent mechanism, ungated attention, or periodic layout.

\paragraph{Consistent Organization, Variable Magnitude.}
First, matched Kimi Linear Base and Instruct checkpoints retain closely aligned PAS locations and ISP boundaries despite magnitude differences, showing persistence across post-training.
Second, across Qwen3.5 scales, PAS and ISP track full attention placement even as magnitude and prominence vary non-monotonically.
Layerwise organization is thus more stable than amplitude across these checkpoints.
Third, recurrence in Nemotron-H and Zamba2 extends the pattern to state-space hybrids with sparse and partly nonuniform attention schedules.
Fourth, across representative inputs from five domains, PAS locations and ISP spans remain stable while magnitudes vary.
Together, these representative traces extend the architectural pattern beyond the M-A-P suite.
Appendices~\ref{app:large-scale-model-suite} and~\ref{app:large-scale-ma-dynamics} provide configurations and all checkpoint and domain traces.
\rev{Five-domain means show higher absolute plateau heights at the attention sink than at either non-sink control in all 12 checkpoints (Appendix~\ref{app:large-population-statistics}).}

\endgroup

\subsection{Controlled Pretraining Studies of PAS and ISP}
\label{sec:controlled-pretraining}
\begingroup

\begin{figure*}[t]
    \centering
    \begin{subfigure}[t]{0.32\textwidth}
        \centering

        \includegraphics[width=\linewidth]{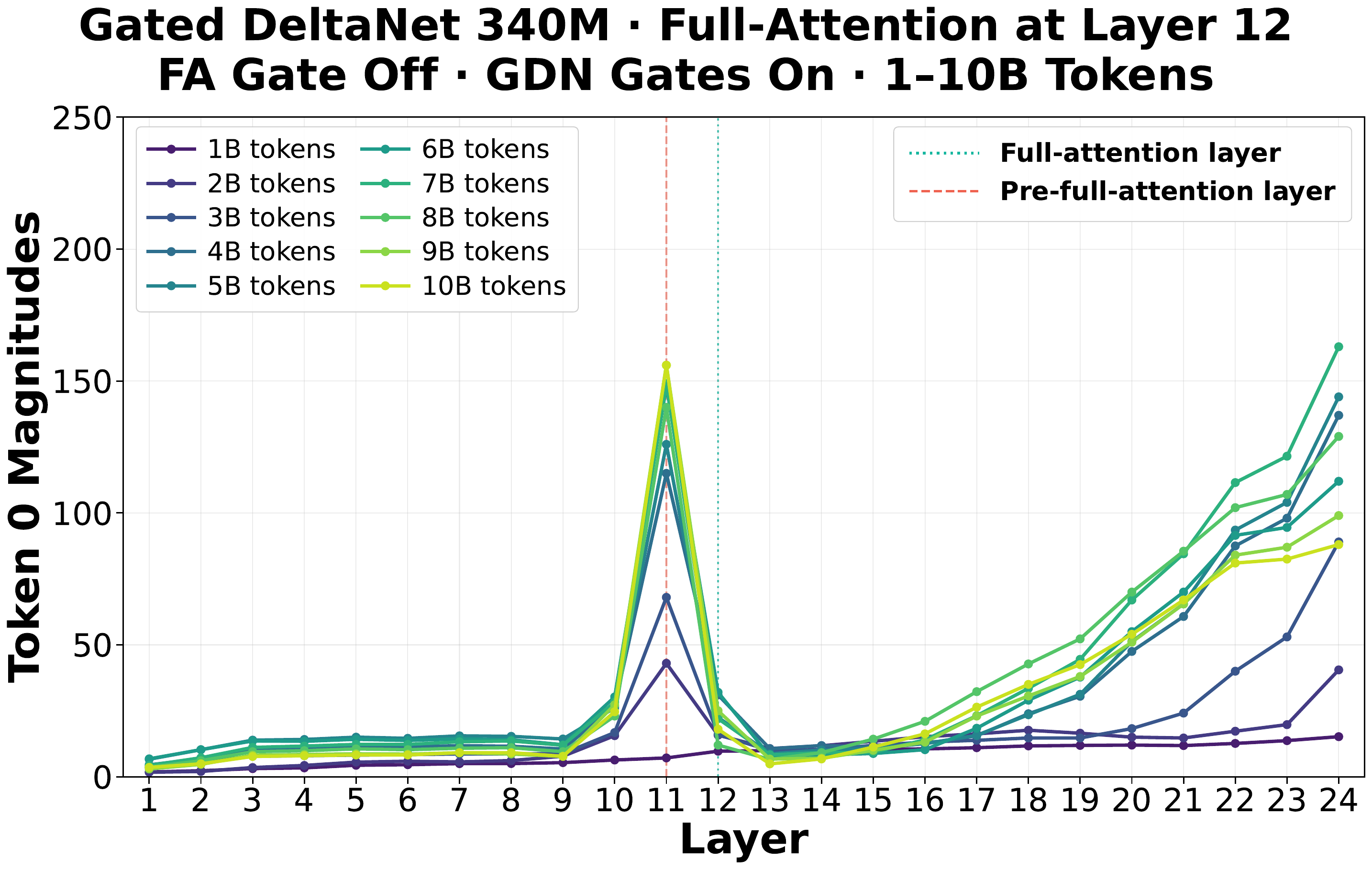}
        \vspace{-3mm}
        \caption{Standard GDN.}
        \label{fig:main-pas-standard}
    \end{subfigure}
    \hfill
    \begin{subfigure}[t]{0.32\textwidth}
        \centering

        \includegraphics[width=\linewidth]{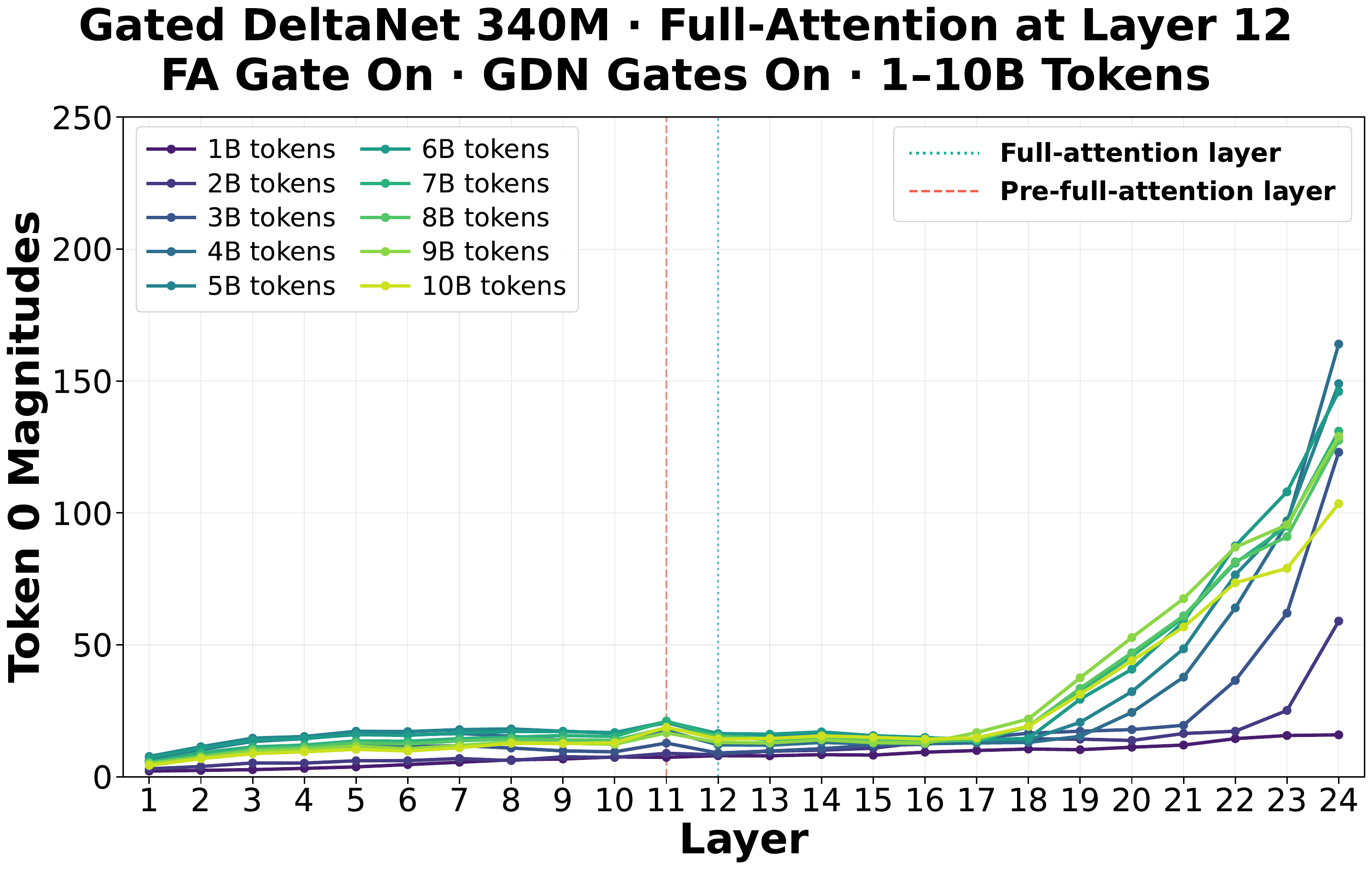}
        \vspace{-3mm}
        \caption{Full attention output gating.}
        \label{fig:main-pas-gatedfa}
    \end{subfigure}
    \hfill
    \begin{subfigure}[t]{0.32\textwidth}
        \centering

        \includegraphics[width=\linewidth]{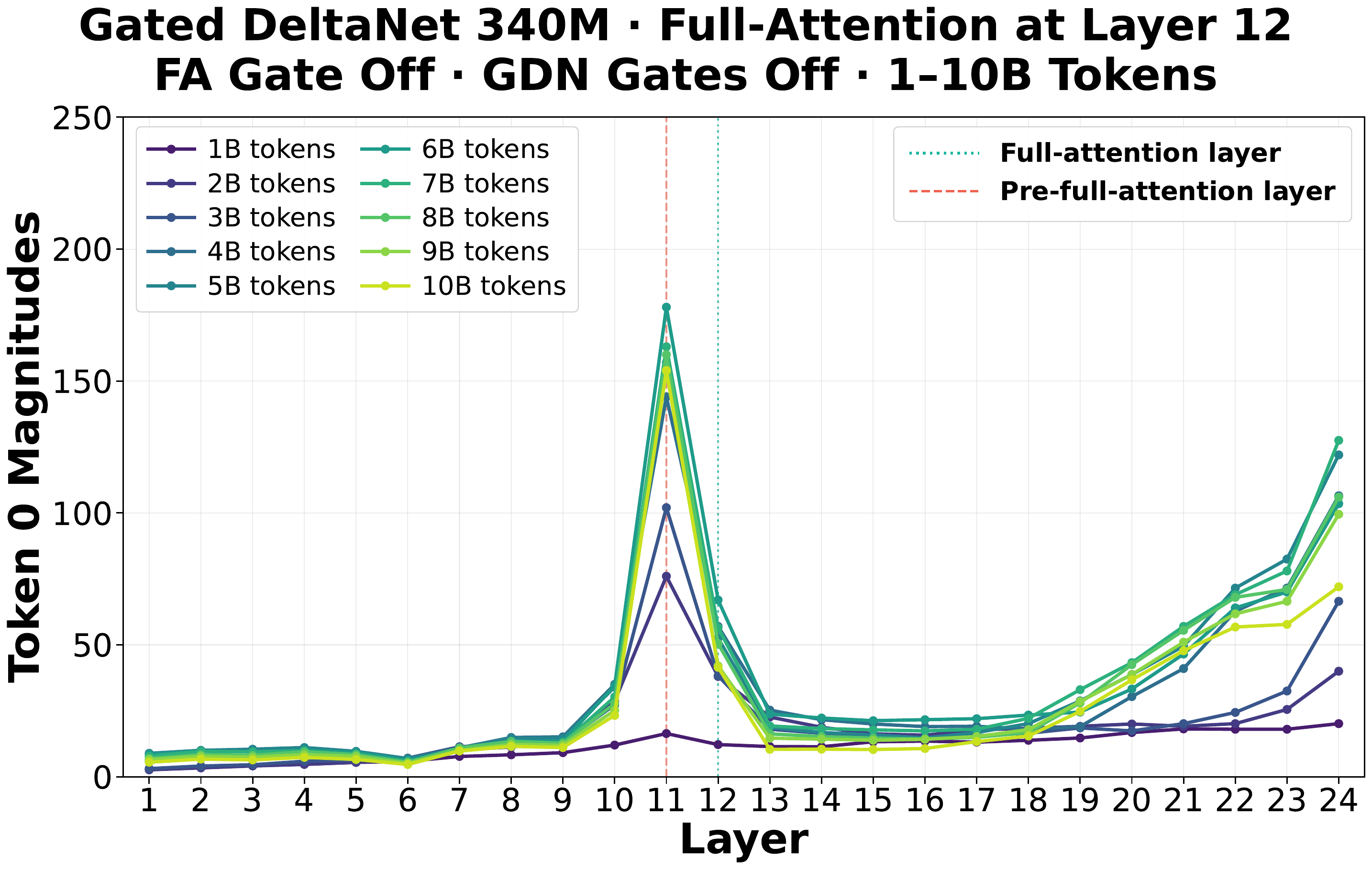}
        \vspace{-3mm}
        \caption{GDN output gates removed.}
        \label{fig:main-pas-nooutgate}
    \end{subfigure}
    \caption{\textbf{PAS emergence and gating response.}
    First-token trajectories at successive 1B-token checkpoints in a 340M GDN model with full attention at layer 12.
    PAS emerges early and strengthens; full attention output gating attenuates but does not eliminate it, whereas removing GDN output gates moderately amplifies it.}
    \label{fig:main-pas-gating-dynamics}
\end{figure*}

\paragraph{Controlled Pretraining Models.}
Inference-time recurrence leaves open when these morphologies arise and how architectural choices shape their development.
Using Flash Linear Attention~\citep{yang2024fla}, we pretrain 24-layer GDN hybrids from scratch at 340M and 1.3B.
Within each comparison, only the targeted layer placement or gating condition changes; all other architectural and training settings remain fixed (Appendix~\ref{app:controlled-training-setup}).

\paragraph{Training-Time Emergence of PAS and ISP.}
To isolate PAS, we place a single full attention layer at an early, middle, or late depth in 340M models and repeat the middle placement at 1.3B.
For ISP, we train both scales at $3{:}1$.
We trace first-token MA trajectories every 1B tokens at 340M and every 5B at 1.3B.
Across these configurations, PAS consistently forms immediately before full attention and shifts with its placement.
Under matched training conditions, this relocation links PAS to full attention placement rather than a preferred absolute depth.
PAS is visible by 1B tokens in the 340M layer-12 configuration (Figure~\ref{fig:main-pas-standard}) and becomes more pronounced and stable.
ISP likewise emerges early and consolidates during optimization (Figure~\ref{fig:controlled-isp-dynamics}).
Both morphologies therefore develop during pretraining, rather than being artifacts of checkpoint selection (Appendices~\ref{app:controlled-pas} and~\ref{app:controlled-isp}).

\paragraph{Asymmetric Effects of Output Gating.}
Recent work reports that element-wise output gating can eliminate MAs and their associated sinks in full attention Transformers~\citep{su2026attention,qiu2026gated}.
We test this in hybrids by adding output gates to full attention layers and removing native GDN output gates.
Full attention output gating markedly attenuates PAS, although a small pre-attention peak persists throughout training (Figure~\ref{fig:main-pas-gatedfa}).
Removing GDN output gates produces only moderate amplification, despite modifying more layers (Figure~\ref{fig:main-pas-nooutgate}).
ISP exhibits the same asymmetry: full attention gating substantially reduces the plateau, whereas removing GDN gates has a comparatively modest effect (Figure~\ref{fig:controlled-isp-gating-dynamics}; Appendix~\ref{app:output-gating}).
Crucially, full attention output gating attenuates MA magnitudes without eliminating their architecture-aligned organization: weak PAS and ISP persist and become more visible during training.
Consistently, Qwen3.5 models with native full attention output gating retain both morphologies (Figures~\ref{fig:qwen35-35b-base}--\ref{fig:qwen35-397b}; Appendix~\ref{app:large-scale-ma-dynamics}).

\endgroup

\section{A Systematic-Outlier Account of PAS and ISP}
\label{sec:lifecycle-analysis}

\begingroup
The preceding analyses establish MA locations and training-time emergence; we now examine their cross-layer mechanisms and functional significance.
Building on systematic-outlier analysis~\citep{an2025systematic,su2025kvsink,su2025unveiling}, we trace signed residual updates and verify a representative PAS at a fixed token--feature coordinate.
Together with recurring cross-model patterns, this supports a shared lifecycle distinguished by cancellation timing.
We examine localized PAS and persistent ISP, then test how spike suppression and extended persistence affect performance.
\endgroup


\subsection{Pre-Attention Spikes: Localized Write--Sink--Cancel}
\label{sec:spike-lifecycle}
\begingroup

\paragraph{Localized Write--Sink--Cancel.}
Across backbones and hybridization configurations, PAS appears as a coordinated cross-layer event rather than an isolated numerical fluctuation.
In a representative 1.3B GDN hybrid at $12{:}1$, large outliers recur immediately before full attention and dissipate afterward (Figure~\ref{fig:pas-lifecycle}).
Signed decomposition reveals recurring write--cancel pairs; fixed-coordinate tracing then tests whether the same feature is written and cancelled.

\begin{figure*}[!htbp]
\centering
\includegraphics[width=\textwidth]{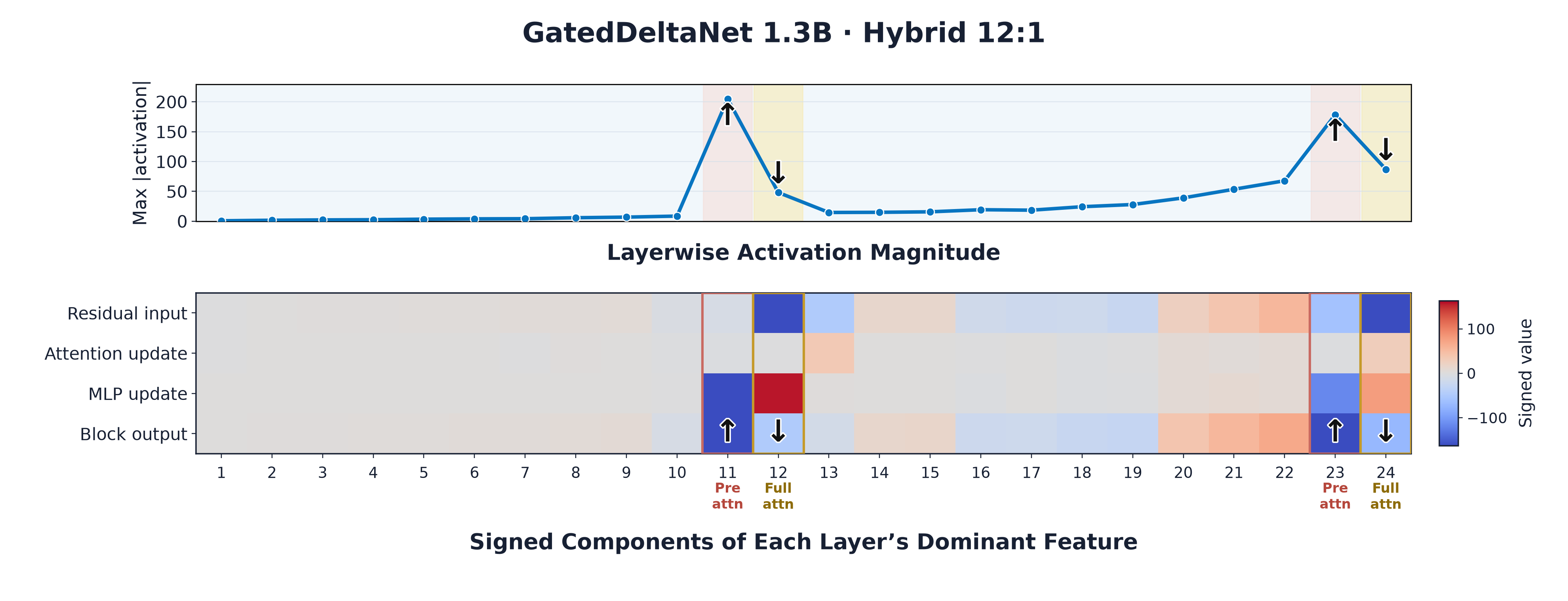}
\vspace{-8mm}
\caption{\textbf{Localized write--sink--cancel lifecycle.}
A 1.3B GDN hybrid at $12{:}1$ writes an outlier before full attention and cancels it with an opposite-signed update.}
\label{fig:pas-lifecycle}
\end{figure*}

\paragraph{Fixed-Coordinate Verification.}
We therefore examine PAS preceding $f=12$ in the same model.
Fixing the consensus sink $t^\star$ and feature $j^\star=\operatorname*{arg\,max}_{j}|X_{t^\star,j}^{(f-1)}|$, we trace signed activations and module-level updates at this coordinate.
Unlike the morphological envelope, this trace follows one feature throughout.
The residual decomposition separates the incoming state, mixer update, FFN update, and block output.
The signs of these components distinguish reinforcement from cancellation.
This trace identifies three stages: layer 11 writes the outlier; during layer-12 full attention, its token attracts a disproportionate share of attention from later queries and acts as a sink; an opposite-signed update in that layer substantially cancels the same outlier.
Signed residual addition verifies cancellation at this fixed coordinate.
Together, these stages verify a localized \emph{write--sink--cancel} event linking residual dynamics to attention-sink behavior.
Appendix~\ref{app:detailed-pas-lifecycle} visualizes each stage; Appendix~\ref{app:layerwise-lifecycle} extends the layerwise analysis to M-A-P and large pretrained linear attention and state-space hybrids.

\endgroup

\subsection{Inter-Spike Plateaus: Extended Outlier Persistence}
\label{sec:plateau-lifecycle}
\begingroup

\paragraph{Extended Persistence through Delayed Cancellation.}
Signed ISP analysis reveals the same write--cancel pattern over greater depth.
Outlier magnitudes rise immediately before full attention and remain elevated across intervening linear attention layers, with opposite-signed updates as the plateau dissipates (Figure~\ref{fig:isp-lifecycle}).
Persistence followed by an opposite-signed update supports two regimes of a shared lifecycle: prompt cancellation localizes PAS, whereas delayed cancellation is consistent with sustained ISP (cross-model traces in Appendix~\ref{app:layerwise-lifecycle}).

\begin{figure*}[!htbp]
    \centering
    \includegraphics[width=\textwidth]{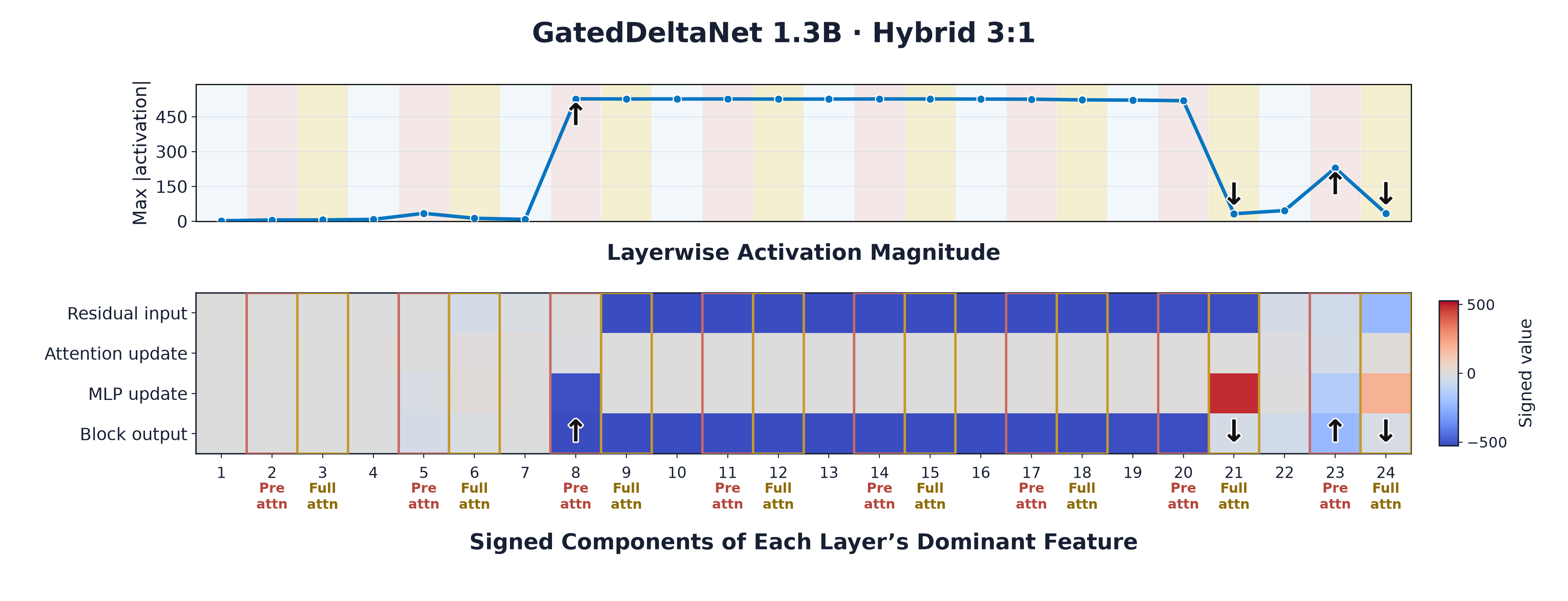}
    \vspace{-8mm}
    \caption{\textbf{ISP dynamics consistent with delayed cancellation.}
    Elevated activations persist between full attention layers, with opposite-signed updates as the plateau dissipates. Signed components are evaluated at each layer's dominant feature.}
    \label{fig:isp-lifecycle}
\end{figure*}

\paragraph{A Continuum toward Full Attention.}
The hybridization trend in \S\ref{sec:ratio-dependence} admits the same interpretation.
As full attention becomes denser, elevated inter-spike activations bridge adjacent PAS, consistent with extended persistence.
At the full attention limit, the distinction between spikes and intervening plateaus disappears, recovering stable MAs across much of the network depth.
PAS, ISP, and persistent full attention MAs thus form a morphological continuum that the cancellation-timing account helps interpret.
Here, persistence refers to the residual activation trajectory across \emph{depth}.
We next test how altering attention-input outliers and their persistence affects model behavior.

\endgroup

\subsection{Functional Interventions: Removing Spikes and Extending Persistence}
\label{sec:functional-interventions}
\begingroup

We investigate the functional significance of PAS--ISP organization through two complementary interventions: suppressing PAS at full attention inputs and connecting successive spikes into sustained plateaus.
We evaluate six released M-A-P GDN checkpoints spanning two scales (340M and 1.3B) and three hybridization ratios ($\rho\in\{12,6,3\}$), with model weights and attention-layer placement fixed.
Both interventions use the same examples and scoring rules as normal inference across eleven benchmarks: ten accuracy tasks and one perplexity benchmark.
Real-world retrieval, including SQuAD, is evaluated by cloze-completion accuracy.
Table~\ref{tab:functional-summary} summarizes results; Appendix~\ref{app:functional-interventions} provides complete protocols and model-specific scores.

\begin{table}[!ht]
\centering
\small
\setlength{\tabcolsep}{5pt}
\caption{\textbf{PAS suppression and spike-to-plateau connection across six frozen GDN hybrids.}
$k=4$ coordinates are selected per full attention input.
Scores are averaged equally across models; parentheses report relative changes from normal inference, $100(\bar S_{\rm int}/\bar S_{\rm normal}-1)$.
Accuracy is in percent ($\uparrow$); perplexity is reported separately ($\downarrow$).
The ten-task mean weights accuracy tasks equally, excluding perplexity; the real-world retrieval mean averages FDA, SWDE, SQuAD, and NQ.
Complete model-specific results for all $k$ appear in Appendix~\ref{app:functional-full-results}.}
\label{tab:functional-summary}
\begingroup
\begin{tabular}{@{}lrrr@{}}
\toprule
Benchmark & Normal & PAS suppression & Spike connection \\
\midrule
PIQA & 70.13 & 66.78 (${-4.77\%}$) & 70.10 (${-0.04\%}$) \\
ARC-Easy & 65.55 & 56.40 (${-13.95\%}$) & 65.38 (${-0.27\%}$) \\
HellaSwag & 51.11 & 45.71 (${-10.57\%}$) & 51.07 (${-0.08\%}$) \\
FDA & 45.37 & 15.32 (${-66.23\%}$) & 46.05 (${+1.50\%}$) \\
SWDE & 35.57 & 16.06 (${-54.85\%}$) & 37.72 (${+6.06\%}$) \\
SQuAD & 37.78 & 32.02 (${-15.26\%}$) & 42.19 (${+11.66\%}$) \\
NQ & 22.42 & 14.14 (${-36.91\%}$) & 22.37 (${-0.19\%}$) \\
NIAH-1 & 71.00 & 19.33 (${-72.77\%}$) & 71.13 (${+0.19\%}$) \\
NIAH-2 & 80.43 & 24.27 (${-69.83\%}$) & 81.13 (${+0.87\%}$) \\
NIAH-3 & 36.93 & 4.27 (${-88.45\%}$) & 37.13 (${+0.54\%}$) \\
WikiText-103 PPL $\downarrow$ & 11.6290 & 20.1000 (${+72.84\%}$) & 11.6238 (${-0.04\%}$) \\
\midrule
Ten-task mean & 51.63 & 29.43 (${-43.00\%}$) & 52.43 (${+1.55\%}$) \\
Real-world retrieval mean & 35.28 & 19.38 (${-45.06\%}$) & 37.08 (${+5.10\%}$) \\
\bottomrule
\end{tabular}
\endgroup
\end{table}

\paragraph{Suppressing PAS.}
Following our first-token analyses, both interventions target the first prompt token.
For each example, let $p_i=f_i-1$ index the residual state entering full attention layer $f_i$, and let $z^{(\ell)}=X^{(\ell)}_{1,:}$ denote its normal reference trajectory before normalization.
At every full attention input, we select $C_i=\operatorname{TopK}(|z^{(p_i)}|,k)$, with $k\in\{1,2,4\}$, and set $\widetilde X^{(p_i)}_{1,j}=0$ for $j\in C_i$.
We fix the coordinate sets from the reference pass and overwrite only the selected entries.
Deleting only four coordinates per full attention input reduces ten-task mean accuracy in all six models, with relative losses of $21.91\%$--$63.60\%$.
The cross-model mean falls from $51.63\%$ to $29.43\%$, accompanied by increased perplexity.
Retrieval is particularly sensitive: cross-model relative losses reach $66.23\%$ on FDA and $88.45\%$ on NIAH-3, compared with $4.77\%$ on PIQA.
These results demonstrate that model performance depends strongly on the selected first-token PAS coordinates.

\paragraph{Connecting Spikes without Retraining.}
We next extend persistence from the first to last full attention input on the union $C=\bigcup_i C_i$.
Here, $k$ specifies the per-layer selection budget, so $C$ may contain more than $k$ distinct coordinates.
Between adjacent inputs, we set
\begin{equation}
\widetilde X^{(\ell)}_{1,j}
=s_{i,j}\max\!\left\{|z^{(p_i)}_j|,\;s_{i,j}z^{(\ell)}_j\right\},
\qquad p_i<\ell<p_{i+1},\quad j\in C,
\label{eq:main-connection}
\end{equation}
where $s_{i,j}=\operatorname{sign}(z^{(p_i)}_j)$, with $+1$ at zero.
The left reference endpoint sets a signed magnitude floor, while larger same-direction reference values are retained.
\rev{The simplified procedure applies Eq.~\ref{eq:main-connection} between FA inputs. At each FA input it uses the residual produced by the preceding layers. The original six-checkpoint connection results below used the combined rule, which also restored the selected reference values at FA inputs.}
A normal forward pass supplies the reference trajectory, and the same rule is applied across examples without answer-based selection or parameter updates.
Across all eighteen model--budget settings, the equal-weight mean accuracy over FDA, SWDE, SQuAD, and NQ improves by $1.06\%$--$12.60\%$ relative to normal inference, with exploratory paired 95\% confidence intervals for the gain above zero (Appendix~\ref{app:functional-full-results}, Table~\ref{tab:retrieval-consistency}).
At $k=4$, the cross-model retrieval mean rises from $35.28\%$ to $37.08\%$, a $5.10\%$ relative gain.
SQuAD and SWDE improve in eighteen and sixteen settings, respectively.
At $k=4$, both improve in all six models, with cross-model relative gains of $11.66\%$ and $6.06\%$.
Task-level regressions occur on FDA, NQ, and NIAH-3, while changes on commonsense benchmarks are comparatively small.
Appendix~\ref{app:open-model-interventions} reports supplementary Qwen3.5-27B results on reasoning, mathematics, and code generation.

\paragraph{Connecting Morphology to Function.}
Together, these interventions demonstrate the functional relevance of MA organization: suppressing selected PAS coordinates can impair performance, whereas reference-conditioned spike-to-plateau connection improves mean real-world retrieval without retraining.
One possible explanation is that extending MA persistence brings hybrid activation dynamics closer to the persistent-MA morphology of full attention Transformers.

\paragraph{\rev{Additional paired controls.}}
\rev{In both M-A-P $\rho=3$ checkpoints, channel-mean replacement suppresses PAS without zeroing and lowers mean retrieval accuracy; random-four deletion stays near normal. Same-rule random-coordinate and non-sink connections stay near normal. With interior editing alone, mean retrieval accuracy rises from 28.91\% to 33.98\% for 340M and from 47.07\% to 50.59\% for 1.3B. Endpoint restoration alone reproduces normal outputs, and combining it with interior editing yields 33.98\% and 49.80\%, respectively. The gains therefore support interior editing alone. Freezing the RMS denominator still leaves substantial deletion losses (Appendices~\ref{app:paired-controls}--\ref{app:normalization-controls}).}

\endgroup

\section{Conclusion}
\label{sec:conclusion}
\begingroup
We identify PAS and ISP as recurring, architecture-aligned MA morphologies in hybrid LLMs.
Controlled pretraining reveals early emergence and asymmetric gating effects that alter magnitude while preserving layerwise organization.
A shared systematic-outlier account links localized write--sink--cancel dynamics in PAS with ISP persistence consistent with delayed cancellation.
Crucially, suppressing PAS coordinates can impair performance, whereas reference-conditioned spike-to-plateau connection consistently improves mean real-world retrieval in the evaluated GDN hybrids without retraining.
Together, these results connect MA organization to model behavior and motivate further study of cancellation depth, effective connection rules, and reference-free retrieval improvement.
\endgroup

\ifpreprint\else\clearpage\fi
\section*{Ethics Statement}
This work analyzes activation dynamics in language models using existing checkpoints and benchmark datasets. The additional interventions do not involve human participants or collect personal information. Benchmark scores characterize the evaluated tasks, not model safety or suitability for deployment. AI assistance is disclosed separately below.

\section*{Reproducibility Statement}
The appendix documents the evaluated checkpoints, datasets, analysis metrics, controlled-pretraining configurations, hyperparameters, and benchmark protocols used in this work.
Appendix~\ref{app:functional-interventions} specifies the functional interventions, evaluation populations, and aggregation rules.
\ifpreprint
Public code is available at \url{https://github.com/StartLuxLabs/Massive-Activations-HLA}. The accompanying supplementary material provides the analysis and intervention code, evaluation configurations, and numerical results.
\else
The anonymized supplementary material provides the analysis and intervention code, evaluation configurations, and numerical results.
\fi

\section*{AI Use Statement}
\begingroup
\rev{The authors led the development of the research ideas, scientific observations, and experiments. We used OpenAI Codex for language editing and to assist with the design, implementation, and verification of the supplemental evaluations. We also used it to check interpretations of the results and to draft the conditional derivations in Appendix~\ref{app:conditional-readout}. The authors reviewed the AI-assisted work and take responsibility for the final manuscript, code, results, and claims.}
\endgroup

\bibliography{iclr2027_conference,functional_references}
\bibliographystyle{iclr2027_conference}

\appendix
\clearpage
\raggedbottom

\section*{Appendix Contents}  %
\startcontents[appendix] 
\begingroup\small
\printcontents[appendix]{}{1}{}
\endgroup
\newpage

\section{Related Work}
\label{app:related_work}
\subsection{Linear Attention and Hybrid Linear Attention Models}
Linear attention replaces pairwise token interactions with recurrent state updates, enabling linear-time sequence processing with fixed-size states.
Representative architectures include RetNet, which introduces decay-based retention; HGRN and GLA, which improve state control through hierarchical and input-dependent gating; and DeltaNet, which employs delta-rule updates to selectively modify associative memory~\citep{sun2023retentive,qin2023hierarchically,yang2023gated,yang2024parallelizing}.
GDN further augments the delta rule with data-dependent forgetting, while closely related state-space models such as Mamba achieve similar efficiency through selective state transitions~\citep{yang2025gated,gu2023mamba}.
Because fixed-size states can limit precise recall, recent LLMs increasingly interleave recurrent mixers with full attention layers.
Representative examples include Jamba and Nemotron-H, which combine Mamba with full attention; MiniMax-01, which combines Lightning Attention with softmax attention; Qwen3-Next and Qwen3.5, which interleave GDN with gated attention; and Kimi Linear, which combines Kimi Delta Attention with Multi-Head Latent Attention~\citep{lieber2024jamba,blakeman2025nemotron,li2025minimax,cao2026qwen3,team2025kimi}.
These hybrid architectures achieve favorable performance--efficiency trade-offs, but their internal activation dynamics remain poorly understood.

\subsection{Massive Activations in Large Language Models}
MAs were first identified as extremely sparse hidden-state outliers whose magnitudes exceed typical activations by several orders of magnitude.
They persist across layers, remain largely invariant to input content, and function as implicit bias terms closely associated with attention sinks~\citep{sun2024massive}.
Subsequent work connects MAs to a broader system of activation, weight, and attention outliers; traces their sources to super experts in MoE LLMs; and links their cross-layer evolution to attention sinks~\citep{an2025systematic,su2025unveiling,su2025kvsink}.
Recent analyses further identify pre-normalization as central to the coupling between MAs and attention sinks, trace their emergence to RMSNorm and FFN computation, and explain their propagation through residual connections~\citep{sun2026spike,shi2026single}.
Beyond their mechanistic role, MAs create a challenging numerical regime for low-bit inference~\citep{zhang2026locate}.
Although MAs are distinct from conventional channel-wise outlier features, their extreme magnitudes broaden the tensor dynamic range; under activation or KV cache quantization, a small number of such values can dominate shared scales, reduce the resolution available to typical values, and amplify quantization error.
This challenge motivates outlier-aware transformations and sink-preserving KV cache quantization methods~\citep{zhang2026beyond,xiao2026exploring,su2025rotatekv,su2026oscar,su2025kvsink,xiong2025dope}.
Training-time studies additionally show that MAs emerge along predictable, architecture-dependent trajectories~\citep{gallego2025hidden}.
Despite this progress, existing work has focused almost exclusively on full attention LLMs, leaving MAs in HLA LLMs largely uncharacterized.

\paragraph{\rev{Relation to softmax and no-update accounts.}}
\rev{\citet{gu2025attention} connect attention sinks to key biases and softmax normalization, and study how sinks emerge during training. \citet{bondarenko2023quantizable} link outliers to attention heads learning no-op or partial residual updates, and reduce them with clipped softmax or gated attention. These findings explain why softmax and outliers are related. Our study examines their depth organization in hybrids: isolated PAS at sparse FA boundaries, ISP through intervening linear layers, and how placement, density and output gating shape the transition. Controlled pretraining, fixed-coordinate lifecycle analysis and targeted deletion/connection tests connect this architectural organization to functional behavior.}

\section{MA Dynamics in the M-A-P Model Suite}
\label{app:map-model-analysis}

\subsection{Model Suite and Pretraining Details}
\label{app:map-model-suite}
Our inference-time analysis uses 52 pretrained checkpoints from the model suite released by~\citet{wang2025systematic}.
The evaluated checkpoints span five linear attention architectures---RetNet, HGRN, GLA, DeltaNet, and GDN---at two parameter scales, 340M and 1.3B.
For each architecture and scale, we consider four hybridization configurations and a pure linear attention baseline, together with a full attention Transformer reference at each scale.
All evaluated checkpoints are publicly available through the M-A-P Hybrid Linear Attention Research collection.\footnote{\href{https://huggingface.co/collections/m-a-p/hybrid-linear-attention-research}{M-A-P model collection on Hugging Face.}}

\paragraph{Linear Attention Families.}
The five architectures considered in our analysis span vector-valued recurrence, matrix-valued associative memory, input-dependent retention, and delta-rule state updates.
They differ primarily in the structure of their recurrent states, the mechanisms governing information retention, and the rules used to incorporate new information.
HGRN maintains a compact vector-valued state controlled by input-dependent forget gates.
RetNet employs a matrix-valued state with fixed exponential decay.
GLA replaces fixed decay with input-dependent multiplicative gates, allowing retention to vary across tokens and state dimensions.
DeltaNet applies the delta rule to correct existing key--value associations rather than unconditionally accumulating new outer products.
GDN combines delta-rule correction with input-dependent forgetting, enabling flexible control over state retention and overwriting.
Table~\ref{tab:linear-attention-families} summarizes these distinctions.

\begin{table}[t]
    \centering
    \small
    \setlength{\tabcolsep}{4pt}
    \caption{
        Linear attention architectures evaluated in our inference-time analysis.
        The architectures differ in state structure, retention mechanism, and update rule.
    }
    \label{tab:linear-attention-families}
    \begin{tabular}{llll}
        \toprule
        \textbf{Architecture}
        & \textbf{State Structure}
        & \textbf{Retention Mechanism}
        & \textbf{Update Rule} \\
        \midrule
        HGRN
        & Vector
        & Input-dependent forgetting
        & Gated recurrence \\
        RetNet
        & Matrix
        & Fixed exponential decay
        & Outer-product accumulation \\
        GLA
        & Matrix
        & Input-dependent decay
        & Gated outer-product accumulation \\
        DeltaNet
        & Matrix
        & No explicit decay
        & Delta-rule correction \\
        GDN
        & Matrix
        & Input-dependent forgetting
        & Gated delta-rule correction \\
        \bottomrule
    \end{tabular}
\end{table}

\paragraph{Hybridization Configurations.}
All evaluated models comprise 24 sequence-mixing layers.
Within each model family, we hold the linear attention architecture fixed while varying the number of full attention layers, enabling controlled comparisons of how hybridization shapes MA dynamics.

We define the hybridization ratio as $\rho=L/L_{\mathrm{FA}}$, where $L=24$ is the total number of sequence-mixing layers and $L_{\mathrm{FA}}$ is the number of full attention layers.
A $\rho{:}1$ configuration therefore contains one full attention layer per $\rho$ sequence-mixing layers.
Larger values of $\rho$ correspond to sparser full attention.
The evaluated checkpoints cover four hybrid configurations, with $\rho\in\{3,6,12,24\}$.
We additionally include a full attention Transformer with $\rho=1$ and, by convention, a pure linear attention model corresponding to $\rho=\infty$.
Together, these configurations span the spectrum from exclusively full attention to exclusively linear attention.

\begin{table}[t]
    \centering
    \small
    \setlength{\tabcolsep}{5pt}
    \caption{
        Hybridization configurations considered in our inference-time analysis.
        All models comprise 24 sequence-mixing layers.
    }
    \label{tab:hybrid-configurations}
    \begin{tabular}{lccc}
        \toprule
        \textbf{Configuration}
        & \(\boldsymbol{\rho}\)
        & \textbf{Linear Attention}
        & \textbf{Full Attention} \\
        \midrule
        Full attention
        & $1$
        & $0$
        & $24$ \\
        $3{:}1$
        & $3$
        & $16$
        & $8$ \\
        $6{:}1$
        & $6$
        & $20$
        & $4$ \\
        $12{:}1$
        & $12$
        & $22$
        & $2$ \\
        $24{:}1$
        & $24$
        & $23$
        & $1$ \\
        Pure linear attention
        & $\infty$
        & $24$
        & $0$ \\
        \bottomrule
    \end{tabular}
\end{table}

\paragraph{Original Pretraining Recipe.}
All checkpoints were pretrained from scratch on FineWeb-Edu~\citep{penedo2024fineweb} using implementations from the Flash Linear Attention library~\citep{yang2024fla}.
Within each parameter scale, all architectures and hybridization configurations share the same training corpus and optimization recipe, minimizing variation unrelated to the sequence-mixing architecture or full attention density.
The 340M models were trained on 20B tokens with a global batch size of 50K tokens, whereas the 1.3B models were trained on 100B tokens with a global batch size of 1M tokens.
Both scales were optimized using AdamW with a cosine learning-rate schedule.
Table~\ref{tab:pretraining-configurations} summarizes the original pretraining configurations.

\begin{table}[t]
    \centering
    \small
    \setlength{\tabcolsep}{5pt}
    \caption{
        Original pretraining configurations of the checkpoints used in our inference-time analysis.
    }
    \label{tab:pretraining-configurations}
    \begin{tabular}{lcc}
        \toprule
        \textbf{Configuration}
        & \textbf{340M}
        & \textbf{1.3B} \\
        \midrule
        Number of layers
        & $24$
        & $24$ \\
        Training corpus
        & FineWeb-Edu
        & FineWeb-Edu \\
        Training tokens
        & $20$B
        & $100$B \\
        Global batch size
        & $50$K tokens
        & $1$M tokens \\
        Optimizer
        & AdamW
        & AdamW \\
        Learning-rate schedule
        & Cosine
        & Cosine \\
        Implementation
        & Flash Linear Attention
        & Flash Linear Attention \\
        \bottomrule
    \end{tabular}
\end{table}

The shared architecture depth and pretraining recipe make this suite a controlled testbed for comparing MA dynamics across linear attention architectures, hybridization ratios, and model scales.

\subsection{Attention-Sink-Guided Tracking and Alignment}
\label{sec:identification}
\begingroup

\begin{figure*}[t]
    \centering

    \begin{subfigure}[t]{0.49\textwidth}
        \centering
        \includegraphics[width=\linewidth]{
            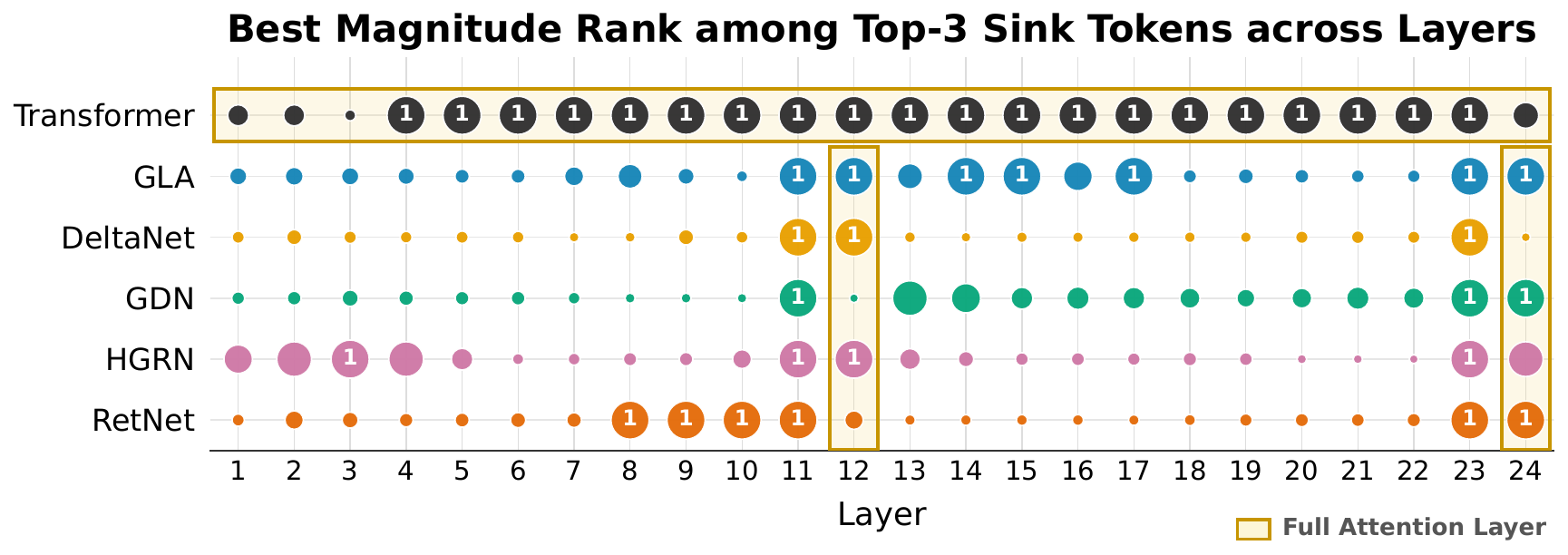
        }
        \vspace{-5mm}
        \caption{
            Best magnitude rank among the top-3 consensus sink tokens across layers.
            Larger bubbles indicate better magnitude--sink alignment.
        }
        \label{fig:top3-sink-magnitude-rank-main}
    \end{subfigure}
    \hfill
    \begin{subfigure}[t]{0.49\textwidth}
        \centering
        \includegraphics[width=\linewidth]{
            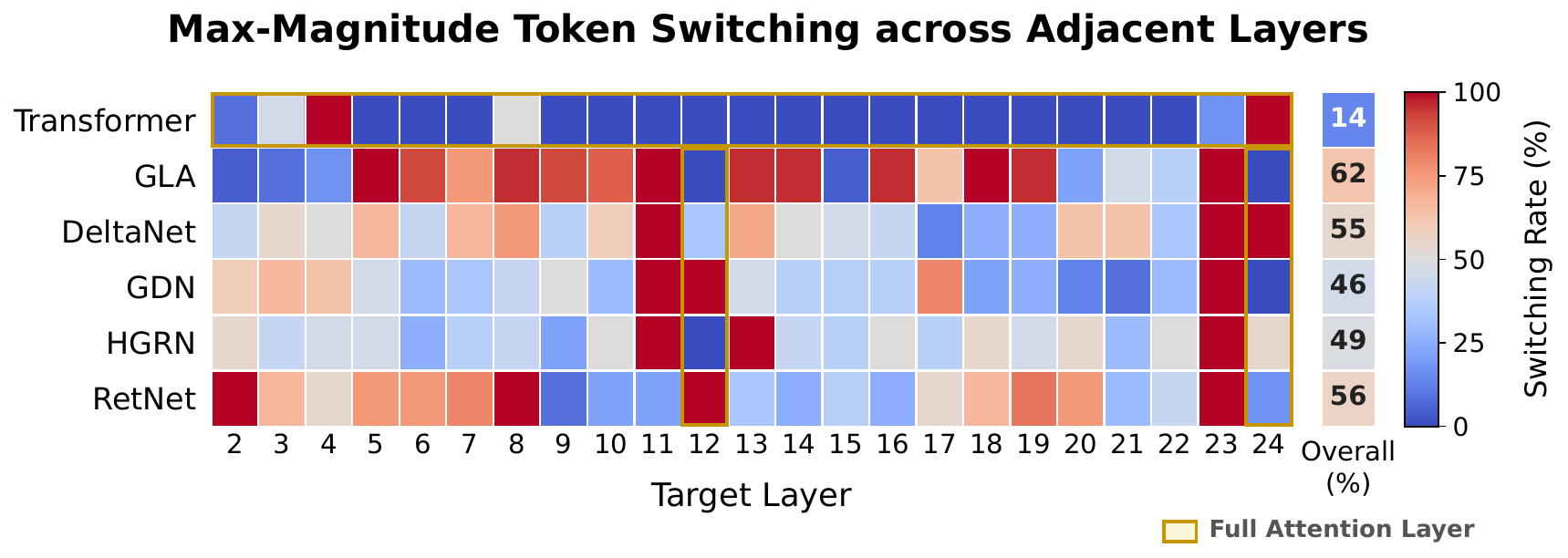
        }
        \vspace{-5mm}
        \caption{
            Mean frequency of changes in the maximally activated token between adjacent layers.
        }
        \label{fig:max-token-switching-main}
    \end{subfigure}

    \vspace{-2mm}
    \caption{
        \textbf{Magnitude--sink correspondence in 1.3B HLA LLMs at a $12{:}1$ hybridization ratio.}
        Compared with the full attention Transformer, HLA models exhibit less consistent alignment between attention sinks and magnitude-ranked tokens (left), together with more frequent switching of maximally activated tokens across layers (right).
        Gold boxes mark full attention layers.
    }
    \label{fig:magnitude-sink-correspondence}
\end{figure*}

\paragraph{Magnitude Ranking and Sink Alignment.}
Prior work detects MAs by selecting the hidden state entries with the largest absolute magnitudes at each layer~\citep{sun2024massive}.
Subsequent studies show that these activations are not merely numerical anomalies but a key component of a systematic outlier mechanism in Transformers; tokens carrying MAs frequently attract a disproportionate share of attention mass and coincide with attention sinks~\citep{an2025systematic,sun2026spike,su2025kvsink,su2026attention}.
In full attention LLMs, this alignment between numerical magnitude and functional attention behavior makes layerwise magnitude ranking a useful proxy for tracing MAs across depth.
Under hybridization, however, attention sink tokens exhibit lower and less stable magnitude rankings than in the full attention baseline, while the maximally activated token switches more frequently between adjacent layers (Figure~\ref{fig:magnitude-sink-correspondence}).
We therefore complement magnitude ranking with attention behavior, using consensus attention sinks as stable, attention-derived token anchors.

\paragraph{Consensus Attention Sinks as Token Anchors.}
Equation~\ref{eq:main-consensus-sink} in \S\ref{sec:experimental-setup} defines the input-specific consensus sink using attention received from later queries, averaged across full attention layers and heads and normalized by the number of eligible queries.
The diagnostic in Figure~\ref{fig:top3-sink-magnitude-rank-main} considers the three highest-scoring tokens; the quantitative alignment and retention analyses use the single dominant sink $t_x^\star$.
Its activation trajectory $m_{x,t_x^\star}^{(\ell)}$ preserves token identity, while the maximally activated feature may change with depth.
Attention selects the anchor independently of activation magnitude; fixed-coordinate lifecycle analysis is conducted separately in Appendix~\ref{app:detailed-pas-lifecycle}.
\paragraph{Sink Token Categories.}
Across the examined models and inputs, sink positions include the first token, punctuation and delimiters (newlines, periods, commas), and semantically light words such as \textit{the}, \textit{is}, \textit{of}, and \textit{in}, consistent with earlier MA observations~\citep{sun2024massive}. The first token is the most prevalent; the main qualitative analysis uses it, while quantitative alignment uses the input-specific consensus sink.
Sink selection and the MA definition allow any eligible token position. Under causal masking, a fixed first token's hidden states are independent of subsequent text, although its received attention can depend on later queries. First-token plots illustrate layerwise organization; other positions appear in Appendix~\ref{app:map-token-dynamics}.

Table~\ref{tab:pas-alignment-domains} reports the domain-specific results underlying Table~\ref{tab:pas-alignment} for both parameter scales.
Each domain contains 100 inputs; Overall denotes their macro-average.
Token/layer controls and uncertainty analyses are provided in Appendix~\ref{app:map-statistical-controls}.
\begin{table*}[t]
    \centering
    \caption{
        \textbf{Sink--spike alignment rates in M-A-P HLA models (\%).}
        Results are reported for different linear attention backbones under the $12{:}1$ hybridization ratio.
        Each entry denotes $1.3\mathrm{B}/340\mathrm{M}$, and Overall is the macro-average across the five evaluation domains.
    }
    \label{tab:pas-alignment-domains}
    \small
    \setlength{\tabcolsep}{5pt}
    \resizebox{\textwidth}{!}{%
    \begin{tabular}{@{}lcccccc@{}}
        \toprule
        \shortstack{\textbf{Linear Attention}}
        & \textbf{WikiText}
        & \shortstack{\textbf{Scientific Papers}}
        & \textbf{GSM8K}
        & \textbf{CodeSearchNet}
        & \textbf{FLORES}
        & \textbf{Overall} \\
        \midrule

        RetNet
        & 99.5 / 100.0
        & 100.0 / 100.0
        & 100.0 / 100.0
        & 100.0 / 100.0
        & 100.0 / 100.0
        & \textbf{99.9 / 100.0} \\

        HGRN
        & 100.0 / 100.0
        & 100.0 / 100.0
        & 100.0 / 100.0
        & 100.0 / 100.0
        & 100.0 / 100.0
        & \textbf{100.0 / 100.0} \\

        GLA
        & 100.0 / 99.5
        & 100.0 / 100.0
        & 100.0 / 100.0
        & 100.0 / 100.0
        & 100.0 / 98.5
        & \textbf{100.0 / 99.6} \\

        DeltaNet
        & 100.0 / 100.0
        & 100.0 / 98.0
        & 100.0 / 100.0
        & 100.0 / 99.5
        & 100.0 / 99.5
        & \textbf{100.0 / 99.4} \\

        GDN
        & 100.0 / 100.0
        & 100.0 / 100.0
        & 100.0 / 100.0
        & 100.0 / 100.0
        & 100.0 / 100.0
        & \textbf{100.0 / 100.0} \\

        \bottomrule
    \end{tabular}%
    }
\end{table*}
\endgroup

\subsection{MA Dynamics across Token Positions}
\label{app:map-token-dynamics}
To complement the first-token analysis in the main text, we examine all eight token positions in the running example, \textit{``Summer is warm. Winter is cold.''}
For each token, we trace its maximum absolute hidden-state activation across depth in GDN models under pure linear attention, the $24{:}1$, $12{:}1$, $6{:}1$, and $3{:}1$ hybrid configurations, and full attention.
The resulting trajectories are presented in Figure~\ref{fig:gdn-token-dynamics}.
Red dashed lines mark layers immediately preceding full attention, while green dotted lines mark full attention layers.
Pronounced MAs concentrate at attention-sink positions, particularly the initial token, ``Summer,'' and the first period.
These tokens exhibit the same ratio-dependent organization identified in the main analysis: isolated PAS arise under sparse full attention, become progressively connected through ISP as full attention becomes denser, and ultimately approach the stable MA morphology of the full attention model.
By contrast, tokens that do not serve as attention sinks exhibit neither pronounced PAS nor ISP.
For this running example, these results show that PAS and ISP emerge selectively at sink-associated positions, supporting the use of the first token as a representative anchor in the main analysis.

\subsection{MA Dynamics across Input Domains}
\label{app:map-domain-dynamics}
We visualize PAS in representative inputs from different domains.
For the running example and representative inputs from WikiText-103, Scientific Papers, GSM8K, CodeSearchNet, and FLORES-200~\citep{merity2016pointer,cohan2018discourse,cobbe2021training,husain2019codesearchnet,costa2022no}, we trace the maximum absolute hidden-state activation of the first token across five HLA architectures at the 1.3B scale and a fixed $12{:}1$ hybridization ratio.
A full attention Transformer is included as a reference.
The resulting trajectories are presented in Figure~\ref{fig:map-cross-domain-dynamics}.

These representative first-token trajectories show qualitatively similar layerwise organization, using the anchor described in Appendix~\ref{sec:identification}.
Across all five linear attention backbones, sink-associated activations attain pronounced maxima immediately before full attention layers.
They complement the input-specific consensus-sink statistics in Table~\ref{tab:pas-alignment}, computed over 100 inputs per domain, which quantify recurrence across domains.

\subsection{MA Dynamics across Linear Attention Backbones and Hybridization Ratios}
\label{app:map-ma-dynamics}

We complement the cross-backbone analysis with within-family comparisons that characterize how MA dynamics vary with the hybridization ratio.
Figure~\ref{fig:ratio-morphologies} presents results for GDN, DeltaNet, GLA, HGRN, and RetNet at both the 1.3B and 340M parameter scales.
For each backbone and scale, we trace the maximum absolute hidden-state activation of the first token, ``Summer,'' across model depth.
The figure compares configurations containing 0, 1, 2, 4, 8, or 24 full attention layers among the 24 sequence-mixing layers, corresponding to pure linear attention, the $24{:}1$, $12{:}1$, $6{:}1$, and $3{:}1$ hybrid configurations, and a full attention Transformer, respectively.
Red dashed lines mark layers immediately preceding full attention, while green dotted lines mark full attention layers.

\subsection{Statistical Controls and Uncertainty}
\label{app:map-statistical-controls}
We complement the main M-A-P analysis with token and layer controls for sink--spike alignment and bootstrap uncertainty for both alignment and ISR.
All analyses use the same 500 inputs as the main experiments, with 100 inputs from each of the five evaluation domains.
Confidence intervals are obtained from 10,000 stratified bootstrap resamples, sampling inputs with replacement within each domain before recomputing the macro-average.

\paragraph{Controls for Sink--Spike Alignment.}
For each input, we aggregate received attention across all full attention layers and select the consensus sink as in Equation~\ref{eq:main-consensus-sink}.
We compare it with two token controls: the first token and the average over eligible tokens outside the top three consensus sinks.
We also include a random-layer baseline, which selects uniformly among the 11 layers in each preceding linear attention block and therefore has an expected alignment rate of $1/11=9.1\%$.

To measure prominence beyond the location of the blockwise maximum, we define the continuous \emph{peak excess}
\begin{equation}
    E_{x,t,f}
    =
    \log_2
    \frac{m_{x,t}^{(f-1)}}
    {\max_{\ell\in\mathcal{B}_f\setminus\{f-1\}}
     m_{x,t}^{(\ell)}}.
\label{eq:pas-peak-excess}
\end{equation}
A positive value indicates that the pre-attention activation exceeds every earlier activation in the block; one unit corresponds to a twofold increase over the strongest earlier layer.

\begin{table*}[t]
    \centering
    \caption{
        \textbf{Token controls for PAS localization and prominence in M-A-P HLA models.}
        Each entry reports $1.3\mathrm{B}/340\mathrm{M}$ under the $12{:}1$ hybridization ratio, macro-averaged over five domains with 100 inputs each.
        Alignment is reported in percent; peak excess is measured in log$_2$ units using Equation~\ref{eq:pas-peak-excess}.
        Non-sink denotes the average over tokens outside the top three consensus sinks.
    }
    \label{tab:pas-alignment-controls}
    \small
    \setlength{\tabcolsep}{4.2pt}
    \resizebox{\textwidth}{!}{%
    \begin{tabular}{@{}lcccccc@{}}
        \toprule
        & \multicolumn{3}{c}{\textbf{Alignment (\%)}}
        & \multicolumn{3}{c}{\textbf{Peak Excess (log$_2$)}} \\
        \cmidrule(lr){2-4}\cmidrule(lr){5-7}
        \textbf{Linear Attention}
        & \textbf{Sink} & \textbf{Non-sink} & \textbf{First Token}
        & \textbf{Sink} & \textbf{Non-sink} & \textbf{First Token} \\
        \midrule
        RetNet   & 99.9 / 100.0 & 57.1 / 71.5 & 99.9 / 100.0 & 2.70 / 2.70 & 0.07 / 0.18  & 2.69 / 2.62 \\
        HGRN     & 100.0 / 100.0 & 93.3 / 84.3 & 100.0 / 100.0 & 4.32 / 4.83 & 0.52 / 0.41  & 4.32 / 4.83 \\
        GLA      & 100.0 / 99.6 & 75.9 / 69.4 & 100.0 / 99.7 & 3.70 / 3.18 & 0.21 / 0.17  & 2.34 / 2.81 \\
        DeltaNet & 100.0 / 99.4 & 57.6 / 40.8 & 99.8 / 99.9 & 2.86 / 0.89 & 0.04 / $-0.08$ & 2.85 / 1.01 \\
        GDN      & 100.0 / 100.0 & 81.2 / 93.5 & 100.0 / 100.0 & 3.04 / 2.01 & 0.28 / 0.54  & 3.03 / 2.01 \\
        \bottomrule
    \end{tabular}%
    }
\end{table*}

Table~\ref{tab:pas-alignment-controls} shows that consensus sinks are more strongly localized and substantially more prominent than the non-sink controls.
Across all ten architecture--scale pairs, the sink-minus-non-sink alignment gap ranges from 6.5 to 58.6 percentage points, while the peak-excess gap ranges from 0.97 to 4.42 log$_2$ units; every paired 95\% bootstrap confidence interval excludes zero.
The first-token control is often comparable to the consensus sink, reflecting that the initial position is itself the dominant sink for most inputs.
Thus, the controls establish that PAS concentrates at sink-associated positions relative to generic tokens.

\paragraph{Uncertainty of the ISR Trend.}
We assess the monotonic ISR trend in Table~\ref{tab:isp-retention} with a paired stratified bootstrap, matching each input across adjacent hybridization ratios.
All 20 architecture--scale comparisons increase with denser full attention.
The $12{:}1\rightarrow6{:}1$ increases range from 5.5 to 44.5 percentage points, and the $6{:}1\rightarrow3{:}1$ increases range from 17.0 to 52.3 points; all paired 95\% confidence intervals exclude zero.
These paired comparisons support higher mean inter-spike retention with denser full attention across the evaluated architectures and scales.

\paragraph{Absolute Inter-Spike Activation.}
Because ISR normalizes each inter-spike activation by its adjacent PAS, we additionally report the unnormalized mean activation magnitude
\begin{equation}
    A_{\mathrm{ISP}}(\mathcal{D})
    =
    \frac{1}{|\mathcal{D}|}
    \sum_{x\in\mathcal{D}}
    \frac{1}{K-1}
    \sum_{i=1}^{K-1}
    \frac{1}{|\mathcal{J}_i|}
    \sum_{\ell\in\mathcal{J}_i}
    m_{x,t_x^\star}^{(\ell)}.
\label{eq:absolute-inter-spike-activation}
\end{equation}
\begin{table}[t]
    \centering
    \caption{
        \textbf{Mean absolute inter-spike activation in M-A-P HLA models.}
        Each entry denotes $1.3\mathrm{B}/340\mathrm{M}$ and is macro-averaged across five domains with 100 inputs per domain.
        Values are computed using Equation~\ref{eq:absolute-inter-spike-activation}; comparisons are meaningful within an architecture and scale because activation scales differ across checkpoints.
    }
    \label{tab:absolute-inter-spike-activation}
    \small
    \setlength{\tabcolsep}{5pt}
    \begin{tabular}{@{}lccc@{}}
        \toprule
        \textbf{Linear Attention} & $\mathbf{12{:}1}$ & $\mathbf{6{:}1}$ & $\mathbf{3{:}1}$ \\
        \midrule
        RetNet   & 25.2 / 15.9 & 102.7 / 23.2 & 236.8 / 89.1 \\
        HGRN     & 14.7 /  8.5 & 132.8 / 70.2 & 368.9 / 165.5 \\
        GLA      & 37.8 / 13.1 &  71.4 / 20.9 & 360.5 /  89.7 \\
        DeltaNet & 42.4 / 35.4 &  85.2 / 57.1 & 405.6 / 290.4 \\
        GDN      & 33.4 / 19.7 &  34.0 / 16.3 & 304.4 / 122.8 \\
        \bottomrule
    \end{tabular}
\end{table}

This companion statistic measures absolute plateau amplitude, rather than retention relative to the neighboring spikes.
Table~\ref{tab:absolute-inter-spike-activation} shows that denser full attention increases mean absolute inter-spike activation in 19 of the 20 adjacent-ratio comparisons; the only exception is the $12{:}1\rightarrow6{:}1$ transition for 340M GDN.
All ten $6{:}1\rightarrow3{:}1$ comparisons increase substantially.
Thus, increasing ISR generally accompanies stronger absolute activation, but the metrics differ: ISR measures persistence relative to adjacent PAS, whereas $A_{\mathrm{ISP}}$ measures absolute amplitude.

\section{MA Dynamics in Large-Scale Pretrained Hybrid Models}
\label{app:large-scale-hybrid-models}

\subsection{Model Families and Evaluated Checkpoints}
\label{app:large-scale-model-suite}
To test whether the architecture-aligned MA organization extends beyond the controlled M-A-P suite, we evaluate 12 publicly available checkpoints from four large-scale hybrid model families.
As summarized in Table~\ref{tab:large-scale-hybrid-models}, these models span 1.2B to 397B total parameters and cover linear attention and state-space mixers, periodic and nonuniform hybridization patterns, and both base and instruction-tuned checkpoints.

\begin{table*}[t]
    \centering
    \small
    \setlength{\tabcolsep}{4pt}
    \renewcommand{\arraystretch}{1.12}
    \caption{
        \textbf{Large-scale pretrained hybrid models used for cross-model evaluation.}
        Mixer counts exclude MLP-only layers where applicable.
        Parameter counts denote total parameters; A$x$B indicates approximately $x$B activated parameters.
    }
    \label{tab:large-scale-hybrid-models}
    \begin{tabularx}{\textwidth}{
        @{}
        >{\raggedright\arraybackslash}p{0.13\textwidth}
        >{\raggedright\arraybackslash}p{0.16\textwidth}
        >{\raggedright\arraybackslash}X
        >{\raggedright\arraybackslash}p{0.27\textwidth}
        @{}
    }
        \toprule
        \textbf{Model Family}
        & \textbf{Linear-Time Mixer}
        & \textbf{Hybridization Pattern}
        & \textbf{Evaluated Checkpoints} \\
        \midrule

        Kimi Linear
        & Kimi Delta Attention
        & 20 KDA and 7 global MLA layers; $\rho=27/7$
        & 48B-A3B Base and Instruct \\

        Qwen3.5
        & Gated DeltaNet
        & Three GDN layers followed by one gated full attention layer
        & 35B-A3B Base and Instruct; 122B-A10B; 397B-A17B \\

        Nemotron-H
        & Mamba-2
        & Mamba-2 and self-attention interleaved with separate MLP layers
        & 8B, 47B, and 56B Base \\

        Zamba2
        & Mamba-2
        & Mamba-2 backbone with periodically inserted shared Transformer blocks
        & 1.2B, 2.7B, and 7B \\

        \bottomrule
    \end{tabularx}
\end{table*}

\paragraph{Kimi Linear.}
Kimi Linear replaces most full attention layers with Kimi Delta Attention (KDA), a linear attention mechanism based on the gated delta rule.
Its 27-layer sequence-mixing stack contains 20 KDA layers and seven global Multi-Head Latent Attention (MLA) layers, with $\rho=27/7$ by the all-mixer definition in \S\ref{sec:preliminaries}. The approximately $3{:}1$ KDA-to-MLA count is a different ratio; the schedule includes an additional MLA layer at the end.
Matched 48B-A3B Base and Instruct checkpoints test PAS--ISP stability across post-training~\citep{team2025kimi}.

\paragraph{Qwen3.5.}
Qwen3.5 uses three GDN layers followed by one full softmax attention layer. This is a $3{:}1$ linear-to-full pattern, but $\rho=4$ under our all-mixer-to-full definition, not the M-A-P $3{:}1$ configuration.
The 35B-A3B, 122B-A10B, and 397B-A17B checkpoints contain 40, 48, and 60 sequence-mixing layers, including 10, 12, and 15 full attention layers, respectively.
These full attention layers employ element-wise output gating, making Qwen3.5 particularly relevant to our controlled gating analysis.
We evaluate 35B-A3B (Base and Instruct), 122B-A10B, and 397B-A17B checkpoints~\citep{yang2025qwen3}.

\paragraph{Nemotron-H.}
Nemotron-H arranges Mamba-2, self-attention, and MLP modules as separate layers.
The 8B and 56B configurations contain 24/4 and 54/10 Mamba-2/self-attention layers, respectively.
The compressed 47B checkpoint retains a nonuniform schedule of 45 Mamba-2 and five self-attention layers.
Nemotron-H therefore complements the fixed periodic designs above with sparser and partly irregular full attention placement~\citep{blakeman2025nemotron}.

\paragraph{Zamba2.}
Zamba2 interleaves a Mamba-2 backbone with hybrid blocks containing shared Transformer attention.
The 1.2B, 2.7B, and 7B configurations contain 38, 54, and 81 layer positions, including six, nine, and thirteen hybrid blocks, respectively.
These blocks occur approximately once every six layers.
The 1.2B model repeatedly invokes one shared Transformer block, whereas the 2.7B and 7B models alternate between two shared blocks; layer-specific adapters specialize their repeated use across depth~\citep{glorioso2024zamba2}.

Together, these families cover linear attention and state-space hybrids, periodic and nonuniform full attention placement, shared and unshared attention modules, multiple parameter scales, and distinct post-training stages.
This diversity allows us to test whether PAS and ISP reflect a general property of layer-interleaved hybrid architectures rather than a particular mixer or model family.

\subsection{MA Dynamics across Models and Input Domains}
\label{app:large-scale-ma-dynamics}
Figure~\ref{fig:large-scale-hybrid-ma-dynamics} presents the complete MA tracing results for the checkpoints summarized in Table~\ref{tab:large-scale-hybrid-models}.
For each checkpoint, six trajectories cover the running example, \textit{``Summer is warm. Winter is cold.''}, and representative inputs from five domains: general prose from WikiText-103, scientific writing from Scientific Papers, mathematical reasoning from GSM8K, Python code from CodeSearchNet, and multilingual text from FLORES-200~\citep{merity2016pointer,cohan2018discourse,cobbe2021training,husain2019codesearchnet,costa2022no}.
Each trajectory traces the maximum absolute hidden-state activation of the first token across model depth.
Red dashed lines mark layers immediately preceding full attention, while green dotted lines mark full attention layers.

\paragraph{Post-Training and Model Scale.}
The matched base and instruction-tuned checkpoints of Kimi Linear and Qwen3.5 exhibit closely aligned PAS locations and ISP spans.
Their activation magnitudes differ, but the correspondence between PAS--ISP organization and full attention placement remains largely intact in both matched pairs.
Across Qwen3.5, Nemotron-H, and Zamba2 checkpoints of different sizes, PAS and ISP remain aligned with full attention placement even as their magnitude and prominence vary.
Their magnitude does not vary monotonically with parameter count, whereas their correspondence with full attention placement persists.

\paragraph{Sequence Mixers and Input Domains.}
PAS and ISP recur in both linear attention hybrids, represented by Kimi Linear and Qwen3.5, and state-space hybrids, represented by Nemotron-H and Zamba2, despite their distinct sequence-mixing mechanisms.
Across both classes, PAS occur immediately before full attention layers, while ISP extend across portions of the intervening linear-time sequence-mixing blocks.
This cross-backbone consistency supports an association with layer-interleaved hybridization rather than a particular recurrent state-transition or update rule.
Within each checkpoint, the six representative inputs show closely aligned PAS locations and ISP spans, with varying magnitudes.
These plots provide representative first-token trajectories across checkpoints, with the interpretation of this anchor specified in Appendix~\ref{sec:identification}.
The broader input-specific sink statistics and token controls are reported for M-A-P in Appendix~\ref{app:map-statistical-controls}.

\begingroup
\color{revisioncolor}
\subsection{\rev{Population Statistics and Token Comparisons}}
\label{app:large-population-statistics}
\rev{We quantify the original SSA and clipped ISR for all twelve checkpoints over five natural domains. The natural-domain means are equally weighted. For each input, S is the Eq.~(5) attention sink, while N8 and N64 are two distinct non-sink positions outside the top three positions by attention score, chosen nearest zero-based positions 8 and 64. The final source position is excluded because it has no future queries. Across checkpoints, this gives 6,000 model--input evaluations and 18,000 paired token observations.}

\rev{Brackets report 95\% paired source-cluster bootstrap intervals with 10,000 resamples. Single-source domain strata are held fixed, so natural-domain intervals are conditional on those observed strata. Repeated leading-prefix trajectories are distinguished from independent content-dependent trajectories. Original SSA and ISR are supplemented by descriptive peak excess (the base-two logarithm of the ratio of the pre-FA peak to the largest earlier value in the preceding linear block) and absolute inter-spike height, averaged over the Eq.~(7) interiors. These two companions were introduced after examining ISR, to distinguish low flat curves from large activation plateaus. We measure activations in the native residual stream.}
\rev{Sink-token absolute plateau height exceeds both non-sink controls in all twelve checkpoints. Non-sink ISR can be higher because it measures retention relative to those tokens' smaller endpoint values. Peak prominence varies across architectures; the combination of alignment, retention, peak excess and absolute height describes this variation explicitly.}
\begingroup
\color{revisioncolor}
\begin{table}[H]
\centering
\small
\setlength{\tabcolsep}{3pt}
\caption{\textbf{PAS alignment and peak strength.} N8/N64: two non-sink controls. Five natural-domain means, equally weighted. Brackets give 95\% conditional source-cluster intervals.}
\label{tab:large-pas-population}
\revisionfit{%
\begin{tabular}{@{}lrrr@{}}
\toprule
\multicolumn{4}{@{}l@{}}{\textbf{SSA (\%)}}\\
\textbf{Model} & \textbf{Sink} & \textbf{Non-sink N8} & \textbf{Non-sink N64}\\
\midrule
Kimi Linear 48B-A3B Base & 100.00\% [100.00, 100.00] & 95.63\% [95.15, 96.10] & 89.80\% [88.78, 90.76]\\
Kimi Linear 48B-A3B Instruct & 100.00\% [100.00, 100.00] & 96.63\% [96.23, 97.05] & 93.34\% [92.56, 94.09]\\
Qwen3.5-35B-A3B Base & 92.30\% [91.93, 92.63] & 85.84\% [85.18, 86.54] & 85.06\% [84.00, 86.29]\\
Qwen3.5-35B-A3B Instruct & 92.14\% [91.78, 92.56] & 86.96\% [85.98, 87.70] & 86.86\% [85.90, 87.76]\\
Qwen3.5-122B-A10B & 90.33\% [89.96, 90.70] & 69.55\% [68.69, 70.30] & 67.43\% [66.47, 68.23]\\
Qwen3.5-397B-A17B & 99.21\% [99.02, 99.40] & 67.83\% [67.19, 68.34] & 65.99\% [65.48, 66.51]\\
Nemotron-H-8B & 75.00\% [75.00, 75.00] & 83.90\% [82.27, 85.32] & 63.80\% [61.22, 66.08]\\
Nemotron-H-47B & 100.00\% [100.00, 100.00] & 83.76\% [82.35, 85.06] & 69.28\% [66.68, 71.36]\\
Nemotron-H-56B & 97.62\% [97.38, 98.24] & 90.22\% [89.58, 90.92] & 74.18\% [72.82, 75.39]\\
Zamba2-1.2B & 100.00\% [100.00, 100.00] & 90.53\% [89.18, 91.67] & 82.57\% [81.29, 83.93]\\
Zamba2-2.7B & 100.00\% [100.00, 100.00] & 92.20\% [91.65, 92.79] & 88.51\% [87.80, 89.20]\\
Zamba2-7B & 100.00\% [100.00, 100.00] & 80.77\% [79.63, 81.83] & 72.02\% [70.84, 73.28]\\
\midrule
\multicolumn{4}{@{}l@{}}{\textbf{Peak excess}}\\
\textbf{Model} & \textbf{Sink} & \textbf{Non-sink N8} & \textbf{Non-sink N64}\\
\midrule
Kimi Linear 48B-A3B Base & 1.7355 [1.7320, 1.7392] & 0.9221 [0.9028, 0.9434] & 0.6612 [0.6389, 0.6844]\\
Kimi Linear 48B-A3B Instruct & 1.7616 [1.7575, 1.7657] & 0.9559 [0.9360, 0.9765] & 0.7144 [0.6933, 0.7353]\\
Qwen3.5-35B-A3B Base & 1.1817 [1.1757, 1.1881] & 0.4811 [0.4684, 0.4947] & 0.3871 [0.3742, 0.4001]\\
Qwen3.5-35B-A3B Instruct & 1.1958 [1.1884, 1.2032] & 0.4775 [0.4633, 0.4908] & 0.3964 [0.3829, 0.4083]\\
Qwen3.5-122B-A10B & 0.6852 [0.6835, 0.6870] & 0.2204 [0.2106, 0.2287] & 0.1680 [0.1601, 0.1759]\\
Qwen3.5-397B-A17B & 0.9694 [0.9661, 0.9724] & 0.2842 [0.2756, 0.2911] & 0.2346 [0.2275, 0.2408]\\
Nemotron-H-8B & 2.6039 [2.5984, 2.6094] & 0.3233 [0.3090, 0.3345] & 0.1274 [0.1089, 0.1488]\\
Nemotron-H-47B & 0.2610 [0.2607, 0.2613] & 0.2942 [0.2750, 0.3074] & 0.1910 [0.1723, 0.2134]\\
Nemotron-H-56B & 1.1018 [1.0991, 1.1049] & 0.4091 [0.3980, 0.4192] & 0.2253 [0.2116, 0.2430]\\
Zamba2-1.2B & 2.0023 [2.0023, 2.0023] & 0.6460 [0.6238, 0.6683] & 0.5155 [0.4913, 0.5446]\\
Zamba2-2.7B & 0.9115 [0.9115, 0.9115] & 0.3358 [0.3296, 0.3417] & 0.3023 [0.2958, 0.3098]\\
Zamba2-7B & 0.5901 [0.5901, 0.5901] & 0.2297 [0.2193, 0.2404] & 0.1764 [0.1640, 0.1908]\\
\bottomrule
\end{tabular}%
}
\end{table}
\clearpage
\begin{table}[H]
\centering
\small
\setlength{\tabcolsep}{3pt}
\caption{\textbf{Inter-spike retention and absolute height.} N8/N64: two non-sink controls. Five natural-domain means, equally weighted. Brackets give 95\% conditional source-cluster intervals.}
\label{tab:large-isp-population}
\revisionfit{%
\begin{tabular}{@{}lrrr@{}}
\toprule
\multicolumn{4}{@{}l@{}}{\textbf{ISR}}\\
\textbf{Model} & \textbf{Sink} & \textbf{Non-sink N8} & \textbf{Non-sink N64}\\
\midrule
Kimi Linear 48B-A3B Base & 0.6864 [0.6861, 0.6866] & 0.8516 [0.8429, 0.8596] & 0.9044 [0.8977, 0.9109]\\
Kimi Linear 48B-A3B Instruct & 0.6997 [0.6995, 0.6999] & 0.8474 [0.8382, 0.8562] & 0.8927 [0.8863, 0.8988]\\
Qwen3.5-35B-A3B Base & 0.7286 [0.7265, 0.7307] & 0.8204 [0.8132, 0.8268] & 0.8563 [0.8506, 0.8617]\\
Qwen3.5-35B-A3B Instruct & 0.7398 [0.7370, 0.7429] & 0.8254 [0.8183, 0.8321] & 0.8592 [0.8537, 0.8651]\\
Qwen3.5-122B-A10B & 0.9020 [0.9018, 0.9021] & 0.9050 [0.9008, 0.9099] & 0.9277 [0.9243, 0.9310]\\
Qwen3.5-397B-A17B & 0.7402 [0.7389, 0.7417] & 0.8844 [0.8812, 0.8882] & 0.9046 [0.9021, 0.9074]\\
Nemotron-H-8B & 0.7248 [0.7247, 0.7250] & 0.9403 [0.9380, 0.9429] & 0.9537 [0.9508, 0.9560]\\
Nemotron-H-47B & 0.8839 [0.8831, 0.8852] & 0.9471 [0.9428, 0.9547] & 0.9489 [0.9404, 0.9532]\\
Nemotron-H-56B & 0.7299 [0.7285, 0.7310] & 0.9229 [0.9187, 0.9290] & 0.9484 [0.9408, 0.9517]\\
Zamba2-1.2B & 0.5697 [0.5697, 0.5697] & 0.9126 [0.9063, 0.9189] & 0.9345 [0.9283, 0.9389]\\
Zamba2-2.7B & 0.8462 [0.8462, 0.8462] & 0.9871 [0.9860, 0.9882] & 0.9877 [0.9862, 0.9890]\\
Zamba2-7B & 0.9526 [0.9526, 0.9526] & 0.9688 [0.9659, 0.9713] & 0.9759 [0.9729, 0.9784]\\
\midrule
\multicolumn{4}{@{}l@{}}{\textbf{Height}}\\
\textbf{Model} & \textbf{Sink} & \textbf{Non-sink N8} & \textbf{Non-sink N64}\\
\midrule
Kimi Linear 48B-A3B Base & 37.60 [37.55, 37.66] & 22.05 [21.50, 22.60] & 18.71 [18.19, 19.19]\\
Kimi Linear 48B-A3B Instruct & 50.59 [50.57, 50.61] & 21.03 [20.51, 21.56] & 18.72 [18.29, 19.11]\\
Qwen3.5-35B-A3B Base & 19.71 [19.61, 19.77] & 1.68 [1.66, 1.70] & 1.88 [1.85, 1.92]\\
Qwen3.5-35B-A3B Instruct & 20.83 [20.73, 20.88] & 1.79 [1.77, 1.82] & 2.06 [2.03, 2.11]\\
Qwen3.5-122B-A10B & 14.29 [14.24, 14.36] & 2.60 [2.57, 2.62] & 2.82 [2.79, 2.86]\\
Qwen3.5-397B-A17B & 23.75 [23.61, 23.89] & 5.69 [5.61, 5.74] & 5.69 [5.63, 5.74]\\
Nemotron-H-8B & 244.59 [244.02, 245.82] & 20.78 [20.45, 21.26] & 16.71 [16.31, 17.05]\\
Nemotron-H-47B & 413.14 [411.99, 414.78] & 78.34 [76.61, 80.77] & 64.05 [62.56, 65.31]\\
Nemotron-H-56B & 342.79 [341.21, 344.31] & 130.57 [126.47, 136.50] & 96.25 [92.50, 99.81]\\
Zamba2-1.2B & 98.20 [98.20, 98.20] & 17.15 [16.72, 17.59] & 15.35 [14.84, 15.90]\\
Zamba2-2.7B & 308.41 [308.41, 308.41] & 73.93 [72.32, 75.56] & 70.70 [68.59, 72.63]\\
Zamba2-7B & 447.40 [447.40, 447.40] & 73.35 [71.87, 74.77] & 66.53 [64.39, 68.52]\\
\bottomrule
\end{tabular}%
}
\end{table}
\endgroup

\clearpage
\endgroup

\section{Controlled Pretraining Studies of PAS and ISP}
\label{app:controlled-training}

\subsection{Experimental Setup and Analysis Overview}
\label{app:controlled-training-setup}
\paragraph{Model Architecture.}
We conduct controlled pretraining experiments using the Flash Linear Attention framework~\citep{yang2024fla}.
All models comprise 24 sequence-mixing layers and use Gated DeltaNet (GDN) as the linear attention backbone.
We consider two parameter scales, 340M and 1.3B.
To instantiate each hybrid configuration, we replace selected GDN mixers with full attention while holding all other architectural components fixed.
Within each scale, this design isolates full attention placement and density from model depth, width, and FFN architecture.

\paragraph{Training Recipe.}
Following the optimization settings of prior GLA and GDN studies~\citep{yang2023gated,yang2025gated}, we train all models from scratch on FineWeb-Edu~\citep{penedo2024fineweb}.
We use AdamW with a sequence length of 4,096, weight decay of 0.01, and gradient clipping at 1.0.
The learning rate is linearly warmed up from \(3\times10^{-5}\) to \(3\times10^{-4}\) and then decayed to \(3\times10^{-5}\) using a cosine schedule.
Unless otherwise specified, the 340M models are trained on 10B tokens with a global batch size of 0.5M tokens and a 0.5B-token warmup, whereas the 1.3B models are trained on 50B tokens with a global batch size of 2M tokens and a 1B-token warmup.

\paragraph{Analysis and Evaluation Overview.}
Our controlled study proceeds in three stages.
First, we characterize the emergence of PAS across full attention placements and model scales (\S\ref{app:controlled-pas}).
Second, we trace ISP formation and consolidation in GDN models at both scales with a \(3{:}1\) hybridization ratio (\S\ref{app:controlled-isp}).
Third, we isolate the role of output gating through two complementary interventions: adding element-wise output gates to full attention layers following gated attention~\citep{qiu2026gated}, and removing the native output gates from GDN layers (\S\ref{app:output-gating}).
Alongside these morphology analyses, we evaluate the trained models on language modeling, commonsense reasoning, and in-context retrieval benchmarks following~\citet{yang2025gated}.
These evaluations contextualize changes in PAS and ISP with downstream model capability.
The complete evaluation protocol and benchmark descriptions are provided in \S\ref{app:controlled-benchmark-setup}.

\subsection{Downstream Evaluation Protocol}
\label{app:controlled-benchmark-setup}
We evaluate all models from our controlled pretraining experiments on language modeling, commonsense reasoning, and in-context retrieval benchmarks.
Unless otherwise specified, we follow the evaluation protocol adopted in the Gated DeltaNet study~\citep{yang2025gated}.

\paragraph{Language Modeling.}
We evaluate next-token prediction on WikiText-103 and report perplexity, with lower values indicating better language modeling performance~\citep{merity2016pointer}.

\paragraph{Commonsense Reasoning.}
We assess zero-shot performance on HellaSwag, PIQA, and ARC-Easy, which test plausible event completion, physical commonsense, and grade-school scientific reasoning, respectively~\citep{zellers2019hellaswag,bisk2020piqa,clark2018think}.
All evaluations are conducted using the Language Model Evaluation Harness~\citep{gao2021framework}.
We report length-normalized accuracy for HellaSwag and accuracy for PIQA and ARC-Easy.

\paragraph{Real-World In-Context Retrieval.}
We evaluate recall-intensive performance on SWDE, FDA, SQuAD, and Natural Questions (NQ).
SWDE assesses structured relation extraction from semi-structured web pages, while FDA assesses key--value retrieval from PDF documents~\citep{lockard2019openceres,arora2024just}.
SQuAD and NQ evaluate question answering over natural-language contexts~\citep{rajpurkar2018know,kwiatkowski2019natural}.
Because our models are pretrained rather than instruction-tuned, we adopt the cloze-completion format of~\citet{arora2024just}, aligning these tasks with the next-token prediction objective.
Inputs are truncated to a maximum context length of 2,048 tokens, and retrieval accuracy is reported for each benchmark.

\paragraph{Synthetic In-Context Retrieval.}
We additionally evaluate the three single-needle NIAH variants from RULER, denoted NIAH-1, NIAH-2, and NIAH-3~\citep{hsieh2024ruler}.
NIAH-1 tests passkey-style retrieval from repeated distractors, NIAH-2 tests numerical-value retrieval from an essay context, and NIAH-3 increases the difficulty by replacing the target value with a UUID.
Each task requires the model to recover the value associated with a target key embedded in a distractor context.
All variants are evaluated at a fixed context length of 4,096 tokens using exact-match accuracy.

\subsection{Training-Time Emergence of PAS}
\label{app:controlled-pas}
\begin{figure*}[t]
    \centering
    \begin{subfigure}[t]{0.49\textwidth}
        \centering
        \includegraphics[width=\linewidth]{
            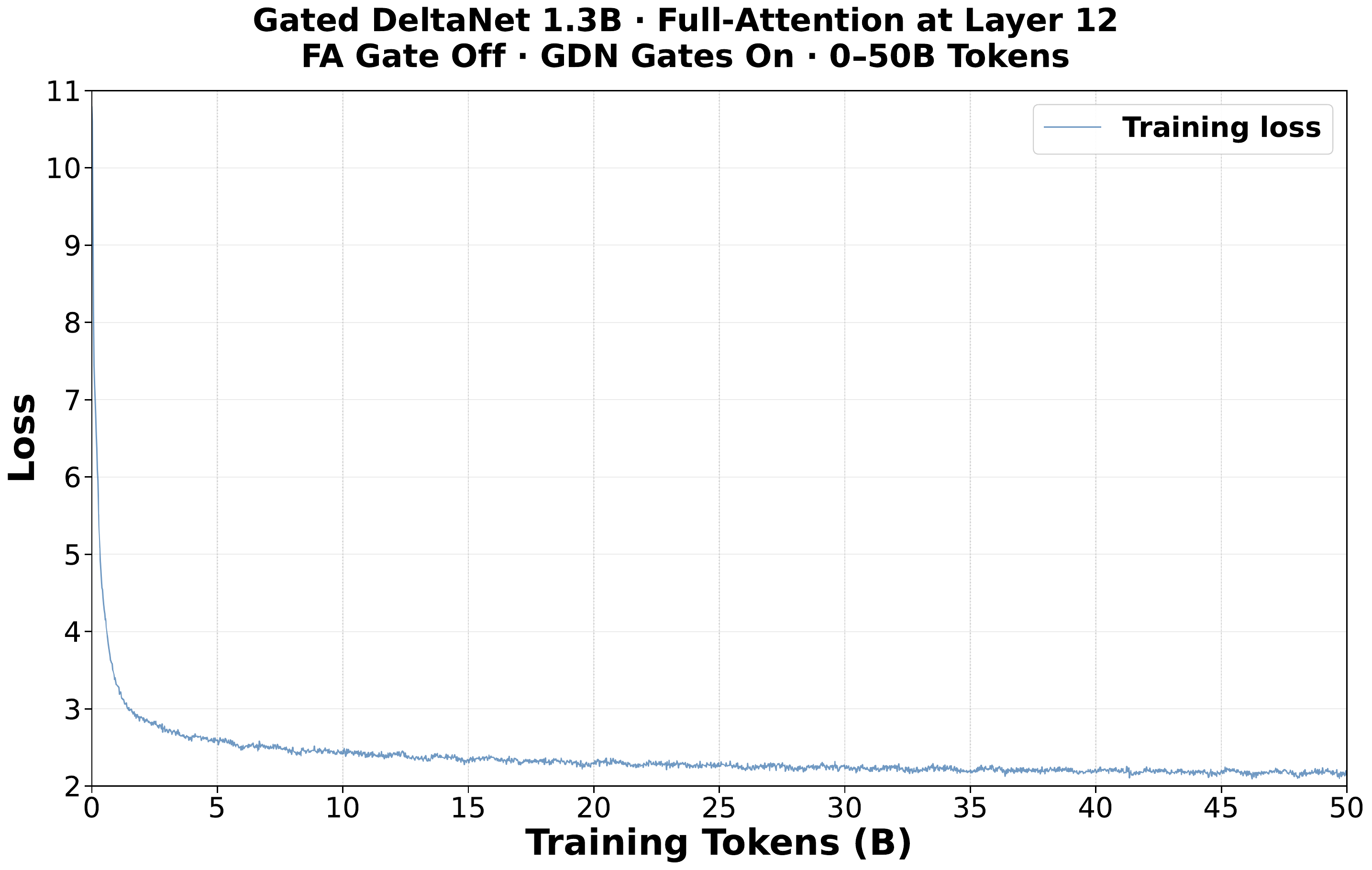
        }
        \caption{
            PAS configuration: full attention only at layer 12.
        }
        \label{fig:controlled-pas-loss}
    \end{subfigure}
    \hfill
    \begin{subfigure}[t]{0.49\textwidth}
        \centering
        \includegraphics[width=\linewidth]{
            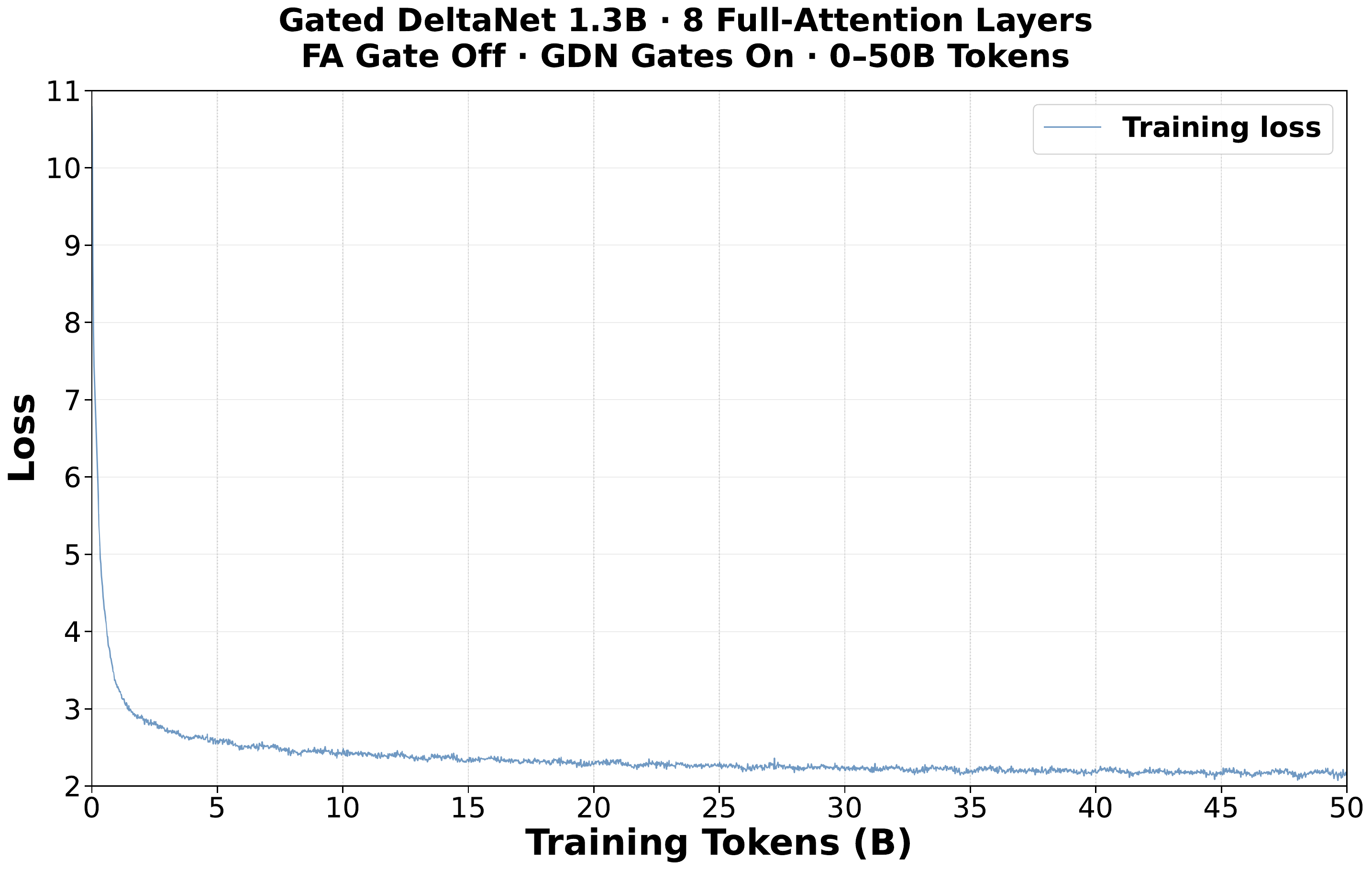
        }
        \caption{
            ISP configuration with a $3{:}1$ hybridization ratio.
        }
        \label{fig:controlled-isp-loss}
    \end{subfigure}
    \caption{
        \textbf{Training loss for representative PAS and ISP configurations.}
        Both curves are obtained from 1.3B GDN models trained on 50B tokens.
        The PAS configuration contains a single full attention layer at layer 12, whereas the ISP configuration interleaves eight full attention layers with 16 GDN layers under a $3{:}1$ hybridization ratio.
    }
    \label{fig:controlled-training-losses}
\end{figure*}

\begin{figure*}[t]
    \centering
    \begin{subfigure}[t]{0.32\textwidth}
        \centering
        \includegraphics[width=\linewidth]{
            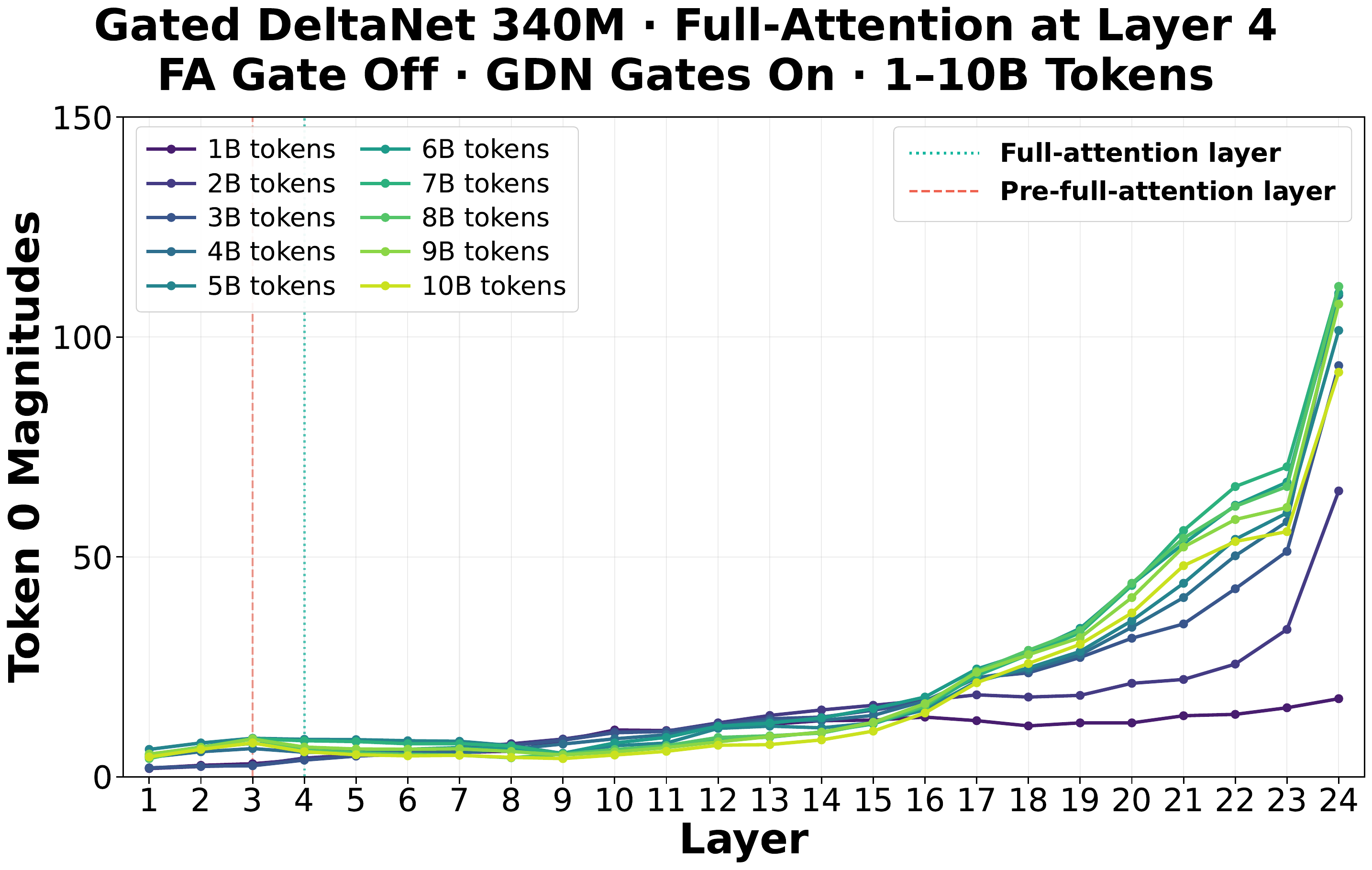
        }
        \caption{
            Early placement: layer 4.
        }
        \label{fig:controlled-pas-placement-early}
    \end{subfigure}
    \hfill
    \begin{subfigure}[t]{0.32\textwidth}
        \centering
        \includegraphics[width=\linewidth]{
            figures/training-time/pas_gdn_340_fa_12_gate_off_la_gate_on.pdf
        }
        \caption{
            Middle placement: layer 12.
        }
        \label{fig:controlled-pas-placement-middle}
    \end{subfigure}
    \hfill
    \begin{subfigure}[t]{0.32\textwidth}
        \centering
        \includegraphics[width=\linewidth]{
            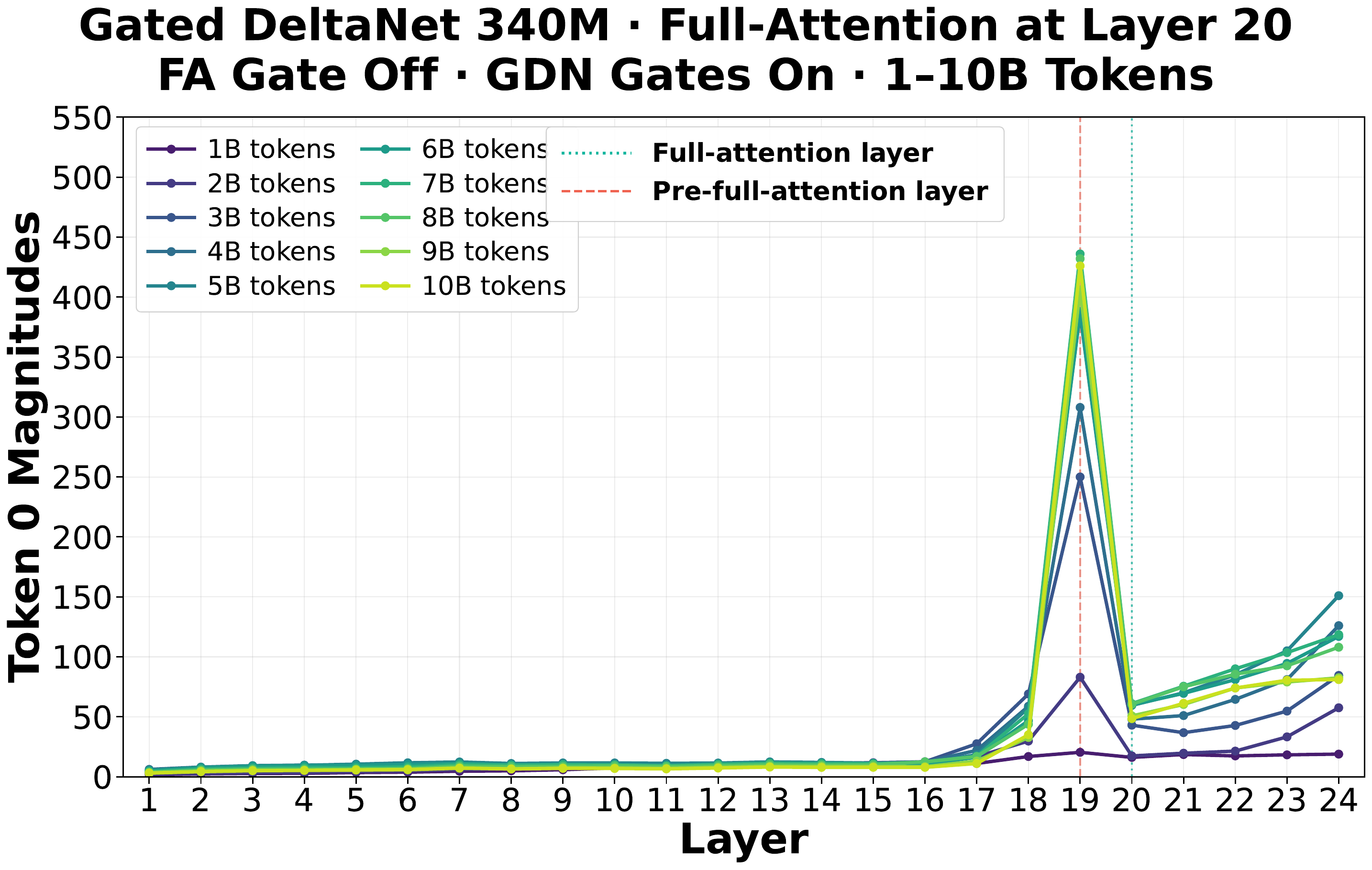
        }
        \caption{
            Late placement: layer 20.
        }
        \label{fig:controlled-pas-placement-late}
    \end{subfigure}
    \caption{
        \textbf{PAS formation across full attention placements.}
        Each panel traces the maximum absolute hidden-state activation of the first token across model depth at successive 1B-token checkpoints for a 340M GDN model containing a single full attention layer.
        PAS remains weak under early placement, becomes pronounced at the middle placement, and reaches its largest magnitude under late placement.
        Full attention placement therefore determines not only where PAS forms but also how prominently it develops.
    }
    \label{fig:controlled-pas-placement}
\end{figure*}

\begin{figure*}[t]
\vspace{-5mm}
    \centering
    \begin{subfigure}[t]{0.49\textwidth}
        \centering
        \includegraphics[width=\linewidth]{
            figures/training-time/pas_gdn_340_fa_12_gate_off_la_gate_on.pdf
        }
        \caption{
            340M model trained on 10B tokens.
        }
        \label{fig:controlled-pas-scale-340m}
    \end{subfigure}
    \hfill
    \begin{subfigure}[t]{0.49\textwidth}
        \centering
        \includegraphics[width=\linewidth]{
            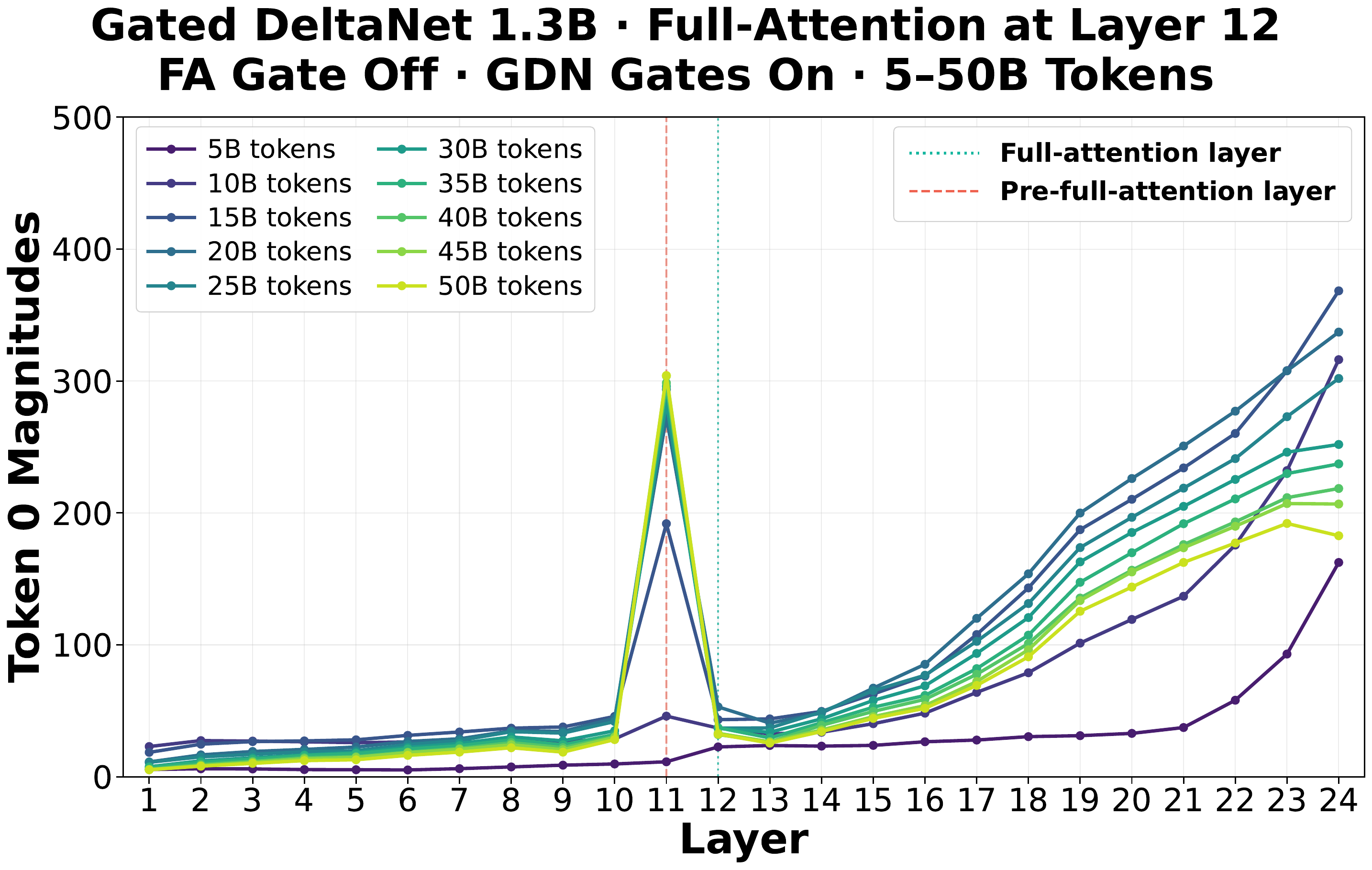
        }
        \caption{
            1.3B model trained on 50B tokens.
        }
        \label{fig:controlled-pas-scale-1p3b}
    \end{subfigure}
    \caption{
        \textbf{PAS formation across model scales.}
        Both models contain 24 sequence-mixing layers with a single full attention layer at layer 12.
        This comparison examines whether the layer-localized PAS organization observed at the 340M scale persists in the 1.3B model.
    }
    \label{fig:controlled-pas-scale-comparison}
\end{figure*}

\begin{figure*}[t]
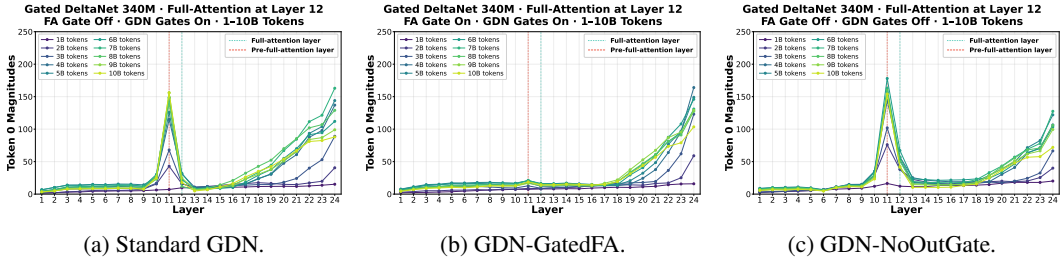

    \centering
    \begin{subfigure}[t]{0.32\textwidth}
        \centering
        \includegraphics[width=\linewidth]{}
        \caption{
            Standard GDN.
        }
        \label{fig:controlled-pas-dynamics}
    \end{subfigure}
    \hfill
    \begin{subfigure}[t]{0.32\textwidth}
        \centering
        \includegraphics[width=\linewidth]{}
        \caption{
            GDN-GatedFA.
        }
        \label{fig:controlled-pas-gatedfa}
    \end{subfigure}
    \hfill
    \begin{subfigure}[t]{0.32\textwidth}
        \centering
        \includegraphics[width=\linewidth]{}
        \caption{
            GDN-NoOutGate.
        }
        \label{fig:controlled-pas-nooutgate}
    \end{subfigure}
    \caption{
        \textbf{PAS dynamics under output-gating interventions.}
        Each panel traces the maximum absolute hidden-state activation of the first token across model depth at successive 1B-token checkpoints for a 340M GDN model with a single full attention layer at layer 12.
        The standard model develops a pronounced PAS.
        Full attention output gating attenuates but does not eliminate PAS; removing GDN output gates moderately amplifies it.
    }
    \label{fig:controlled-pas-gating-dynamics}
\end{figure*}

\begin{table*}[t]
    \centering
    \caption{
        \textbf{PAS and downstream performance under controlled pretraining.}
        FA Layer gives the one-based full attention position among 24 layers.
        Align.\ (sink--spike alignment) and downstream accuracies are percentages; PPL denotes perplexity.
    }
    \label{tab:controlled-pas-results}
    \scriptsize
    \setlength{\tabcolsep}{3.5pt}

    \textbf{(a) PAS alignment, language modeling, and real-world retrieval}
    \par\smallskip

    \resizebox{0.9\textwidth}{!}{%
    \begin{tabular}{lccccccccc}
        \toprule
        \textbf{Variant}
        & \textbf{Scale}
        & \textbf{FA Layer}
        & \textbf{Tokens}
        & \textbf{Align.}
        & \textbf{PPL}
        & \textbf{FDA}
        & \textbf{SWDE}
        & \textbf{NQ}
        & \textbf{SQuAD} \\
        \midrule

        GDN
        & 340M & $4/24$ & 10B
        & 99.53 & 14.00 & 8.02 & 20.21 & 14.95 & 25.30 \\

        GDN
        & 340M & $12/24$ & 10B
        & 100.00 & 13.78 & 60.43 & 40.49 & 19.61 & 32.81 \\

        GDN
        & 340M & $20/24$ & 10B
        & 100.00 & 13.84 & 53.71 & 35.80 & 21.10 & 35.42 \\

        GDN
        & 1.3B & $12/24$ & 50B
        & 100.00 & 9.79 & 66.58 & 45.36 & 23.57 & 36.96 \\

        \midrule

        GDN-NoOutGate
        & 340M & $12/24$ & 10B
        & 100.00 & 13.96 & 49.86 & 41.77 & 19.61 & 34.32 \\

        GDN-GatedFA
        & 340M & $12/24$ & 10B
        & 100.00 & 13.80 & 53.38 & 34.21 & 18.37 & 31.70 \\

        \bottomrule
    \end{tabular}%
    }

    \medskip
    \textbf{(b) Synthetic retrieval and commonsense reasoning}
    \par\smallskip

    \resizebox{\textwidth}{!}{%
    \begin{tabular}{lccccccccc}
        \toprule
        \textbf{Variant}
        & \textbf{Scale}
        & \textbf{FA Layer}
        & \textbf{Tokens}
        & \textbf{NIAH-1}
        & \textbf{NIAH-2}
        & \textbf{NIAH-3}
        & \textbf{HellaSwag}
        & \textbf{PIQA}
        & \textbf{ARC-E} \\
        \midrule

        GDN
        & 340M & $4/24$ & 10B
        & 57.40 & 27.20 & 2.40 & 38.92 & 65.61 & 50.72 \\

        GDN
        & 340M & $12/24$ & 10B
        & 61.80 & 97.00 & 25.80 & 39.61 & 65.56 & 50.17 \\

        GDN
        & 340M & $20/24$ & 10B
        & 90.60 & 87.20 & 69.00 & 39.04 & 64.42 & 52.27 \\

        GDN
        & 1.3B & $12/24$ & 50B
        & 99.40 & 93.60 & 2.60 & 53.37 & 70.40 & 60.06 \\

        \midrule

        GDN-NoOutGate
        & 340M & $12/24$ & 10B
        & 80.40 & 94.40 & 39.40 & 38.87 & 65.78 & 49.33 \\

        GDN-GatedFA
        & 340M & $12/24$ & 10B
        & 41.60 & 16.80 & 30.80 & 39.73 & 65.13 & 50.63 \\

        \bottomrule
    \end{tabular}%
    }
\end{table*}

\paragraph{Experimental Setup.}
We study PAS formation by inserting a single full attention layer into an otherwise homogeneous 24-layer GDN model.
At the 340M scale, we consider three insertion points---layers 4, 12, and 20---representing early, middle, and late full attention placements.
All three models are trained from scratch on 10B tokens using the controlled pretraining recipe described in \S\ref{app:controlled-training-setup}.
To trace the emergence and evolution of PAS, we record the maximum absolute hidden-state activation of the first token across model depth at 1B-token intervals.
At the final checkpoint, we compute the sink--spike alignment rate and evaluate downstream performance.
For layer 12, we also train a 1.3B counterpart on 50B tokens and record trajectories every 5B tokens, testing recurrence at a larger scale and training budget.

\paragraph{Results.}
Figure~\ref{fig:controlled-pas-loss} confirms stable optimization of the 1.3B layer-12 configuration through 50B tokens.
Figure~\ref{fig:controlled-pas-placement} compares PAS formation across early, middle, and late full attention placements, while Figure~\ref{fig:controlled-pas-scale-comparison} provides the corresponding cross-scale comparison.
Table~\ref{tab:controlled-pas-results} reports the final sink--spike alignment rates and downstream performance.
The output-gating variants are analyzed in \S\ref{app:output-gating}.
First, PAS consistently localizes immediately before the inserted full attention layer, but its magnitude depends strongly on placement depth.
As shown in Figure~\ref{fig:controlled-pas-placement}, placement at layer 4 produces only a weak local maximum, whereas layer 12 produces a pronounced spike and layer 20 yields the strongest PAS.
Second, PAS emerges early and recurs at the larger scale.
In the 340M layer-12 configuration, it is already visible after 1B training tokens and becomes progressively sharper and more stable thereafter.
The 1.3B trajectory shows a weak precursor at 5B tokens, a pronounced PAS by 10B, and persistent localization immediately before full attention through 50B tokens.
Both scales achieve 100\% final sink--spike alignment despite differences in training budget and activation profile.
Third, full attention placement substantially affects retrieval performance despite having little effect on the final sink--spike alignment rate.
All three configurations achieve near-perfect alignment and comparable language-modeling and commonsense performance.
Nevertheless, middle and late placements outperform early placement on most retrieval tasks, showing that alignment alone does not predict retrieval performance.

\subsection{Training-Time Emergence of ISP}
\label{app:controlled-isp}
\paragraph{Experimental Setup.}
We study ISP formation using 24-layer GDN models with a fixed $3{:}1$ hybridization ratio, in which eight full attention layers are interleaved with 16 GDN layers.
At the 340M scale, the model is trained from scratch on 10B tokens following the controlled pretraining recipe described in \S\ref{app:controlled-training-setup}.
To trace the emergence and evolution of ISP, we record the maximum absolute hidden-state activation of the first token across model depth at 1B-token intervals.
At the final checkpoint, we compute the inter-spike retention score and evaluate downstream performance.
We also train a 1.3B counterpart on 50B tokens, recording trajectories every 5B tokens to test recurrence at a larger scale and training budget.

\begin{figure*}[t]
    \centering
    \begin{subfigure}[t]{0.49\textwidth}
        \centering
        \includegraphics[width=\linewidth]{
            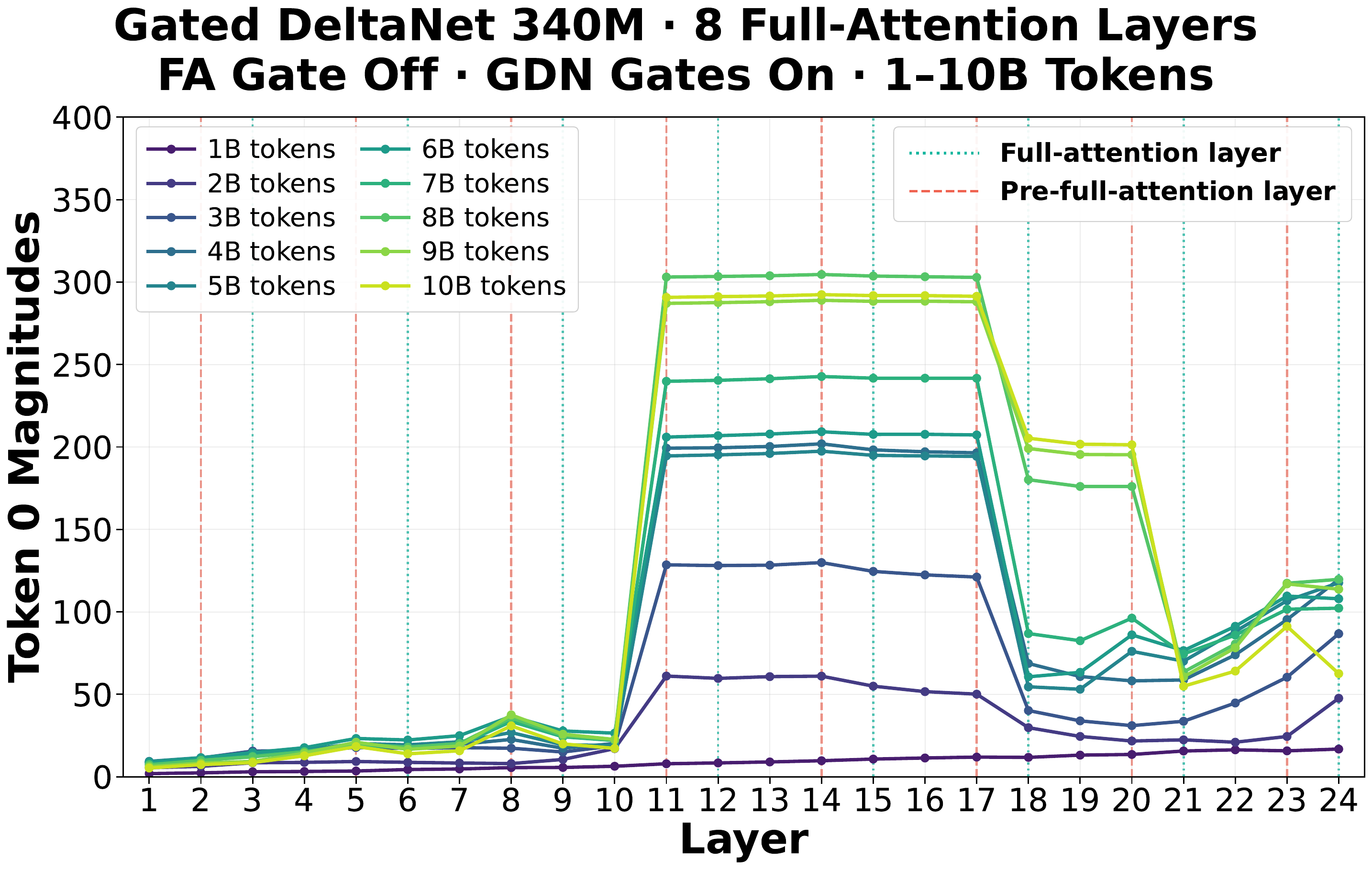
        }
        \caption{
            340M model trained on 10B tokens.
        }
        \label{fig:controlled-isp-scale-340m}
    \end{subfigure}
    \hfill
    \begin{subfigure}[t]{0.49\textwidth}
        \centering
        \includegraphics[width=\linewidth]{
            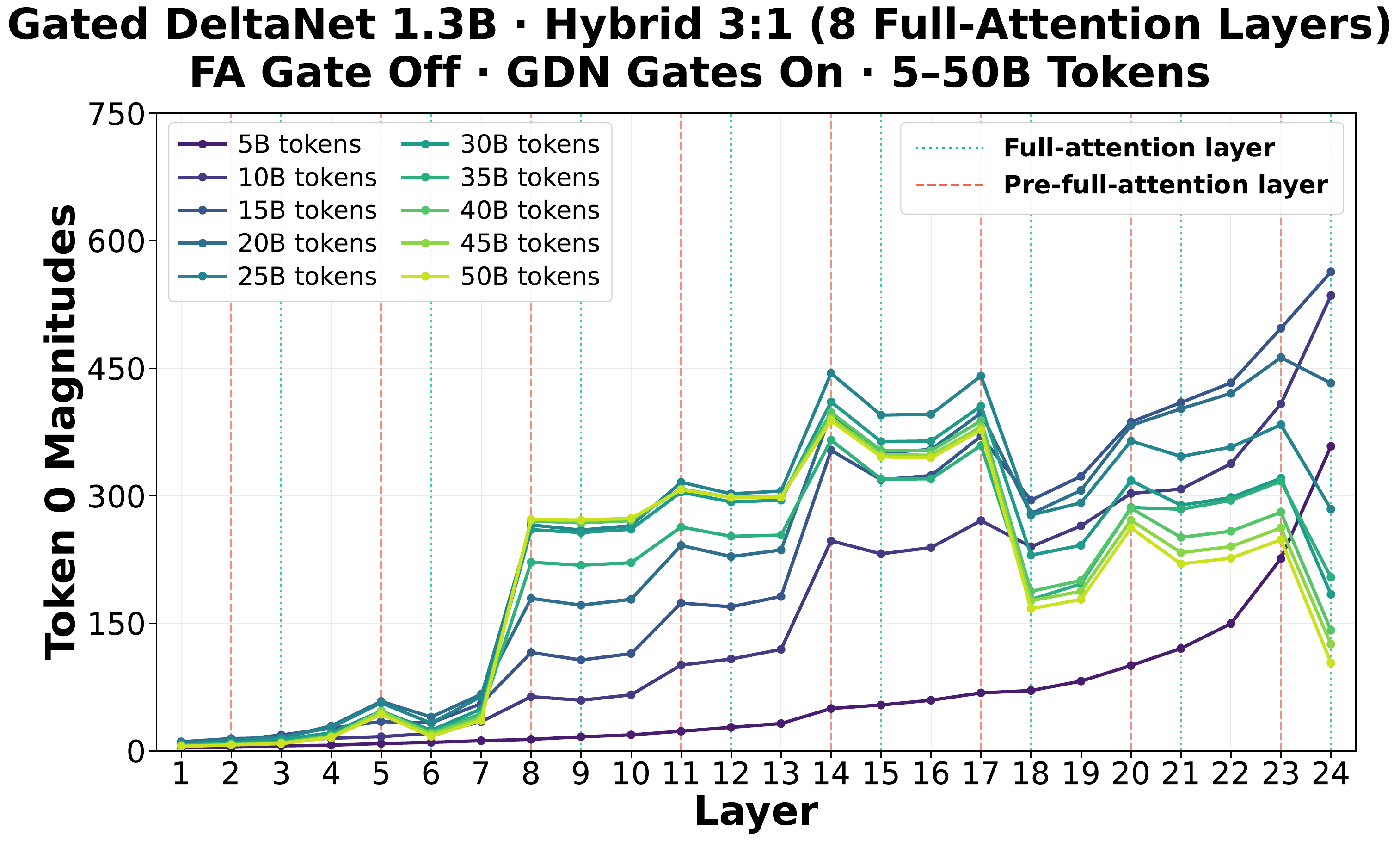
        }
        \caption{
            1.3B model trained on 50B tokens.
        }
        \label{fig:controlled-isp-scale-1p3b}
    \end{subfigure}
    \caption{
        \textbf{ISP formation across model scales.}
        Both models contain 24 sequence-mixing layers and use a $3{:}1$ hybridization ratio.
Inter-spike persistence recurs at 1.3B despite differences in magnitude and training budget.
    }
    \label{fig:controlled-isp-scale-comparison}
\end{figure*}

\paragraph{Results.}
Figure~\ref{fig:controlled-isp-loss} confirms stable optimization of the 1.3B ISP configuration through 50B tokens, while Figure~\ref{fig:controlled-isp-dynamics} traces the 340M first-token activation trajectory throughout pretraining.
Figure~\ref{fig:controlled-isp-scale-comparison} provides the corresponding cross-scale comparison, and Table~\ref{tab:controlled-isp-results} reports the final inter-spike retention scores and downstream performance.
The output-gating variants are analyzed in \S\ref{app:output-gating}.
First, ISP emerges early during pretraining and becomes progressively more pronounced and stable across subsequent checkpoints.
An elevated activation region between adjacent PAS is already visible at the first recorded checkpoint after 1B training tokens and gradually consolidates into a sustained plateau.
Second, the same organization recurs at 1.3B.
In the larger model, a broad plateau becomes evident over the early checkpoints and remains aligned with the interleaved full attention schedule through 50B tokens, although its absolute magnitude evolves non-monotonically during training.
Final ISR is 90.56\% at 340M and 94.47\% at 1.3B, confirming strong relative retention at both scales; the different training budgets preclude attributing this difference to scale alone.
These trajectories establish ISP as a training-time morphology recurring at both scales.

\subsection{Effects of Output Gating on PAS and ISP}
\label{app:output-gating}

\begin{figure*}[t]
    \centering
    \begin{subfigure}[t]{0.32\textwidth}
        \centering
        \includegraphics[width=\linewidth]{
            figures/training-time/isp_gdn_340_fa_3_1_gate_off_la_gate_on.pdf
        }
        \caption{
            Standard GDN.
        }
        \label{fig:controlled-isp-dynamics}
    \end{subfigure}
    \hfill
    \begin{subfigure}[t]{0.32\textwidth}
        \centering
        \includegraphics[width=\linewidth]{
            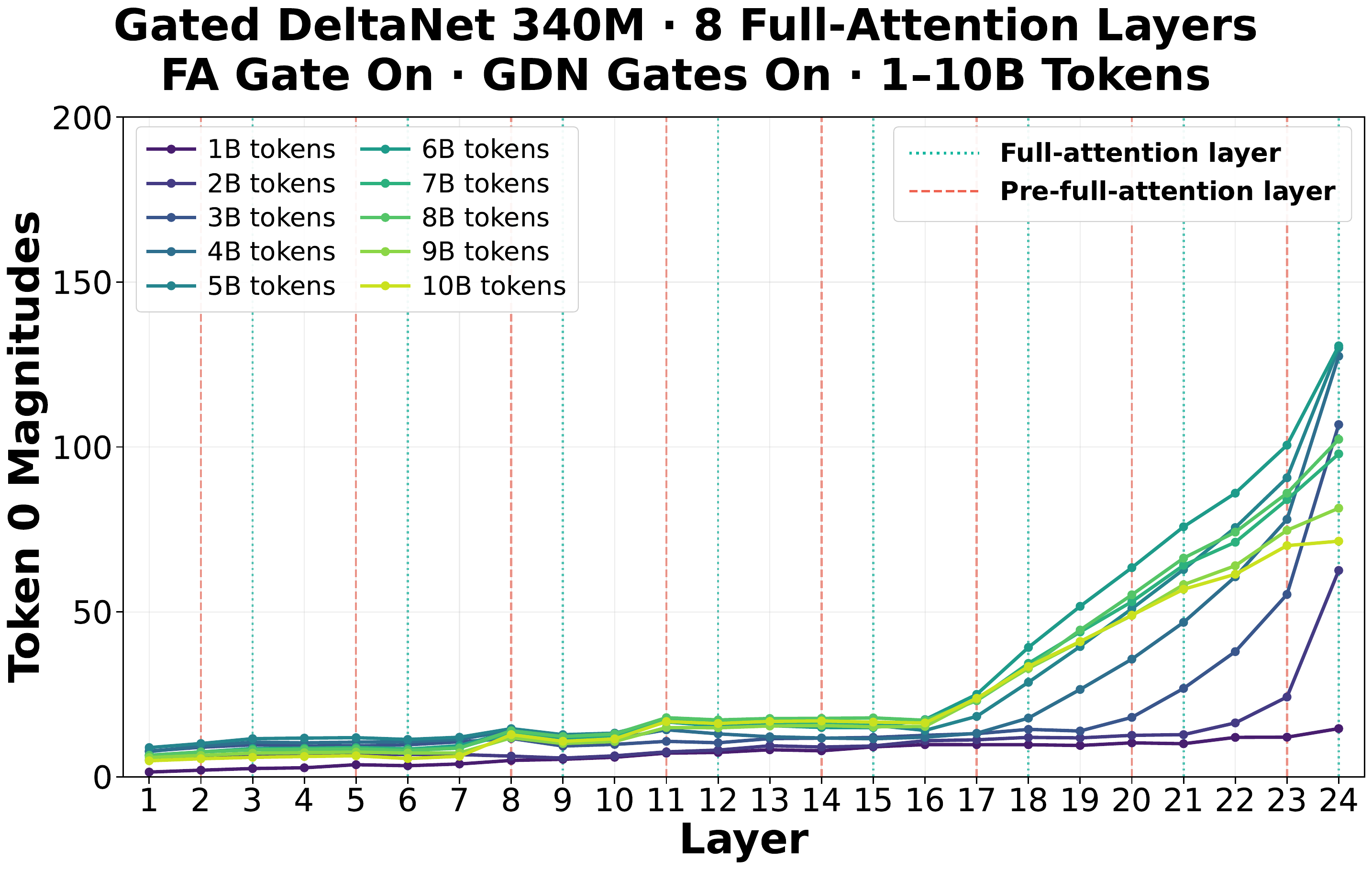
        }
        \caption{
            GDN-GatedFA.
        }
        \label{fig:controlled-isp-gatedfa}
    \end{subfigure}
    \hfill
    \begin{subfigure}[t]{0.32\textwidth}
        \centering
        \includegraphics[width=\linewidth]{
            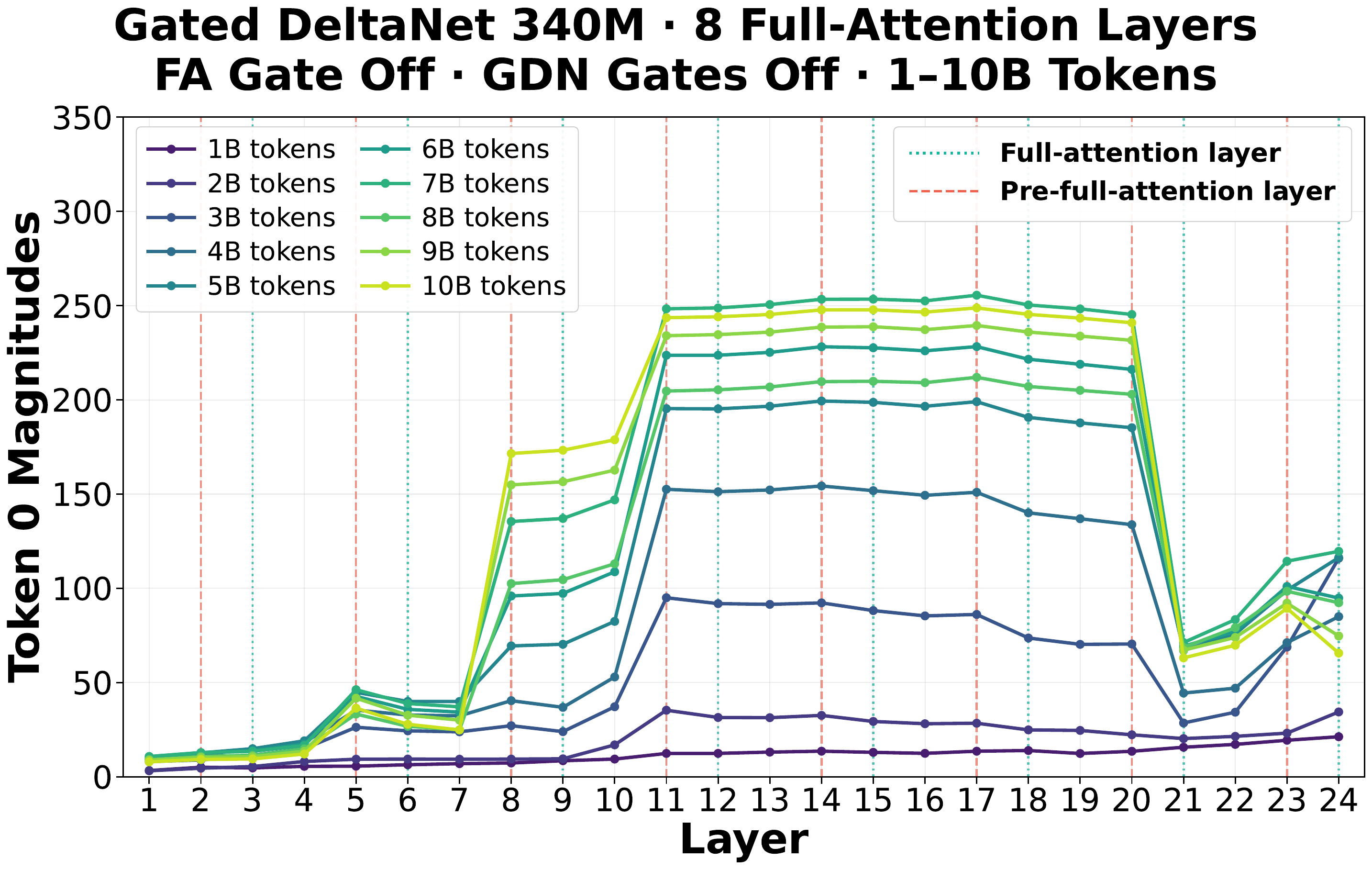
        }
        \caption{
            GDN-NoOutGate.
        }
        \label{fig:controlled-isp-nooutgate}
    \end{subfigure}
    \caption{
        \textbf{ISP dynamics under output-gating interventions.}
        Each panel traces the maximum absolute hidden-state activation of the first token across model depth at successive 1B-token checkpoints for a 340M GDN model with a $3{:}1$ hybridization ratio trained on 10B tokens.
        The standard model develops a sustained ISP.
        Adding output gates to the full attention layers markedly reduces its absolute magnitude, although a weak inter-spike plateau remains visible and persists throughout training.
Removing GDN output gates moderately amplifies the plateau.
    }
    \label{fig:controlled-isp-gating-dynamics}
\end{figure*}

\begin{table*}[t]
    \centering
    \caption{
        \textbf{ISP and downstream performance under controlled pretraining.}
        All configurations use 24 layers at $3{:}1$.
        ISR and downstream accuracies are percentages; PPL denotes perplexity.
        ISR measures persistence relative to each variant's own PAS endpoints.
    }
    \label{tab:controlled-isp-results}
    \small
    \setlength{\tabcolsep}{5pt}

    \textbf{(a) ISP retention, language modeling, and real-world retrieval}
    \par\smallskip

    \resizebox{0.92\textwidth}{!}{%
    \begin{tabular}{lcccccccc}
        \toprule
        \textbf{Variant}
        & \textbf{Scale}
        & \textbf{Tokens}
        & \textbf{ISR}
        & \textbf{PPL}
        & \textbf{FDA}
        & \textbf{SWDE}
        & \textbf{NQ}
        & \textbf{SQuAD} \\
        \midrule

        GDN
        & 340M & 10B
        & 90.56 & 13.74 & 54.62 & 39.68 & 21.63 & 38.40 \\

        GDN
        & 1.3B & 50B
        & 94.47 & 9.80 & 64.82 & 45.92 & 24.90 & 40.65 \\
        \midrule

        GDN-NoOutGate
        & 340M & 10B
        & 96.48 & 13.78 & 49.32 & 39.33 & 21.82 & 35.25 \\

        GDN-GatedFA
        & 340M & 10B
        & 99.48 & 13.68 & 65.12 & 40.49 & 21.32 & 36.60 \\

        \bottomrule
    \end{tabular}%
    }

    \medskip
    \textbf{(b) Synthetic retrieval and commonsense reasoning}
    \par\smallskip

    \resizebox{0.92\textwidth}{!}{%
    \begin{tabular}{lcccccccc}
        \toprule
        \textbf{Variant}
        & \textbf{Scale}
        & \textbf{Tokens}
        & \textbf{NIAH-1}
        & \textbf{NIAH-2}
        & \textbf{NIAH-3}
        & \textbf{HellaSwag}
        & \textbf{PIQA}
        & \textbf{ARC-E} \\
        \midrule

        GDN
        & 340M & 10B
        & 78.60 & 99.40 & 63.60 & 39.56 & 65.67 & 50.93 \\

        GDN
        & 1.3B & 50B
        & 96.00 & 100.00 & 90.00 & 53.36 & 71.55 & 61.11 \\

        \midrule

        GDN-NoOutGate
        & 340M & 10B
        & 57.20 & 48.20 & 14.20 & 39.57 & 66.16 & 50.25 \\

        GDN-GatedFA
        & 340M & 10B
        & 61.00 & 98.20 & 5.00 & 39.38 & 66.00 & 50.34 \\

        \bottomrule
    \end{tabular}%
    }
\end{table*}
\paragraph{Experimental Setup.}
We examine how output gating regulates PAS and ISP using the representative 340M configurations from Sections~\ref{app:controlled-pas} and~\ref{app:controlled-isp}.
The PAS configuration is a 24-layer GDN model with a single full attention layer at layer 12, whereas the ISP configuration uses a $3{:}1$ hybridization ratio, interleaving eight full attention layers with 16 GDN layers.
All models are trained from scratch on 10B tokens under the same controlled recipe.
For each configuration, we construct two variants while holding all other architectural components fixed.
\textsc{GDN-GatedFA} adds element-wise output gates to the full attention layers following gated attention~\citep{qiu2026gated}, while retaining the native GDN gates.
Conversely, \textsc{GDN-NoOutGate} removes all GDN output gates while leaving full attention ungated.
These interventions isolate the module-specific effects of output gating.
We record the maximum absolute hidden-state activation of the first token across model depth at successive 1B-token checkpoints and, at the final checkpoint, compute sink--spike alignment for PAS and inter-spike retention for ISP.

\paragraph{Results.}
Figures~\ref{fig:controlled-pas-gating-dynamics} and~\ref{fig:controlled-isp-gating-dynamics} compare the evolution of PAS and ISP under the standard configuration and the two gating interventions.
Tables~\ref{tab:controlled-pas-results} and~\ref{tab:controlled-isp-results} report the corresponding morphology metrics and downstream performance.
We assess absolute ISP magnitude from the trajectories and relative persistence from ISR, which normalizes by adjacent PAS.
First, adding output gates to full attention substantially attenuates the absolute magnitudes of PAS and ISP without eliminating their layerwise organization.
Weak spikes and plateaus persist and become increasingly visible during pretraining: the trajectories retain a low-magnitude plateau, while ISR confirms continued inter-spike persistence.
Consistently, Qwen3.5 models with native full attention output gating still exhibit both morphologies (Figures~\ref{fig:qwen35-35b-base}--\ref{fig:qwen35-397b}).
Second, removing the native GDN output gates increases PAS and ISP magnitudes, with a more visible effect on the plateau, but the changes remain smaller than those induced by full attention gating.
Thus, native GDN output gates modulate amplification and propagation but are not necessary for either morphology in these configurations.
Third, gating the substantially fewer full attention layers produces larger changes in the absolute activation trajectories than removing gates from all GDN layers.
This contrast motivates examining gate placement rather than gate count alone, while distinguishing the addition of full attention gates from the removal of GDN gates.

\FloatBarrier
\begingroup
\color{revisioncolor}
\subsection{Matched Pretraining with SiTU-GLU Soft Caps}
\label{app:situ-softcap}
\rev{Following the suggested SoftCap comparison, we train a matched GDN pair using standard SwiGLU and Kimi-K3-style SiTU-GLU. Let $c_b(x)=b\tanh(x/b)$; the modified MLP is $W_{\rm down}[c_4(g)\,\sigma(g)\,c_{25}(u)]$, with $g=W_gx$ and $u=W_ux$. Soft-capping is the scalar operation; SiTU-GLU is the specific gated-MLP construction using it. The original gate sigmoid is retained. The operation is installed before the down projection in all 24 MLPs.}

\rev{Both models have 392,896,144 parameters and six FA layers, at zero-based indices 3/7/11/15/19/23 ($\rho=4$ under the paper's definition). This independent training pair is distinct from the public $\rho=3$ checkpoints in the paired inference controls. Both models process 10,000,269,312 input tokens over 19,074 updates at a sequence length of 4,096 and use the same seed, data, tokenizer, global batch size, and learning-rate schedule. We report results for this matched seed.}
\begin{table}[H]
\centering
\small
\setlength{\tabcolsep}{5pt}
\caption{\textbf{Downstream evaluation after matched 10B-token pretraining.} Accuracy is in percent; perplexity is lower-is-better.}
\label{tab:trained-softcap-quality}
\begin{tabular}{@{}lrr@{}}
\toprule
\textbf{Benchmark} & \textbf{SwiGLU} & \textbf{SiTU-GLU}\\
\midrule
FDA & 34.15\% & 34.88\%\\
SWDE & 25.87\% & 24.46\%\\
Based SQuAD & 30.40\% & 29.86\%\\
Based NQ & 17.87\% & 17.83\%\\
PIQA & 65.13\% & 64.85\%\\
ARC-Easy & 55.51\% & 54.97\%\\
HellaSwag (acc\_norm) & 38.62\% & 38.61\%\\
NIAH-1 & 87.80\% & 69.60\%\\
NIAH-2 & 98.20\% & 95.20\%\\
NIAH-3 & 40.40\% & 17.40\%\\
\midrule
Ten-task mean & 49.39\% & 44.77\%\\
WikiText PPL & 16.7079 & 16.9549\\
\bottomrule
\end{tabular}
\end{table}
\begin{table}[H]
\centering
\small
\setlength{\tabcolsep}{5pt}
\caption{\textbf{Morphology after pretraining: original tie-aware SSA and clipped Eq.~(7) ISR.} The absolute peak is an additional descriptive statistic.}
\label{tab:trained-softcap-morphology}
\begin{tabular}{@{}lrr@{}}
\toprule
\textbf{Measure} & \textbf{SwiGLU} & \textbf{SiTU-GLU}\\
\midrule
First token SSA & 99.09\% & 98.96\%\\
First token ISR & 0.8503 & 0.8411\\
Midpoint SSA & 61.85\% & 61.07\%\\
Midpoint ISR & 0.9905 & 0.9901\\
First-token absolute FA peak & 261.32 & 241.26\\
\bottomrule
\end{tabular}
\end{table}
\rev{Soft caps moderate absolute peak amplitude, but PAS alignment and substantial inter-spike retention remain. In this setting, they do not improve downstream quality: the ten-task mean decreases from 49.39\% to 44.77\%, and perplexity increases from 16.7079 to 16.9549. Thus attenuating branch magnitudes does not remove the architecture-aligned organization or produce a quality gain in the matched comparison.}

\endgroup
\FloatBarrier

\section{Additional Systematic-Outlier Analyses of PAS and ISP}
\label{app:lifecycle-analysis}

\subsection{Fixed-Coordinate Systematic-Outlier Analysis of PAS}
\label{app:detailed-pas-lifecycle}
\begin{figure*}[t]
    \centering
    \includegraphics[width=\textwidth]{
        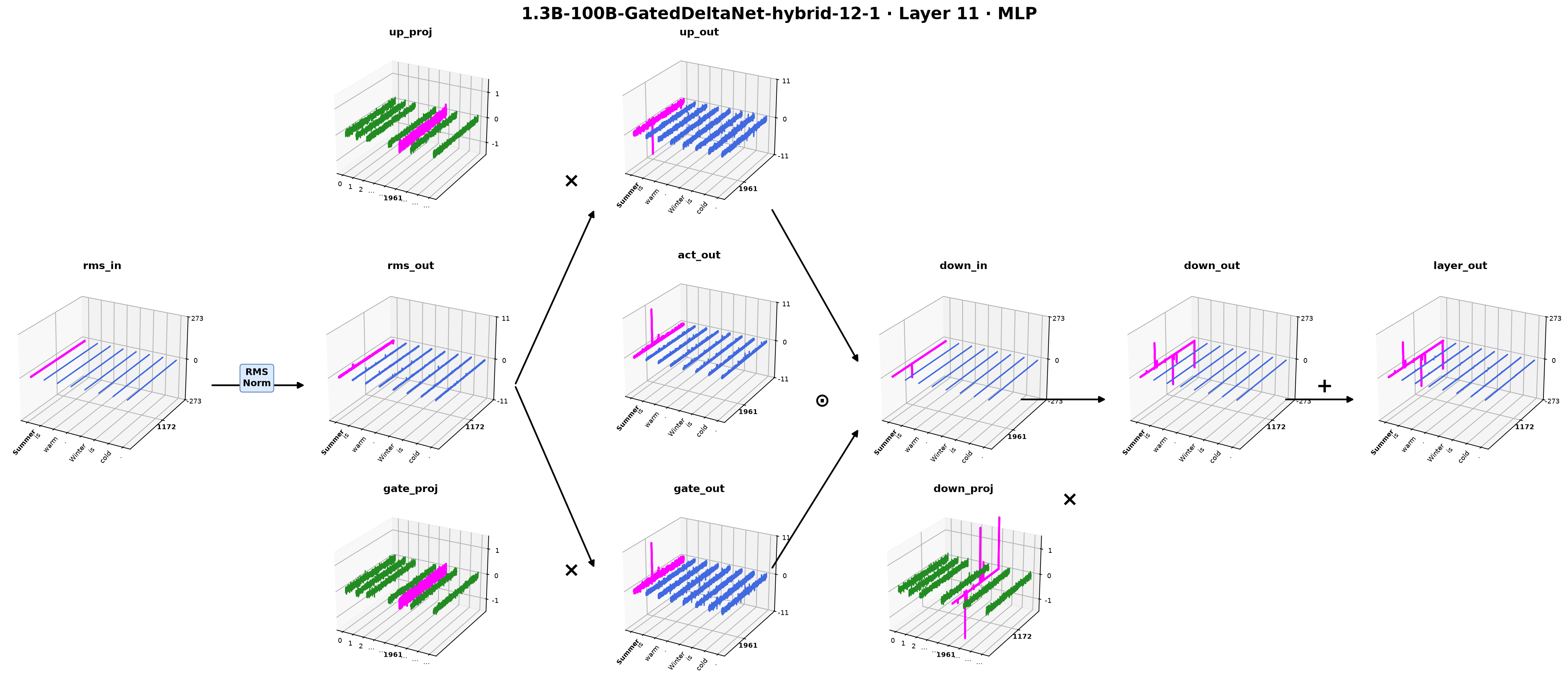
    }
    \caption{
        \textbf{Fixed-coordinate analysis of PAS formation.}
        At $(t^\star,j^\star)$, layer 11 produces an extreme update that writes a large outlier into the residual stream immediately before full attention.
    }
    \label{fig:pas-formation-detail}
\end{figure*}
\begin{figure*}[t]
    \centering
    \includegraphics[width=\textwidth]{
        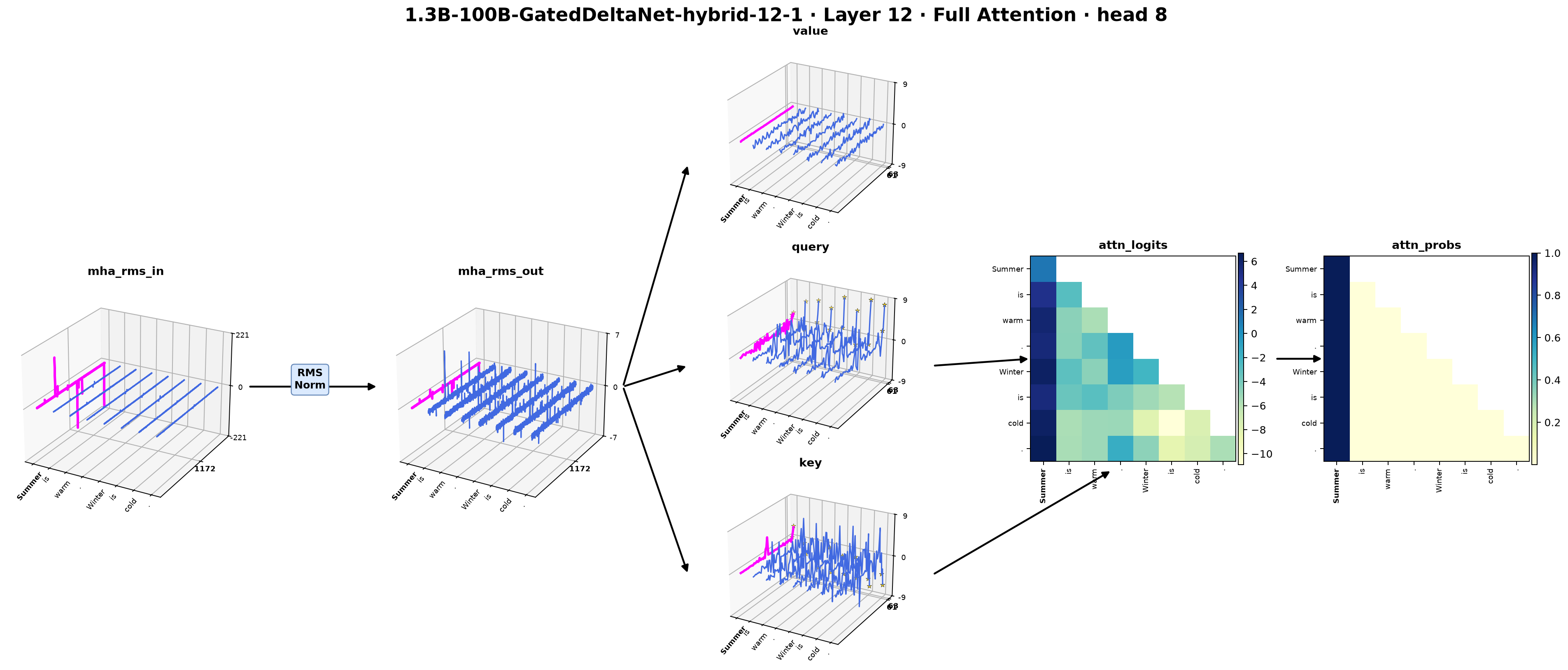
    }
    \caption{
        \textbf{Coupling between PAS and full attention sink behavior.}
        During full attention at layer 12, token $t^\star$ receives a disproportionate share of attention from subsequent query positions and acts as the dominant attention sink.
    }
    \vspace{+5mm}
    \label{fig:pas-sink-detail}
\end{figure*}
\begin{figure*}[t]
    \centering
    \includegraphics[width=\textwidth]{
        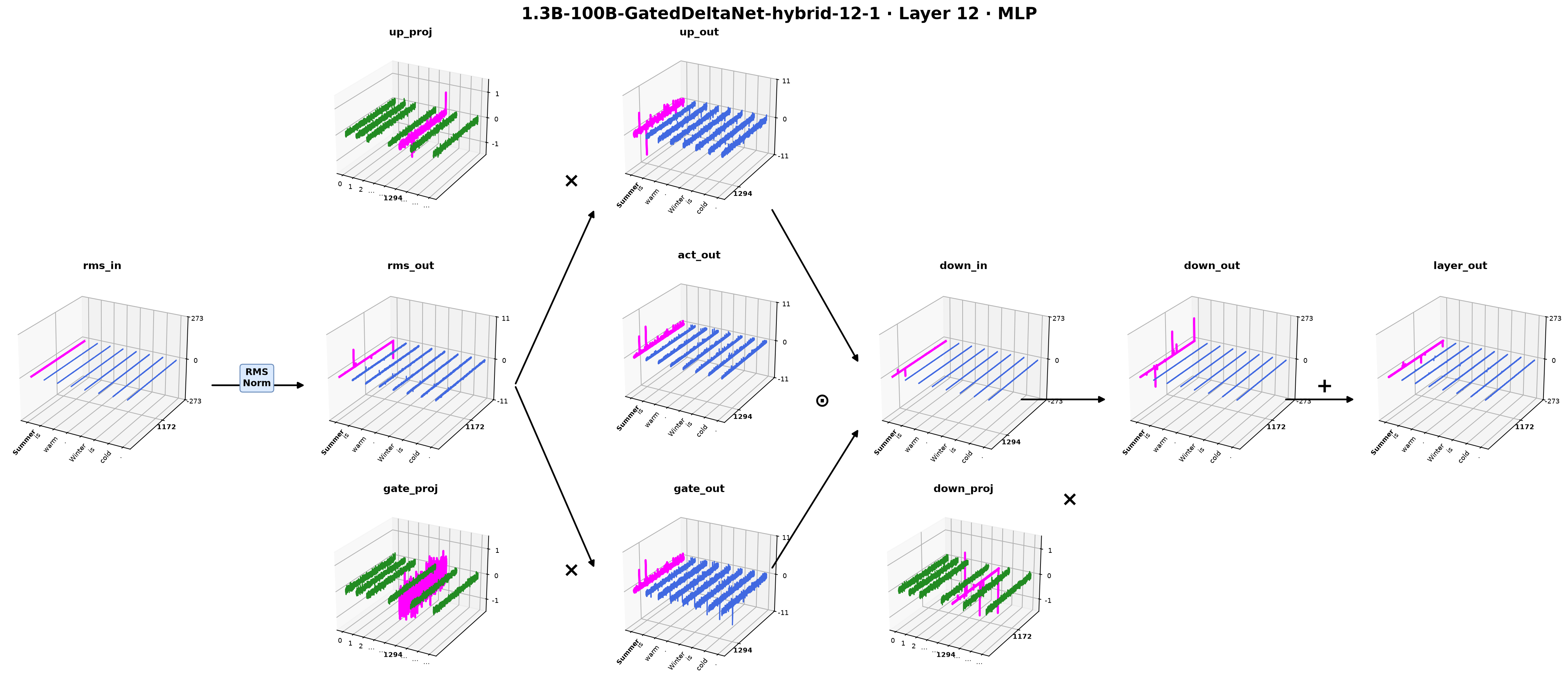
    }
    \caption{
        \textbf{Fixed-coordinate analysis of PAS cancellation.}
        At $(t^\star,j^\star)$, an opposite-signed layer-12 update substantially cancels the incoming outlier.
    }
    \label{fig:pas-dissipation-detail}
\end{figure*}

We examine the PAS immediately preceding the full attention layer $f=12$ in a representative 1.3B GDN model with a $12{:}1$ hybridization ratio.
Following \S\ref{sec:spike-lifecycle}, let $t^\star$ denote the consensus sink token and $j^\star=\operatorname*{arg\,max}_{j}|X_{t^\star,j}^{(f-1)}|$ its maximally activated feature at the corresponding pre-attention layer.
We fix this token--feature coordinate and trace its signed activation and module-level updates through three stages: outlier formation, attention-sink coupling, and cancellation.
Figures~\ref{fig:pas-formation-detail}, \ref{fig:pas-sink-detail}, and~\ref{fig:pas-dissipation-detail} provide detailed evidence for these stages, respectively.

\paragraph{Outlier Formation.}
Figure~\ref{fig:pas-formation-detail} shows that the pre-attention layer produces an extreme update at the fixed coordinate, writing a large outlier into the residual stream immediately before full attention.
This localization establishes that PAS arises from a concentrated update at a specific token--feature coordinate rather than a diffuse increase across tokens or features.

\paragraph{Attention-Sink Coupling.}
As shown in Figure~\ref{fig:pas-sink-detail}, token $t^\star$ receives a disproportionate share of attention from subsequent query positions during full attention at layer 12, acting as the dominant attention sink.
This links the fixed-coordinate outlier to full attention sink behavior.

\paragraph{Outlier Cancellation.}
Figure~\ref{fig:pas-dissipation-detail} tracks this coordinate through layer 12, where an opposite-signed update substantially cancels the incoming outlier and sharply reduces its residual magnitude.
Following one coordinate links formation, attention-sink coupling, and cancellation within a coherent PAS lifecycle rather than unrelated layerwise extrema.

\FloatBarrier
\begingroup
\color{revisioncolor}
\paragraph{Fixed-coordinate ISP extension.}
\rev{We track the sink token in the M-A-P GDN 340M and 1.3B hybrids ($\rho=3$). Top-1 identity compares the largest-magnitude coordinate at adjacent residual boundaries. Top-4 overlap is the fraction of their four largest coordinates by magnitude that they share. Anchor membership measures how often the largest-magnitude coordinate at the left FA input remains in the top four through the next FA input. Table~\ref{tab:gdn-isp-coordinate} summarizes five natural domains.}
\begin{table}[H]
\centering
\small
\setlength{\tabcolsep}{5pt}
\caption{\textbf{Coordinate identity in the two M-A-P GDN hybrids with $\rho=3$.}}
\label{tab:gdn-isp-coordinate}
\begin{tabular}{@{}lrrr@{}}
\toprule
\textbf{Model} & \textbf{Top-1 identity} & \textbf{Top-4 overlap} & \textbf{Anchor in top 4}\\
\midrule
340M & 73.91\% & 90.22\% & 100.00\%\\
1.3B & 86.96\% & 86.96\% & 90.48\%\\
\bottomrule
\end{tabular}
\end{table}
\rev{The high identity, overlap, and anchor membership indicate that plateaus typically involve a fixed coordinate or a stable set of dominant coordinates. This agrees with the fixed-coordinate PAS analysis and supports the write--sink--cancel account. In each model, the inputs from all five domains repeat the same leading-token sink trajectory.}

\endgroup
\FloatBarrier

\subsection{Cross-Model Systematic-Outlier Analyses}
\label{app:layerwise-lifecycle}
We extend \S\ref{sec:lifecycle-analysis} with layerwise systematic-outlier visualizations for M-A-P and large pretrained hybrids, tracing the first token of \textit{``Summer is warm. Winter is cold.''}
Each figure pairs the MA envelope with signed residual input, mixer update, FFN update, and block output at each layer's dominant feature.
The dominant feature is reselected at each layer, so these plots characterize layerwise write--cancel structure; Appendix~\ref{app:detailed-pas-lifecycle} separately follows one fixed coordinate for a representative PAS.
In particular, an ISP envelope can persist while its dominant feature changes.

\begin{figure*}[p]
    \centering
    \includegraphics[
        width=\textwidth,
        height=0.72\textheight,
        keepaspectratio
    ]{
        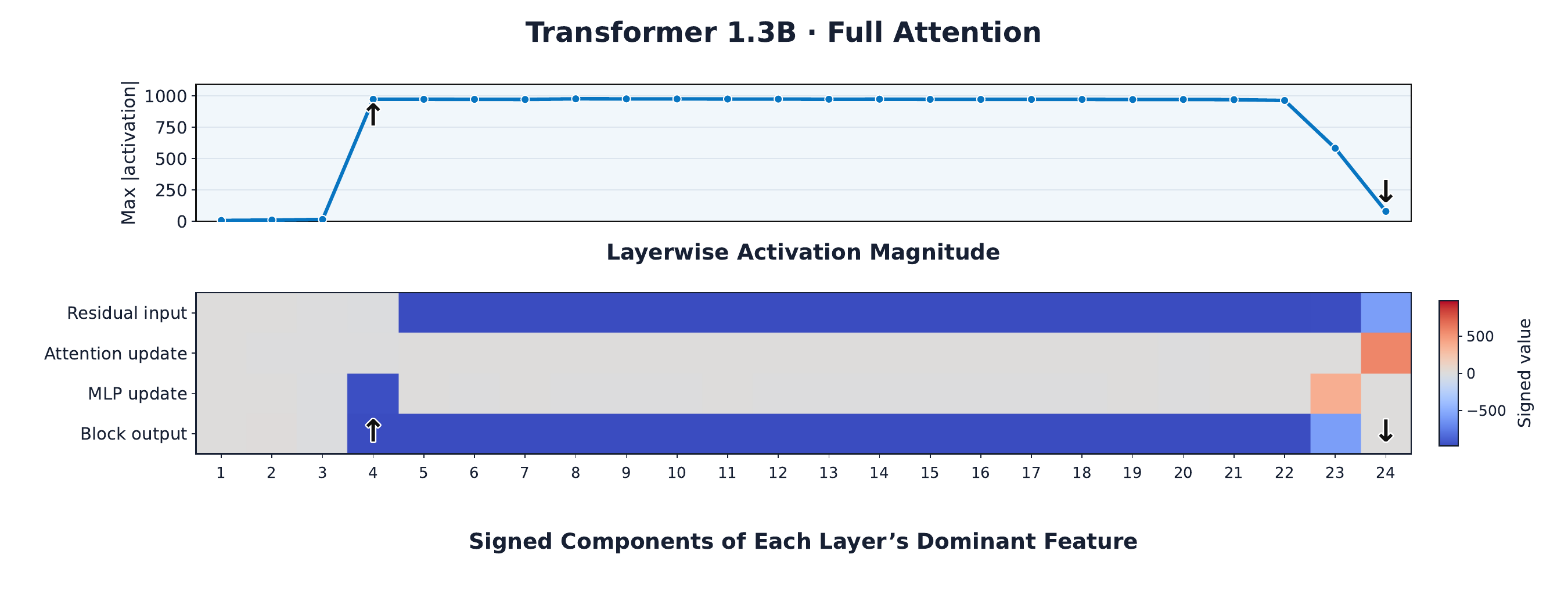
    }
    \vspace{-10mm}
    \caption{
        \textbf{Layerwise systematic-outlier pattern in the 1.3B
        full attention Transformer from the M-A-P suite.}
        MAs persist across much of the model depth, providing the full attention reference for the PAS--ISP continuum.
    }
    \label{fig:lifecycle-full-attention}
\end{figure*}

\paragraph{M-A-P Model Suite.}
We analyze 1.3B GDN, DeltaNet, GLA, HGRN, and RetNet models~\citep{wang2025systematic} at $24{:}1$, $12{:}1$, $6{:}1$, and $3{:}1$ (Figures~\ref{fig:lifecycle-gdn-ratios}--\ref{fig:lifecycle-retnet-ratios}).
Figure~\ref{fig:lifecycle-full-attention} shows the common full attention reference once, since this limit is independent of the linear backbone.
Across configurations, dominant outliers arise before full attention and are subsequently accompanied by opposite-signed updates.
Denser full attention leaves larger portions of the inter-spike intervals elevated, consistent with delayed cancellation; at the full attention limit, spikes and plateaus merge into persistent MAs.
These signed traces complement the quantitative morphology analysis in \S\ref{sec:ratio-dependence}.

\paragraph{Large-Scale Open-Source Hybrid Models.}
Figures~\ref{fig:lifecycle-qwen-models} and~\ref{fig:lifecycle-nemotron-models} examine three Qwen3.5 and three Nemotron-H checkpoints, respectively.
Qwen3.5 interleaves GDN with output-gated full attention; Nemotron-H combines Mamba-2 with full attention.
Both families show amplification near full attention boundaries and extended high-activation regions between them, despite different mixers, scales, and schedules.
Thus, the layerwise pattern also occurs with native full attention output gating and in state-space hybrids, broadening the cancellation-timing interpretation beyond M-A-P.

\FloatBarrier
\begingroup
\color{revisioncolor}
\subsection{A Conditional Account of Readout and Persistence}
\label{app:conditional-readout}
\rev{An MLP can write a large residual component before full attention, and later updates can oppose it. Whether the component is still large at the next FA input depends on the FA spacing and the size of the opposing updates. We describe the conditions for attention to read the component and for the component to persist across layers.}

\paragraph{From a large component to attention readout.}
\rev{At source token $s$, write the $d$-dimensional residual vector entering FA as $x=a u+r$, with $\|u\|=1$ and $u^\top r=0$. For fixed RMSNorm weights $\Gamma$, as $\|r\|/|a|\to0$ and $\epsilon d/a^2\to0$, $\Gamma x/\sqrt{\|x\|^2/d+\epsilon}\to\operatorname{sign}(a)\sqrt d\,\Gamma u$. In this regime the component dominates the normalized direction. Its projected key contributes to the attention score, and its projected value contributes to the readout. Let $T$ be the number of source tokens visible to query $q$. If the post-RoPE logit for $s$ exceeds that of every other visible token $t$ by at least $\gamma>0$, then its attention weight satisfies}
\begin{equation}
 P_{qs}\ge\frac{1}{1+(T-1)e^{-\gamma}}.
\end{equation}
\rev{A sufficiently large margin concentrates attention on the source token. The FA output and subsequent residual updates can oppose the written direction, producing a localized spike.}

\paragraph{Why a shorter inter-FA interval can retain the component.}
\rev{Fix a signed direction and let $a_0>0$ denote its initial projected amplitude. After $m$ subsequent updates, let $P_m$ be the sum of positive projections and $C_m$ the sum of the absolute values of negative projections. For any retention fraction $0<\theta\le1$, the projected amplitude obeys}
\begin{equation}
 a_m=a_0+P_m-C_m,\qquad
 a_m\ge\theta a_0\iff C_m-P_m\le(1-\theta)a_0.
\end{equation}
\rev{If the net opposing update $C_m-P_m$ never exceeds this threshold before the next FA input, the component stays elevated between the inputs. This predicts that shorter FA gaps can connect neighboring spikes into a plateau. In the placement and density experiments, peaks follow the FA boundaries, and denser FA placement promotes persistence. Appendix~\ref{app:detailed-pas-lifecycle} tracks the coordinates that persist across these gaps.}

\paragraph{Objective versus intervention.}
\rev{Pretraining minimizes next-token loss over the training distribution. This objective supervises long-range retrieval indirectly, at positions whose next token depends on distant context. A model can learn to cancel a component after one readout even when retaining it would improve later retrieval. The reference-conditioned edit in Eq.~\ref{eq:main-connection} uses the unedited trajectory to set a signed floor, keeping the component available across a longer interval.}

\endgroup
\FloatBarrier

\section{Functional Intervention Protocols and Results}
\label{app:functional-interventions}
\begingroup
\subsection{Intervention Definition and Evaluation Scope}
\label{app:functional-protocol}
\paragraph{Checkpoints and coordinates.}
We use official M-A-P GDN checkpoints, distinct from the controlled-pretraining models in \S\ref{sec:controlled-pretraining}. Both scales have 24 blocks; $\rho=12,6,3$ corresponds to two, four, and eight full attention layers. Both interventions use the first prompt token as a fixed anchor, following \S\ref{sec:architecture-dependence}. As in \S\ref{sec:functional-interventions}, $p_i=f_i-1$ indexes the residual state entering full attention and $z^{(\ell)}=X^{(\ell)}_{1,:}$ its normal reference trajectory. We select $C_i=\operatorname{TopK}(|z^{(p_i)}|,k)$ and use the union $C=\bigcup_i C_i$ for connection.

\paragraph{Deletion and connection.}
Deletion sets $\widetilde{X}_{1,j}^{(p_i)} = 0$ for every $j\in C_i$ and every $i$. Coordinates selected in the normal pass remain fixed throughout deletion. Connection acts on $C$ and sets, for $p_i<\ell<p_{i+1}$,
\begin{equation}
\widetilde X^{(\ell)}_{1,j}
=s_{i,j}\max\!\left(s_{i,j}z^{(\ell)}_j,\,|z^{(p_i)}_j|\right),
\quad s_{i,j}=\begin{cases}1,&z^{(p_i)}_j\geq0,\\-1,&z^{(p_i)}_j<0,\end{cases}
\qquad j\in C.
\label{eq:connection-intervention}
\end{equation}
\rev{At each FA input, the interior-only procedure uses the residual produced by the preceding layers. The original combined procedure also sets $\widetilde X^{(p_i)}_{1,j}=z^{(p_i)}_j$ for $j\in C$ at every endpoint; unselected entries follow the intervened computation. Appendix~\ref{app:paired-controls} reports both variants and the endpoint-only control.} Equation~\ref{eq:connection-intervention} uses the normal reference trajectory throughout, allowing signs to change at endpoints. The budget $k$ applies \emph{per input layer}, so the union $C$ may contain more than $k$ coordinates. Every example is evaluated, including identity cases already satisfying the target and weak-activation cases. Coordinate selection and reference values depend exclusively on activations.

\paragraph{Checks and scoring.}
\rev{We check coordinate-level target attainment. For the original combined procedure we also verify the selected endpoint assignments; the endpoint-only control reproduces the normal retrieval outputs.} Accelerated graph execution is checked against native forwards, and reused normal results must match the frozen baseline. Models run in BF16 with greedy retrieval decoding; frozen manifests record the execution seed and numerical configuration. For autoregressive retrieval, the intervention edits the original first-token state during prompt processing, retaining its token identity during generation. Model weights remain frozen.
We use the benchmark formats and metrics in Appendix~\ref{app:controlled-benchmark-setup}. PIQA, ARC-Easy, and HellaSwag contain 1,838, 2,376, and 10,042 examples; HellaSwag uses length-normalized accuracy. FDA, SWDE, SQuAD, and NQ contain 1,101, 1,067, 2,977, and 3,157 cloze-completion examples, with up to 2,048 prompt and 48 generated tokens. SQuAD measures retrieval accuracy, not extractive F1. Each NIAH variant has 500 examples and a 4,096-token budget (up to 3,968 prompt plus 128 generated tokens), scored by exact match. WikiText-103 uses 2,048-token windows with stride 1,024: 326 windows cover 334,660 scored target tokens. Perplexity exponentiates their mean NLL.

\paragraph{Aggregation and interpretation.}
Accuracy is reported as a percentage. For each task, Table~\ref{tab:functional-summary} averages the six models equally. Relative change is $100(\bar S_{\rm int}/\bar S_{\rm normal}-1)$: the ratio of model means, not the mean of relative changes or a percentage-point difference. The ten-task mean weights accuracy tasks equally, excluding perplexity. Each model's perplexity exponentiates aggregate token NLL; averaging these perplexities is descriptive, not pooled likelihood. Appendix~\ref{app:functional-full-results} retains every score, including regressions, and reports exploratory paired uncertainty for retrieval.
Deletion precedes normalization and can alter the full normalized representation. The results quantify dependence on the selected PAS coordinates. \rev{The original results in Appendices~\ref{app:functional-full-results} and~\ref{app:open-model-interventions} concern the combined rule. The ablation in Appendix~\ref{app:paired-controls} shows that interior edits suffice for the mean gain, so endpoint restoration is omitted in the simplified procedure. The interior floor still uses the normal reference trajectory.}
\endgroup

\subsection{Complete Model-Specific Results}
\label{app:functional-full-results}
Tables~\ref{tab:functional-2fa340}--\ref{tab:functional-8fa13b_official} retain all per-model results; relative changes use each model's own normal score.

\begingroup
The retrieval mean weights FDA, SWDE, SQuAD, and NQ equally, retaining negative task changes. All eighteen model--$k$ settings improve this mean by $1.06\%$--$12.60\%$ relative to normal inference. These settings share checkpoints and evaluation examples.

\paragraph{Exploratory Paired Uncertainty.}
Using existing paired win/loss/tie counts, we bootstrap binary correctness differences within each task and recompute the four-task mean (100,000 resamples; seed 20260919). Pointwise 95\% percentile intervals are positive for all eighteen means; Table~\ref{tab:retrieval-consistency} reports $k=4$, with all budgets in the supplementary data. Two-sided exact McNemar tests, Holm-corrected over 72 model--budget--task contrasts, detect 24 positive changes and no negative changes at $\alpha=0.05$; non-rejection does not establish equivalence. This post-hoc analysis assumes independent items within tasks, without document/site clustering or training-seed uncertainty.

Appendix~\ref{app:functional-protocol} specifies scoring and context limits. The following tables retain the complete task-level results, including regressions and non-monotonic deletion responses.
\endgroup
\begin{table}[!ht]
\centering
\small
\setlength{\tabcolsep}{5pt}
\caption{\textbf{Connection gains and paired uncertainty.} $k=4$. Scores average FDA, SWDE, SQuAD, and NQ equally. Relative change is measured against normal inference; $\Delta$ and its exploratory item-paired 95\% bootstrap interval are in percentage points (pp). Intervals are pointwise and do not account for document/site clustering or training-seed variation.}
\label{tab:retrieval-consistency}
\begingroup
\begin{tabular}{@{}llrrrl@{}}
\toprule
Scale & Ratio & Normal (\%) & Connect (\%) & Relative (\%) & $\Delta$ [95\% CI] (pp) \\
\midrule
340M & $12{:}1$ & 23.16 & 23.88 & $+3.11$ & $+0.72\ [0.41,1.04]$ \\
     & $6{:}1$ & 23.36 & 24.37 & $+4.29$ & $+1.00\ [0.61,1.39]$ \\
     & $3{:}1$ & 30.95 & 34.85 & $+12.60$ & $+3.90\ [3.23,4.57]$ \\
\midrule
1.3B & $12{:}1$ & 42.82 & 43.31 & $+1.14$ & $+0.49\ [0.18,0.79]$ \\
     & $6{:}1$ & 45.04 & 47.15 & $+4.68$ & $+2.11\ [1.54,2.68]$ \\
     & $3{:}1$ & 46.36 & 48.95 & $+5.57$ & $+2.58\ [2.04,3.14]$ \\
\bottomrule
\end{tabular}
\endgroup
\end{table}

\begin{table}[H]
\centering
\footnotesize
\setlength{\tabcolsep}{3pt}
\caption{\textbf{340M GDN, $12{:}1$.} Accuracy (\%) and PPL; relative changes in parentheses.}
\label{tab:functional-2fa340}
\begingroup
\begin{tabular}{@{}lrrrrrrr@{}}
\toprule
Benchmark & Normal & \multicolumn{3}{c}{PAS suppression} & \multicolumn{3}{c}{Spike connection} \\
\midrule
 &  & $k=1$ & $k=2$ & $k=4$ & $k=1$ & $k=2$ & $k=4$ \\
\midrule
PIQA & 67.41 & \shortstack[r]{66.27\\ $(-1.69\%)$} & \shortstack[r]{66.92\\ $(-0.73\%)$} & \shortstack[r]{67.36\\ $(-0.08\%)$} & \shortstack[r]{67.46\\ $(+0.08\%)$} & \shortstack[r]{67.74\\ $(+0.48\%)$} & \shortstack[r]{67.57\\ $(+0.24\%)$} \\
ARC-Easy & 58.54 & \shortstack[r]{57.07\\ $(-2.52\%)$} & \shortstack[r]{56.86\\ $(-2.88\%)$} & \shortstack[r]{56.61\\ $(-3.31\%)$} & \shortstack[r]{58.63\\ $(+0.14\%)$} & \shortstack[r]{58.84\\ $(+0.50\%)$} & \shortstack[r]{58.63\\ $(+0.14\%)$} \\
HellaSwag & 42.17 & \shortstack[r]{40.22\\ $(-4.63\%)$} & \shortstack[r]{40.21\\ $(-4.65\%)$} & \shortstack[r]{40.02\\ $(-5.10\%)$} & \shortstack[r]{42.08\\ $(-0.21\%)$} & \shortstack[r]{42.17\\ $(+0.00\%)$} & \shortstack[r]{42.04\\ $(-0.31\%)$} \\
FDA & 22.98 & \shortstack[r]{5.63\\ $(-75.49\%)$} & \shortstack[r]{6.36\\ $(-72.33\%)$} & \shortstack[r]{7.08\\ $(-69.17\%)$} & \shortstack[r]{23.34\\ $(+1.58\%)$} & \shortstack[r]{22.98\\ $(+0.00\%)$} & \shortstack[r]{22.71\\ $(-1.19\%)$} \\
SWDE & 22.21 & \shortstack[r]{8.72\\ $(-60.76\%)$} & \shortstack[r]{11.81\\ $(-46.84\%)$} & \shortstack[r]{12.46\\ $(-43.88\%)$} & \shortstack[r]{22.40\\ $(+0.84\%)$} & \shortstack[r]{22.40\\ $(+0.84\%)$} & \shortstack[r]{22.59\\ $(+1.69\%)$} \\
SQuAD & 31.94 & \shortstack[r]{19.42\\ $(-39.22\%)$} & \shortstack[r]{25.16\\ $(-21.24\%)$} & \shortstack[r]{28.08\\ $(-12.09\%)$} & \shortstack[r]{34.97\\ $(+9.46\%)$} & \shortstack[r]{35.27\\ $(+10.41\%)$} & \shortstack[r]{34.50\\ $(+7.99\%)$} \\
NQ & 15.52 & \shortstack[r]{12.35\\ $(-20.41\%)$} & \shortstack[r]{12.45\\ $(-19.80\%)$} & \shortstack[r]{12.80\\ $(-17.55\%)$} & \shortstack[r]{15.74\\ $(+1.43\%)$} & \shortstack[r]{15.77\\ $(+1.63\%)$} & \shortstack[r]{15.74\\ $(+1.43\%)$} \\
NIAH-1 & 36.80 & \shortstack[r]{15.60\\ $(-57.61\%)$} & \shortstack[r]{13.60\\ $(-63.04\%)$} & \shortstack[r]{13.40\\ $(-63.59\%)$} & \shortstack[r]{36.80\\ $(+0.00\%)$} & \shortstack[r]{36.60\\ $(-0.54\%)$} & \shortstack[r]{37.00\\ $(+0.54\%)$} \\
NIAH-2 & 50.40 & \shortstack[r]{15.80\\ $(-68.65\%)$} & \shortstack[r]{20.00\\ $(-60.32\%)$} & \shortstack[r]{22.00\\ $(-56.35\%)$} & \shortstack[r]{51.60\\ $(+2.38\%)$} & \shortstack[r]{51.80\\ $(+2.78\%)$} & \shortstack[r]{51.20\\ $(+1.59\%)$} \\
NIAH-3 & 12.00 & \shortstack[r]{2.60\\ $(-78.33\%)$} & \shortstack[r]{2.00\\ $(-83.33\%)$} & \shortstack[r]{1.60\\ $(-86.67\%)$} & \shortstack[r]{12.40\\ $(+3.33\%)$} & \shortstack[r]{12.40\\ $(+3.33\%)$} & \shortstack[r]{11.20\\ $(-6.67\%)$} \\
\shortstack[l]{WikiText-103\\PPL} & 14.0183 & \shortstack[r]{15.8006\\ $(+12.71\%)$} & \shortstack[r]{15.6746\\ $(+11.82\%)$} & \shortstack[r]{15.8038\\ $(+12.74\%)$} & \shortstack[r]{14.0152\\ $(-0.02\%)$} & \shortstack[r]{14.0127\\ $(-0.04\%)$} & \shortstack[r]{14.0109\\ $(-0.05\%)$} \\
\midrule
Ten-task mean & 36.00 & \shortstack[r]{24.37\\ $(-32.31\%)$} & \shortstack[r]{25.54\\ $(-29.06\%)$} & \shortstack[r]{26.14\\ $(-27.38\%)$} & \shortstack[r]{36.54\\ $(+1.51\%)$} & \shortstack[r]{36.60\\ $(+1.66\%)$} & \shortstack[r]{36.32\\ $(+0.89\%)$} \\
\bottomrule
\end{tabular}
\endgroup
\end{table}
\begin{table}[H]
\centering
\footnotesize
\setlength{\tabcolsep}{3pt}
\caption{\textbf{1.3B GDN, $12{:}1$.} Accuracy (\%) and PPL; relative changes in parentheses.}
\label{tab:functional-2fa13b}
\begingroup
\begin{tabular}{@{}lrrrrrrr@{}}
\toprule
Benchmark & Normal & \multicolumn{3}{c}{PAS suppression} & \multicolumn{3}{c}{Spike connection} \\
\midrule
 &  & $k=1$ & $k=2$ & $k=4$ & $k=1$ & $k=2$ & $k=4$ \\
\midrule
PIQA & 73.29 & \shortstack[r]{72.96\\ $(-0.45\%)$} & \shortstack[r]{73.07\\ $(-0.30\%)$} & \shortstack[r]{70.62\\ $(-3.64\%)$} & \shortstack[r]{73.23\\ $(-0.07\%)$} & \shortstack[r]{73.12\\ $(-0.22\%)$} & \shortstack[r]{73.07\\ $(-0.30\%)$} \\
ARC-Easy & 72.64 & \shortstack[r]{71.76\\ $(-1.22\%)$} & \shortstack[r]{71.59\\ $(-1.45\%)$} & \shortstack[r]{65.78\\ $(-9.44\%)$} & \shortstack[r]{72.18\\ $(-0.64\%)$} & \shortstack[r]{72.39\\ $(-0.35\%)$} & \shortstack[r]{72.43\\ $(-0.29\%)$} \\
HellaSwag & 60.99 & \shortstack[r]{60.77\\ $(-0.36\%)$} & \shortstack[r]{60.74\\ $(-0.41\%)$} & \shortstack[r]{55.90\\ $(-8.36\%)$} & \shortstack[r]{60.83\\ $(-0.26\%)$} & \shortstack[r]{60.75\\ $(-0.39\%)$} & \shortstack[r]{60.85\\ $(-0.23\%)$} \\
FDA & 59.22 & \shortstack[r]{60.22\\ $(+1.69\%)$} & \shortstack[r]{59.67\\ $(+0.77\%)$} & \shortstack[r]{4.36\\ $(-92.64\%)$} & \shortstack[r]{59.58\\ $(+0.61\%)$} & \shortstack[r]{59.49\\ $(+0.46\%)$} & \shortstack[r]{59.58\\ $(+0.61\%)$} \\
SWDE & 45.17 & \shortstack[r]{45.45\\ $(+0.62\%)$} & \shortstack[r]{43.49\\ $(-3.73\%)$} & \shortstack[r]{13.59\\ $(-69.92\%)$} & \shortstack[r]{44.80\\ $(-0.83\%)$} & \shortstack[r]{45.08\\ $(-0.21\%)$} & \shortstack[r]{45.27\\ $(+0.21\%)$} \\
SQuAD & 41.42 & \shortstack[r]{44.41\\ $(+7.22\%)$} & \shortstack[r]{43.84\\ $(+5.84\%)$} & \shortstack[r]{31.44\\ $(-24.09\%)$} & \shortstack[r]{43.40\\ $(+4.79\%)$} & \shortstack[r]{43.53\\ $(+5.11\%)$} & \shortstack[r]{43.10\\ $(+4.06\%)$} \\
NQ & 25.47 & \shortstack[r]{25.34\\ $(-0.50\%)$} & \shortstack[r]{25.47\\ $(+0.00\%)$} & \shortstack[r]{14.86\\ $(-41.67\%)$} & \shortstack[r]{25.31\\ $(-0.62\%)$} & \shortstack[r]{25.37\\ $(-0.37\%)$} & \shortstack[r]{25.28\\ $(-0.75\%)$} \\
NIAH-1 & 82.60 & \shortstack[r]{81.40\\ $(-1.45\%)$} & \shortstack[r]{76.60\\ $(-7.26\%)$} & \shortstack[r]{2.40\\ $(-97.09\%)$} & \shortstack[r]{82.60\\ $(+0.00\%)$} & \shortstack[r]{82.60\\ $(+0.00\%)$} & \shortstack[r]{82.60\\ $(+0.00\%)$} \\
NIAH-2 & 100.00 & \shortstack[r]{100.00\\ $(+0.00\%)$} & \shortstack[r]{100.00\\ $(+0.00\%)$} & \shortstack[r]{0.00\\ $(-100.00\%)$} & \shortstack[r]{100.00\\ $(+0.00\%)$} & \shortstack[r]{100.00\\ $(+0.00\%)$} & \shortstack[r]{100.00\\ $(+0.00\%)$} \\
NIAH-3 & 64.20 & \shortstack[r]{63.80\\ $(-0.62\%)$} & \shortstack[r]{56.80\\ $(-11.53\%)$} & \shortstack[r]{0.00\\ $(-100.00\%)$} & \shortstack[r]{65.40\\ $(+1.87\%)$} & \shortstack[r]{65.00\\ $(+1.25\%)$} & \shortstack[r]{64.80\\ $(+0.93\%)$} \\
\shortstack[l]{WikiText-103\\PPL} & 8.5199 & \shortstack[r]{8.5109\\ $(-0.11\%)$} & \shortstack[r]{8.5187\\ $(-0.01\%)$} & \shortstack[r]{11.8805\\ $(+39.44\%)$} & \shortstack[r]{8.5097\\ $(-0.12\%)$} & \shortstack[r]{8.5094\\ $(-0.12\%)$} & \shortstack[r]{8.5091\\ $(-0.13\%)$} \\
\midrule
Ten-task mean & 62.50 & \shortstack[r]{62.61\\ $(+0.18\%)$} & \shortstack[r]{61.13\\ $(-2.20\%)$} & \shortstack[r]{25.89\\ $(-58.57\%)$} & \shortstack[r]{62.73\\ $(+0.37\%)$} & \shortstack[r]{62.73\\ $(+0.38\%)$} & \shortstack[r]{62.70\\ $(+0.32\%)$} \\
\bottomrule
\end{tabular}
\endgroup
\end{table}
\begin{table}[H]
\centering
\footnotesize
\setlength{\tabcolsep}{3pt}
\caption{\textbf{340M GDN, $6{:}1$.} Accuracy (\%) and PPL; relative changes in parentheses.}
\label{tab:functional-4fa340}
\begingroup
\begin{tabular}{@{}lrrrrrrr@{}}
\toprule
Benchmark & Normal & \multicolumn{3}{c}{PAS suppression} & \multicolumn{3}{c}{Spike connection} \\
\midrule
 &  & $k=1$ & $k=2$ & $k=4$ & $k=1$ & $k=2$ & $k=4$ \\
\midrule
PIQA & 66.59 & \shortstack[r]{64.36\\ $(-3.35\%)$} & \shortstack[r]{62.46\\ $(-6.21\%)$} & \shortstack[r]{62.40\\ $(-6.29\%)$} & \shortstack[r]{66.70\\ $(+0.16\%)$} & \shortstack[r]{66.54\\ $(-0.08\%)$} & \shortstack[r]{66.54\\ $(-0.08\%)$} \\
ARC-Easy & 59.43 & \shortstack[r]{55.81\\ $(-6.09\%)$} & \shortstack[r]{50.21\\ $(-15.51\%)$} & \shortstack[r]{47.98\\ $(-19.26\%)$} & \shortstack[r]{59.13\\ $(-0.50\%)$} & \shortstack[r]{58.71\\ $(-1.20\%)$} & \shortstack[r]{58.84\\ $(-0.99\%)$} \\
HellaSwag & 41.65 & \shortstack[r]{38.24\\ $(-8.18\%)$} & \shortstack[r]{37.22\\ $(-10.62\%)$} & \shortstack[r]{36.39\\ $(-12.63\%)$} & \shortstack[r]{41.70\\ $(+0.14\%)$} & \shortstack[r]{41.67\\ $(+0.07\%)$} & \shortstack[r]{41.72\\ $(+0.19\%)$} \\
FDA & 20.07 & \shortstack[r]{5.99\\ $(-70.14\%)$} & \shortstack[r]{4.54\\ $(-77.38\%)$} & \shortstack[r]{10.45\\ $(-47.96\%)$} & \shortstack[r]{18.80\\ $(-6.33\%)$} & \shortstack[r]{19.26\\ $(-4.07\%)$} & \shortstack[r]{19.35\\ $(-3.62\%)$} \\
SWDE & 23.34 & \shortstack[r]{6.19\\ $(-73.49\%)$} & \shortstack[r]{5.06\\ $(-78.31\%)$} & \shortstack[r]{11.06\\ $(-52.61\%)$} & \shortstack[r]{24.18\\ $(+3.61\%)$} & \shortstack[r]{24.37\\ $(+4.42\%)$} & \shortstack[r]{23.71\\ $(+1.61\%)$} \\
SQuAD & 32.72 & \shortstack[r]{25.56\\ $(-21.87\%)$} & \shortstack[r]{27.31\\ $(-16.53\%)$} & \shortstack[r]{28.69\\ $(-12.32\%)$} & \shortstack[r]{39.17\\ $(+19.71\%)$} & \shortstack[r]{38.80\\ $(+18.58\%)$} & \shortstack[r]{37.55\\ $(+14.78\%)$} \\
NQ & 17.33 & \shortstack[r]{13.18\\ $(-23.95\%)$} & \shortstack[r]{12.39\\ $(-28.52\%)$} & \shortstack[r]{12.73\\ $(-26.51\%)$} & \shortstack[r]{16.98\\ $(-2.01\%)$} & \shortstack[r]{17.04\\ $(-1.65\%)$} & \shortstack[r]{16.85\\ $(-2.74\%)$} \\
NIAH-1 & 40.60 & \shortstack[r]{23.00\\ $(-43.35\%)$} & \shortstack[r]{16.00\\ $(-60.59\%)$} & \shortstack[r]{9.20\\ $(-77.34\%)$} & \shortstack[r]{41.00\\ $(+0.99\%)$} & \shortstack[r]{41.00\\ $(+0.99\%)$} & \shortstack[r]{41.00\\ $(+0.99\%)$} \\
NIAH-2 & 51.00 & \shortstack[r]{21.40\\ $(-58.04\%)$} & \shortstack[r]{14.80\\ $(-70.98\%)$} & \shortstack[r]{11.00\\ $(-78.43\%)$} & \shortstack[r]{53.00\\ $(+3.92\%)$} & \shortstack[r]{52.80\\ $(+3.53\%)$} & \shortstack[r]{52.80\\ $(+3.53\%)$} \\
NIAH-3 & 4.00 & \shortstack[r]{3.20\\ $(-20.00\%)$} & \shortstack[r]{3.20\\ $(-20.00\%)$} & \shortstack[r]{1.40\\ $(-65.00\%)$} & \shortstack[r]{4.00\\ $(+0.00\%)$} & \shortstack[r]{4.20\\ $(+5.00\%)$} & \shortstack[r]{4.00\\ $(+0.00\%)$} \\
\shortstack[l]{WikiText-103\\PPL} & 14.6162 & \shortstack[r]{16.5302\\ $(+13.10\%)$} & \shortstack[r]{16.9186\\ $(+15.75\%)$} & \shortstack[r]{17.2137\\ $(+17.77\%)$} & \shortstack[r]{14.6422\\ $(+0.18\%)$} & \shortstack[r]{14.6407\\ $(+0.17\%)$} & \shortstack[r]{14.6333\\ $(+0.12\%)$} \\
\midrule
Ten-task mean & 35.67 & \shortstack[r]{25.69\\ $(-27.97\%)$} & \shortstack[r]{23.32\\ $(-34.63\%)$} & \shortstack[r]{23.13\\ $(-35.16\%)$} & \shortstack[r]{36.47\\ $(+2.23\%)$} & \shortstack[r]{36.44\\ $(+2.15\%)$} & \shortstack[r]{36.24\\ $(+1.58\%)$} \\
\bottomrule
\end{tabular}
\endgroup
\end{table}
\begin{table}[H]
\centering
\footnotesize
\setlength{\tabcolsep}{3pt}
\caption{\textbf{1.3B GDN, $6{:}1$.} Accuracy (\%) and PPL; relative changes in parentheses.}
\label{tab:functional-4fa13b}
\begingroup
\begin{tabular}{@{}lrrrrrrr@{}}
\toprule
Benchmark & Normal & \multicolumn{3}{c}{PAS suppression} & \multicolumn{3}{c}{Spike connection} \\
\midrule
 &  & $k=1$ & $k=2$ & $k=4$ & $k=1$ & $k=2$ & $k=4$ \\
\midrule
PIQA & 72.80 & \shortstack[r]{70.67\\ $(-2.91\%)$} & \shortstack[r]{64.36\\ $(-11.58\%)$} & \shortstack[r]{70.51\\ $(-3.14\%)$} & \shortstack[r]{72.85\\ $(+0.07\%)$} & \shortstack[r]{72.85\\ $(+0.07\%)$} & \shortstack[r]{72.96\\ $(+0.22\%)$} \\
ARC-Easy & 72.52 & \shortstack[r]{68.94\\ $(-4.93\%)$} & \shortstack[r]{58.96\\ $(-18.69\%)$} & \shortstack[r]{66.25\\ $(-8.65\%)$} & \shortstack[r]{71.13\\ $(-1.92\%)$} & \shortstack[r]{71.13\\ $(-1.92\%)$} & \shortstack[r]{72.05\\ $(-0.64\%)$} \\
HellaSwag & 60.44 & \shortstack[r]{54.22\\ $(-10.28\%)$} & \shortstack[r]{44.89\\ $(-25.72\%)$} & \shortstack[r]{57.48\\ $(-4.89\%)$} & \shortstack[r]{60.17\\ $(-0.44\%)$} & \shortstack[r]{60.17\\ $(-0.44\%)$} & \shortstack[r]{60.36\\ $(-0.13\%)$} \\
FDA & 61.13 & \shortstack[r]{61.49\\ $(+0.59\%)$} & \shortstack[r]{40.96\\ $(-32.99\%)$} & \shortstack[r]{52.59\\ $(-13.97\%)$} & \shortstack[r]{62.40\\ $(+2.08\%)$} & \shortstack[r]{61.94\\ $(+1.34\%)$} & \shortstack[r]{63.22\\ $(+3.42\%)$} \\
SWDE & 48.08 & \shortstack[r]{41.05\\ $(-14.62\%)$} & \shortstack[r]{13.96\\ $(-70.96\%)$} & \shortstack[r]{19.49\\ $(-59.45\%)$} & \shortstack[r]{50.23\\ $(+4.48\%)$} & \shortstack[r]{49.95\\ $(+3.90\%)$} & \shortstack[r]{50.33\\ $(+4.68\%)$} \\
SQuAD & 41.99 & \shortstack[r]{46.83\\ $(+11.52\%)$} & \shortstack[r]{46.09\\ $(+9.76\%)$} & \shortstack[r]{46.83\\ $(+11.52\%)$} & \shortstack[r]{46.09\\ $(+9.76\%)$} & \shortstack[r]{46.32\\ $(+10.32\%)$} & \shortstack[r]{46.12\\ $(+9.84\%)$} \\
NQ & 28.98 & \shortstack[r]{27.87\\ $(-3.83\%)$} & \shortstack[r]{24.11\\ $(-16.83\%)$} & \shortstack[r]{26.10\\ $(-9.95\%)$} & \shortstack[r]{28.67\\ $(-1.09\%)$} & \shortstack[r]{28.57\\ $(-1.42\%)$} & \shortstack[r]{28.95\\ $(-0.11\%)$} \\
NIAH-1 & 95.60 & \shortstack[r]{87.80\\ $(-8.16\%)$} & \shortstack[r]{54.20\\ $(-43.31\%)$} & \shortstack[r]{56.00\\ $(-41.42\%)$} & \shortstack[r]{94.60\\ $(-1.05\%)$} & \shortstack[r]{94.60\\ $(-1.05\%)$} & \shortstack[r]{94.80\\ $(-0.84\%)$} \\
NIAH-2 & 99.40 & \shortstack[r]{97.40\\ $(-2.01\%)$} & \shortstack[r]{70.20\\ $(-29.38\%)$} & \shortstack[r]{88.20\\ $(-11.27\%)$} & \shortstack[r]{99.40\\ $(+0.00\%)$} & \shortstack[r]{99.20\\ $(-0.20\%)$} & \shortstack[r]{99.20\\ $(-0.20\%)$} \\
NIAH-3 & 54.80 & \shortstack[r]{49.20\\ $(-10.22\%)$} & \shortstack[r]{29.40\\ $(-46.35\%)$} & \shortstack[r]{13.00\\ $(-76.28\%)$} & \shortstack[r]{58.20\\ $(+6.20\%)$} & \shortstack[r]{58.00\\ $(+5.84\%)$} & \shortstack[r]{59.80\\ $(+9.12\%)$} \\
\shortstack[l]{WikiText-103\\PPL} & 8.5247 & \shortstack[r]{8.9381\\ $(+4.85\%)$} & \shortstack[r]{12.3677\\ $(+45.08\%)$} & \shortstack[r]{9.3092\\ $(+9.20\%)$} & \shortstack[r]{8.4998\\ $(-0.29\%)$} & \shortstack[r]{8.5004\\ $(-0.28\%)$} & \shortstack[r]{8.4969\\ $(-0.33\%)$} \\
\midrule
Ten-task mean & 63.57 & \shortstack[r]{60.55\\ $(-4.76\%)$} & \shortstack[r]{44.71\\ $(-29.66\%)$} & \shortstack[r]{49.64\\ $(-21.91\%)$} & \shortstack[r]{64.37\\ $(+1.26\%)$} & \shortstack[r]{64.27\\ $(+1.10\%)$} & \shortstack[r]{64.78\\ $(+1.90\%)$} \\
\bottomrule
\end{tabular}
\endgroup
\end{table}
\begin{table}[H]
\centering
\footnotesize
\setlength{\tabcolsep}{3pt}
\caption{\textbf{340M GDN, $3{:}1$.} Accuracy (\%) and PPL; relative changes in parentheses.}
\label{tab:functional-8fa340_official}
\begingroup
\begin{tabular}{@{}lrrrrrrr@{}}
\toprule
Benchmark & Normal & \multicolumn{3}{c}{PAS suppression} & \multicolumn{3}{c}{Spike connection} \\
\midrule
 &  & $k=1$ & $k=2$ & $k=4$ & $k=1$ & $k=2$ & $k=4$ \\
\midrule
PIQA & 67.25 & \shortstack[r]{65.45\\ $(-2.67\%)$} & \shortstack[r]{65.07\\ $(-3.24\%)$} & \shortstack[r]{63.87\\ $(-5.02\%)$} & \shortstack[r]{67.03\\ $(-0.32\%)$} & \shortstack[r]{67.19\\ $(-0.08\%)$} & \shortstack[r]{66.92\\ $(-0.49\%)$} \\
ARC-Easy & 57.91 & \shortstack[r]{55.51\\ $(-4.14\%)$} & \shortstack[r]{54.08\\ $(-6.61\%)$} & \shortstack[r]{52.44\\ $(-9.45\%)$} & \shortstack[r]{57.83\\ $(-0.15\%)$} & \shortstack[r]{58.00\\ $(+0.15\%)$} & \shortstack[r]{58.16\\ $(+0.44\%)$} \\
HellaSwag & 41.18 & \shortstack[r]{40.46\\ $(-1.74\%)$} & \shortstack[r]{38.72\\ $(-5.97\%)$} & \shortstack[r]{39.31\\ $(-4.52\%)$} & \shortstack[r]{41.43\\ $(+0.60\%)$} & \shortstack[r]{41.45\\ $(+0.65\%)$} & \shortstack[r]{41.32\\ $(+0.34\%)$} \\
FDA & 41.24 & \shortstack[r]{7.27\\ $(-82.38\%)$} & \shortstack[r]{1.63\\ $(-96.04\%)$} & \shortstack[r]{0.64\\ $(-98.46\%)$} & \shortstack[r]{43.23\\ $(+4.85\%)$} & \shortstack[r]{43.05\\ $(+4.41\%)$} & \shortstack[r]{42.69\\ $(+3.52\%)$} \\
SWDE & 27.65 & \shortstack[r]{9.75\\ $(-64.75\%)$} & \shortstack[r]{1.31\\ $(-95.25\%)$} & \shortstack[r]{0.84\\ $(-96.95\%)$} & \shortstack[r]{33.36\\ $(+20.68\%)$} & \shortstack[r]{33.46\\ $(+21.02\%)$} & \shortstack[r]{33.93\\ $(+22.71\%)$} \\
SQuAD & 35.77 & \shortstack[r]{32.82\\ $(-8.26\%)$} & \shortstack[r]{18.98\\ $(-46.95\%)$} & \shortstack[r]{8.87\\ $(-75.21\%)$} & \shortstack[r]{41.08\\ $(+14.84\%)$} & \shortstack[r]{40.88\\ $(+14.27\%)$} & \shortstack[r]{44.21\\ $(+23.57\%)$} \\
NQ & 19.13 & \shortstack[r]{14.89\\ $(-22.19\%)$} & \shortstack[r]{7.13\\ $(-62.75\%)$} & \shortstack[r]{3.61\\ $(-81.13\%)$} & \shortstack[r]{18.15\\ $(-5.13\%)$} & \shortstack[r]{18.06\\ $(-5.63\%)$} & \shortstack[r]{18.56\\ $(-2.98\%)$} \\
NIAH-1 & 71.80 & \shortstack[r]{15.20\\ $(-78.83\%)$} & \shortstack[r]{6.00\\ $(-91.64\%)$} & \shortstack[r]{0.00\\ $(-100.00\%)$} & \shortstack[r]{73.00\\ $(+1.67\%)$} & \shortstack[r]{73.20\\ $(+1.95\%)$} & \shortstack[r]{72.80\\ $(+1.39\%)$} \\
NIAH-2 & 82.60 & \shortstack[r]{10.40\\ $(-87.41\%)$} & \shortstack[r]{5.20\\ $(-93.70\%)$} & \shortstack[r]{0.00\\ $(-100.00\%)$} & \shortstack[r]{84.40\\ $(+2.18\%)$} & \shortstack[r]{84.40\\ $(+2.18\%)$} & \shortstack[r]{84.40\\ $(+2.18\%)$} \\
NIAH-3 & 21.40 & \shortstack[r]{0.00\\ $(-100.00\%)$} & \shortstack[r]{0.00\\ $(-100.00\%)$} & \shortstack[r]{0.00\\ $(-100.00\%)$} & \shortstack[r]{21.40\\ $(+0.00\%)$} & \shortstack[r]{21.80\\ $(+1.87\%)$} & \shortstack[r]{21.80\\ $(+1.87\%)$} \\
\shortstack[l]{WikiText-103\\PPL} & 14.6004 & \shortstack[r]{17.3538\\ $(+18.86\%)$} & \shortstack[r]{26.8624\\ $(+83.98\%)$} & \shortstack[r]{51.6760\\ $(+253.94\%)$} & \shortstack[r]{14.6175\\ $(+0.12\%)$} & \shortstack[r]{14.5953\\ $(-0.03\%)$} & \shortstack[r]{14.5937\\ $(-0.05\%)$} \\
\midrule
Ten-task mean & 46.59 & \shortstack[r]{25.17\\ $(-45.97\%)$} & \shortstack[r]{19.81\\ $(-57.48\%)$} & \shortstack[r]{16.96\\ $(-63.60\%)$} & \shortstack[r]{48.09\\ $(+3.22\%)$} & \shortstack[r]{48.15\\ $(+3.34\%)$} & \shortstack[r]{48.48\\ $(+4.05\%)$} \\
\bottomrule
\end{tabular}
\endgroup
\end{table}
\begin{table}[H]
\centering
\footnotesize
\setlength{\tabcolsep}{3pt}
\caption{\textbf{1.3B GDN, $3{:}1$.} Accuracy (\%) and PPL; relative changes in parentheses.}
\label{tab:functional-8fa13b_official}
\begingroup
\begin{tabular}{@{}lrrrrrrr@{}}
\toprule
Benchmark & Normal & \multicolumn{3}{c}{PAS suppression} & \multicolumn{3}{c}{Spike connection} \\
\midrule
 &  & $k=1$ & $k=2$ & $k=4$ & $k=1$ & $k=2$ & $k=4$ \\
\midrule
PIQA & 73.45 & \shortstack[r]{72.96\\ $(-0.67\%)$} & \shortstack[r]{72.74\\ $(-0.96\%)$} & \shortstack[r]{65.94\\ $(-10.22\%)$} & \shortstack[r]{73.34\\ $(-0.15\%)$} & \shortstack[r]{73.50\\ $(+0.07\%)$} & \shortstack[r]{73.56\\ $(+0.15\%)$} \\
ARC-Easy & 72.26 & \shortstack[r]{71.55\\ $(-0.99\%)$} & \shortstack[r]{71.09\\ $(-1.63\%)$} & \shortstack[r]{49.37\\ $(-31.68\%)$} & \shortstack[r]{72.18\\ $(-0.12\%)$} & \shortstack[r]{72.22\\ $(-0.06\%)$} & \shortstack[r]{72.14\\ $(-0.17\%)$} \\
HellaSwag & 60.25 & \shortstack[r]{59.16\\ $(-1.80\%)$} & \shortstack[r]{59.43\\ $(-1.36\%)$} & \shortstack[r]{45.17\\ $(-25.02\%)$} & \shortstack[r]{60.12\\ $(-0.21\%)$} & \shortstack[r]{60.16\\ $(-0.15\%)$} & \shortstack[r]{60.13\\ $(-0.20\%)$} \\
FDA & 67.57 & \shortstack[r]{60.04\\ $(-11.16\%)$} & \shortstack[r]{54.31\\ $(-19.62\%)$} & \shortstack[r]{16.80\\ $(-75.13\%)$} & \shortstack[r]{69.57\\ $(+2.96\%)$} & \shortstack[r]{69.21\\ $(+2.42\%)$} & \shortstack[r]{68.76\\ $(+1.75\%)$} \\
SWDE & 46.95 & \shortstack[r]{44.05\\ $(-6.19\%)$} & \shortstack[r]{42.17\\ $(-10.18\%)$} & \shortstack[r]{38.89\\ $(-17.17\%)$} & \shortstack[r]{50.33\\ $(+7.19\%)$} & \shortstack[r]{50.42\\ $(+7.39\%)$} & \shortstack[r]{50.52\\ $(+7.58\%)$} \\
SQuAD & 42.86 & \shortstack[r]{50.22\\ $(+17.16\%)$} & \shortstack[r]{54.11\\ $(+26.25\%)$} & \shortstack[r]{48.20\\ $(+12.46\%)$} & \shortstack[r]{50.25\\ $(+17.24\%)$} & \shortstack[r]{49.21\\ $(+14.81\%)$} & \shortstack[r]{47.67\\ $(+11.21\%)$} \\
NQ & 28.06 & \shortstack[r]{27.56\\ $(-1.81\%)$} & \shortstack[r]{24.93\\ $(-11.17\%)$} & \shortstack[r]{14.76\\ $(-47.40\%)$} & \shortstack[r]{28.92\\ $(+3.05\%)$} & \shortstack[r]{28.73\\ $(+2.37\%)$} & \shortstack[r]{28.86\\ $(+2.82\%)$} \\
NIAH-1 & 98.60 & \shortstack[r]{99.40\\ $(+0.81\%)$} & \shortstack[r]{99.40\\ $(+0.81\%)$} & \shortstack[r]{35.00\\ $(-64.50\%)$} & \shortstack[r]{98.40\\ $(-0.20\%)$} & \shortstack[r]{98.60\\ $(+0.00\%)$} & \shortstack[r]{98.60\\ $(+0.00\%)$} \\
NIAH-2 & 99.20 & \shortstack[r]{71.00\\ $(-28.43\%)$} & \shortstack[r]{85.40\\ $(-13.91\%)$} & \shortstack[r]{24.40\\ $(-75.40\%)$} & \shortstack[r]{99.20\\ $(+0.00\%)$} & \shortstack[r]{99.20\\ $(+0.00\%)$} & \shortstack[r]{99.20\\ $(+0.00\%)$} \\
NIAH-3 & 65.20 & \shortstack[r]{52.20\\ $(-19.94\%)$} & \shortstack[r]{40.20\\ $(-38.34\%)$} & \shortstack[r]{9.60\\ $(-85.28\%)$} & \shortstack[r]{60.20\\ $(-7.67\%)$} & \shortstack[r]{59.80\\ $(-8.28\%)$} & \shortstack[r]{61.20\\ $(-6.13\%)$} \\
\shortstack[l]{WikiText-103\\PPL} & 9.4945 & \shortstack[r]{9.5874\\ $(+0.98\%)$} & \shortstack[r]{9.5905\\ $(+1.01\%)$} & \shortstack[r]{14.7169\\ $(+55.00\%)$} & \shortstack[r]{9.4952\\ $(+0.01\%)$} & \shortstack[r]{9.4986\\ $(+0.04\%)$} & \shortstack[r]{9.4987\\ $(+0.04\%)$} \\
\midrule
Ten-task mean & 65.44 & \shortstack[r]{60.81\\ $(-7.07\%)$} & \shortstack[r]{60.38\\ $(-7.74\%)$} & \shortstack[r]{34.81\\ $(-46.80\%)$} & \shortstack[r]{66.25\\ $(+1.24\%)$} & \shortstack[r]{66.11\\ $(+1.01\%)$} & \shortstack[r]{66.06\\ $(+0.95\%)$} \\
\bottomrule
\end{tabular}
\endgroup
\end{table}

\subsection{Exploratory Extension to Qwen3.5-27B}
\label{app:open-model-interventions}

We evaluate reference-conditioned Top-4 spike-to-plateau connection on Qwen3.5-27B without updating model weights.
The evaluation covers scientific reasoning, mathematics, and code generation.
Table~\ref{tab:functional-qwen} reports paired comparisons between normal inference and reference-conditioned connection on the same evaluation examples.

\begin{table}[!htbp]
\centering
\small
\caption{\textbf{Qwen3.5-27B with Top-4 spike-to-plateau connection.}
Scores are percentages; changes are relative to normal inference.}
\label{tab:functional-qwen}
\begin{tabular}{@{}lrrr@{}}
\toprule
Benchmark & Normal (\%) & Connect (\%) & Relative $\Delta$ (\%) \\
\midrule
GPQA-Diamond & 77.78 & 78.79 & +1.30 \\
MATH500 & 98.20 & 98.40 & +0.20 \\
LiveCodeBench & 87.11 & 88.09 & +1.12 \\
AIME2025 & 80.00 & 80.00 & +0.00 \\
HumanEval & 96.95 & 98.17 & +1.26 \\
\bottomrule
\end{tabular}
\end{table}

Connection yields small net improvements on GPQA-Diamond, MATH500, LiveCodeBench and HumanEval, with unchanged AIME2025 accuracy.

\FloatBarrier
\begingroup
\color{revisioncolor}
\subsection{Paired Controls and Connection Components}
\label{app:paired-controls}
\rev{We use the two released GDN hybrids with $\rho=3$ (nominally 340M and 1.3B) and hold the prompts, decoding rule and normal baseline fixed across the controls below. The actual parameter counts are 382.6M and 1.399B; their public pretraining budgets are 20B and 100B tokens. As in the original completion protocol, we count a response as correct if its first generated line contains an answer alias, ignoring case. FDA and SWDE are field-extraction tasks; Based SQuAD and NQ are completion tasks. Four-task means give the benchmarks equal weight.}

\paragraph{Complete deletion and replacement.}
\rev{At every FA input, we select the four largest-magnitude coordinates of the first token in the normal pass. We zero these coordinates, replace each with its mean over the normal prompt tokens at that FA input, or zero four coordinates sampled uniformly without replacement from the remaining coordinates. All operations are complete assignments at every FA input. Mean replacement retains an input-dependent contextual average; random deletion matches the coordinate count.}
\begin{table}[H]
\centering
\small
\setlength{\tabcolsep}{5pt}
\caption{\textbf{Complete PAS-coordinate controls.} All scores are percentages. The same normal row is used in every paired-control table.}
\label{tab:paired-deletion}
\begin{tabular}{@{}llrrrr@{}}
\toprule
Model & Benchmark & Normal & Top4 zero & Mean replacement & Random4 zero\\
\midrule
340M & FDA & 38.28\% & 2.34\% & 13.28\% & 39.06\%\\
340M & SWDE & 30.47\% & 2.34\% & 35.94\% & 30.47\%\\
340M & Based SQuAD & 27.34\% & 10.16\% & 29.69\% & 27.34\%\\
340M & Based NQ & 19.53\% & 3.12\% & 8.59\% & 20.31\%\\
\cmidrule(l){2-6}
340M & Four-task mean & 28.91\% & 4.49\% & 21.88\% & 29.30\%\\
\midrule
1.3B & FDA & 64.84\% & 16.41\% & 14.84\% & 65.62\%\\
1.3B & SWDE & 53.12\% & 41.41\% & 18.75\% & 53.12\%\\
1.3B & Based SQuAD & 43.75\% & 46.09\% & 40.62\% & 42.97\%\\
1.3B & Based NQ & 26.56\% & 11.72\% & 13.28\% & 26.56\%\\
\cmidrule(l){2-6}
1.3B & Four-task mean & 47.07\% & 28.91\% & 21.88\% & 47.07\%\\
\bottomrule
\end{tabular}
\end{table}

\rev{Mean replacement lowers the aggregate at both scales, while random-four deletion remains close to normal. Individual tasks need not follow the mean: 1.3B SQuAD improves under zeroing, and 340M SWDE and SQuAD improve under mean replacement. These task-level results are retained in Table~\ref{tab:paired-deletion}.}

\paragraph{Same-rule connection controls.}
\rev{The random-coordinate and non-sink-token controls use the original combined rule: they apply the same Eq.~\ref{eq:main-connection} to their own normal reference trajectories and endpoints. Random coordinates are sampled uniformly without replacement from all hidden dimensions, with the same number as the target's cross-FA coordinate union; the draw is fixed across layers. All four predetermined seeds are averaged within each input. A non-sink token is selected outside the three largest Eq.~(5) attention scores near the prompt midpoint, and uses its own local-top4 union and endpoints. The resulting natural update magnitudes may differ.}
\begin{table}[H]
\centering
\small
\setlength{\tabcolsep}{5pt}
\caption{\textbf{Same-rule target, random-coordinate and non-sink-token connection.} Random denotes the mean over four fixed coordinate draws.}
\label{tab:paired-own-connection}
\begin{tabular}{@{}llrrrr@{}}
\toprule
Model & Benchmark & Normal & Target & Random & Non-sink\\
\midrule
340M & FDA & 38.28\% & 39.06\% & 37.89\% & 39.06\%\\
340M & SWDE & 30.47\% & 38.28\% & 30.66\% & 30.47\%\\
340M & Based SQuAD & 27.34\% & 41.41\% & 27.15\% & 26.56\%\\
340M & Based NQ & 19.53\% & 17.19\% & 19.73\% & 20.31\%\\
\cmidrule(l){2-6}
340M & Four-task mean & 28.91\% & 33.98\% & 28.86\% & 29.10\%\\
\midrule
1.3B & FDA & 64.84\% & 67.97\% & 65.43\% & 64.84\%\\
1.3B & SWDE & 53.12\% & 55.47\% & 53.71\% & 53.12\%\\
1.3B & Based SQuAD & 43.75\% & 49.22\% & 42.97\% & 42.97\%\\
1.3B & Based NQ & 26.56\% & 26.56\% & 26.56\% & 26.56\%\\
\cmidrule(l){2-6}
1.3B & Four-task mean & 47.07\% & 49.80\% & 47.17\% & 46.88\%\\
\bottomrule
\end{tabular}
\end{table}

\rev{Target connection increases four-task mean accuracy by 5.08 and 2.73 percentage points. The paired 95\% source-cluster intervals are [2.50, 7.84] and [0.52, 5.13] points. Target-minus-random differences are 5.13 [2.54, 7.83] and 2.64 [0.61, 4.81] points; target-minus-non-sink differences are 4.88 [2.39, 7.47] and 2.93 [0.76, 5.22] points. Intervals use 10,000 paired source-cluster resamples, with each input's four random seeds first averaged, and do not measure training-seed uncertainty.}

\paragraph{Interior edits versus endpoint restoration.}
\rev{Interior-only connection applies the signed reference floor between FA inputs. Endpoint-only assigns the selected normal reference values at FA inputs only. The combined arm performs both operations. Endpoint-only reproduces every normal generated output. Interior-only already yields the mean retrieval improvement. We therefore adopt interior edits alone in \S\ref{sec:functional-interventions} and Appendix~\ref{app:functional-protocol}; endpoint restoration is retained here as an ablation.}
\begin{table}[H]
\centering
\small
\setlength{\tabcolsep}{5pt}
\caption{\textbf{Connection-component ablation using the same paired normal baseline.}}
\label{tab:paired-factorial}
\begin{tabular}{@{}llrrrr@{}}
\toprule
Model & Benchmark & Normal & Interior only & Endpoint only & Combined\\
\midrule
340M & FDA & 38.28\% & 40.62\% & 38.28\% & 39.06\%\\
340M & SWDE & 30.47\% & 39.84\% & 30.47\% & 38.28\%\\
340M & Based SQuAD & 27.34\% & 40.62\% & 27.34\% & 41.41\%\\
340M & Based NQ & 19.53\% & 14.84\% & 19.53\% & 17.19\%\\
\cmidrule(l){2-6}
340M & Four-task mean & 28.91\% & 33.98\% & 28.91\% & 33.98\%\\
\midrule
1.3B & FDA & 64.84\% & 69.53\% & 64.84\% & 67.97\%\\
1.3B & SWDE & 53.12\% & 56.25\% & 53.12\% & 55.47\%\\
1.3B & Based SQuAD & 43.75\% & 50.00\% & 43.75\% & 49.22\%\\
1.3B & Based NQ & 26.56\% & 26.56\% & 26.56\% & 26.56\%\\
\cmidrule(l){2-6}
1.3B & Four-task mean & 47.07\% & 50.59\% & 47.07\% & 49.80\%\\
\bottomrule
\end{tabular}
\end{table}

\endgroup
\FloatBarrier

\FloatBarrier
\begingroup
\color{revisioncolor}
\subsection{Normalization and Native Magnitude-Matching Controls}
\label{app:normalization-controls}
\rev{We completely zero the reference-selected local top4 at all FA inputs, and compare ordinary normalization with a frozen local denominator. For $R(x)=\sqrt{\|x\|^2/d+\epsilon}$ and a coordinate mask $P$, the two normalized vectors are $\Gamma Px/R(Px)$ and $\Gamma Px/R(x)$. The frozen denominator is the branch's current pre-deletion denominator at each edited site. The implementation preserves the native pre-deletion normalized output exactly on unremoved coordinates and sets the selected outputs to zero.}
\begin{table}[H]
\centering
\small
\setlength{\tabcolsep}{5pt}
\caption{\textbf{Complete deletion with ordinary or frozen local RMS denominator.}}
\label{tab:frozen-rms-control}
\begin{tabular}{@{}llrrr@{}}
\toprule
Model & Benchmark & Normal & Ordinary zero & Frozen RMS zero\\
\midrule
340M & FDA & 38.28\% & 2.34\% & 7.03\%\\
340M & SWDE & 30.47\% & 2.34\% & 29.69\%\\
340M & Based SQuAD & 27.34\% & 10.16\% & 14.84\%\\
340M & Based NQ & 19.53\% & 3.12\% & 9.38\%\\
\cmidrule(l){2-5}
340M & Four-task mean & 28.91\% & 4.49\% & 15.23\%\\
\midrule
1.3B & FDA & 64.84\% & 16.41\% & 10.94\%\\
1.3B & SWDE & 53.12\% & 41.41\% & 19.53\%\\
1.3B & Based SQuAD & 43.75\% & 46.09\% & 37.50\%\\
1.3B & Based NQ & 26.56\% & 11.72\% & 21.09\%\\
\cmidrule(l){2-5}
1.3B & Four-task mean & 47.07\% & 28.91\% & 22.27\%\\
\bottomrule
\end{tabular}
\end{table}

\rev{Freezing the denominator raises the smaller model's mean from 4.49\% to 15.23\%, and lowers the larger model's mean from 28.91\% to 22.27\%. Both remain below normal inference (28.91\% and 47.07\%). Normalization therefore changes the deletion consequence, but its rescaling does not account for the entire performance gap.}

\paragraph{Magnitude, direction and morphology metrics.}
\rev{For $N(x)=\Gamma x/R(x)$ on the domain $R(x)>0$, with diagonal $\Gamma$ and $\epsilon\ge0$, the local derivative is}
\begin{equation}
 D N_x[h]=\Gamma\left(\frac{h}{R(x)}-\frac{x(x^\top h)}{dR(x)^3}\right),\qquad
 D N_x[-\eta x]=-\frac{\eta\epsilon\Gamma x}{R(x)^3}.
\end{equation}
\rev{At $\epsilon=0$, positive uniform scaling leaves the normalized direction unchanged. Complete coordinate deletion generally changes direction as well as norm. SSA describes peak localization and ISR describes relative retention; neither asserts that the normalized representation is fixed. The frozen-denominator arm isolates the immediate shared rescaling at the deletion site, with subsequent computation evaluated normally.}

\paragraph{Native complete-deletion matching.}
\rev{We additionally searched for four non-PAS coordinates with deletion magnitude within 5\% of the selected PAS four, on the same normal token and FA input. No candidate met even the four-dimensional L2 criterion over any of the 4,096 FA events per checkpoint. The median largest-eligible-four to target norm ratios are 0.0718 and 0.1723. Largest-outside coordinates are therefore not labeled magnitude-matched.}

\rev{For two predeclared neighboring layer positions, we also required both operations to be native four-coordinate complete deletions with exactly equal squared norm, evaluated using exact BF16 rational values. No input admitted a matched schedule at all eight intervention sites at either position (0/512 per position and checkpoint). The result concerns these four-coordinate neighboring-site controls; it supplies no matched-control accuracy column. Tables~\ref{tab:paired-deletion} and~\ref{tab:frozen-rms-control} establish dependence on the selected coordinates while keeping this geometric distinction explicit.}

\endgroup
\FloatBarrier

\FloatBarrier
\begingroup
\color{revisioncolor}
\subsection{Perplexity and Commonsense Reasoning Controls}
\label{app:quality-extension}
\rev{We extend the paired interventions to PIQA validation, ARC-Easy test and WikiText-103 continuation perplexity, using both public GDN $\rho=3$ checkpoints. The intervention is applied to the prompt-context prefill; each answer continuation is then evaluated with its own native cache. The reference, channel means, token selection and connection schedule use only the context preceding that continuation. For PIQA and ARC-Easy, we choose the answer with the highest total log-likelihood and report accuracy. WikiText-103 reports token-weighted perplexity on disjoint continuations after 1,024 context tokens.}

\rev{The same normal and target conditions are reused across the three tables below. Random-coordinate connection averages four fixed draws within each input; for perplexity, mean negative log-likelihood is averaged before exponentiation.}
\begin{table}[H]
\centering
\small
\setlength{\tabcolsep}{5pt}
\caption{\textbf{Complete deletion and replacement.} Accuracy is reported as a percentage. Lower continuation perplexity is better.}
\label{tab:quality-deletion}
\begin{tabular}{@{}llrrr@{}}
\toprule
Model & Condition & PIQA (\%) & ARC-Easy (\%) & WikiText-103 PPL\\
\midrule
340M & Normal & 67.41 & 58.29 & 14.61\\
340M & PAS top4 zero & 63.93 & 52.74 & 51.97\\
340M & Channel mean & 63.60 & 54.50 & 19.96\\
340M & Random4 zero & 67.41 & 58.04 & 14.61\\
\midrule
1.3B & Normal & 73.34 & 72.26 & 9.50\\
1.3B & PAS top4 zero & 66.43 & 49.41 & 14.73\\
1.3B & Channel mean & 68.44 & 61.99 & 14.88\\
1.3B & Random4 zero & 73.34 & 72.26 & 9.50\\
\bottomrule
\end{tabular}
\end{table}
\begin{table}[H]
\centering
\small
\setlength{\tabcolsep}{5pt}
\caption{\textbf{Same-rule connection controls.} Accuracy is reported as a percentage. Lower continuation perplexity is better.}
\label{tab:quality-own}
\begin{tabular}{@{}llrrr@{}}
\toprule
Model & Condition & PIQA (\%) & ARC-Easy (\%) & WikiText-103 PPL\\
\midrule
340M & Normal & 67.41 & 58.29 & 14.61\\
340M & Target connection & 67.63 & 58.38 & 14.60\\
340M & Random own connection & 67.37 & 58.07 & 14.64\\
340M & Non-sink own connection & 67.36 & 58.00 & 14.61\\
\midrule
1.3B & Normal & 73.34 & 72.26 & 9.50\\
1.3B & Target connection & 73.34 & 72.31 & 9.50\\
1.3B & Random own connection & 73.31 & 72.30 & 9.50\\
1.3B & Non-sink own connection & 73.29 & 72.18 & 9.50\\
\bottomrule
\end{tabular}
\end{table}
\begin{table}[H]
\centering
\small
\setlength{\tabcolsep}{5pt}
\caption{\textbf{Connection-component ablation.} Accuracy is reported as a percentage. Lower continuation perplexity is better.}
\label{tab:quality-factorial}
\begin{tabular}{@{}llrrr@{}}
\toprule
Model & Condition & PIQA (\%) & ARC-Easy (\%) & WikiText-103 PPL\\
\midrule
340M & Normal & 67.41 & 58.29 & 14.61\\
340M & Interior only & 67.19 & 58.21 & 14.60\\
340M & Endpoint only & 67.41 & 58.29 & 14.61\\
340M & Combined & 67.63 & 58.38 & 14.60\\
\midrule
1.3B & Normal & 73.34 & 72.26 & 9.50\\
1.3B & Interior only & 73.18 & 72.26 & 9.49\\
1.3B & Endpoint only & 73.34 & 72.26 & 9.50\\
1.3B & Combined & 73.34 & 72.31 & 9.50\\
\bottomrule
\end{tabular}
\end{table}
\rev{Complete PAS-coordinate deletion lowers both commonsense scores and increases continuation perplexity at both scales. Channel-mean replacement also lowers commonsense accuracy and raises perplexity, while random-four deletion remains close to normal. Target connection and its component ablations produce much smaller changes on these tasks than on the retrieval benchmarks. Thus the retrieval gains do not imply a corresponding general-purpose accuracy gain.}
\endgroup
\FloatBarrier

\FloatBarrier
\Needspace{15\baselineskip}
\begingroup
\color{revisioncolor}
\subsection{Prepared-Plan Prefill and Generation Latency}
\label{app:prepared-cost}
\rev{We time batch-one inference on an A800 with the prepared schedule of the original combined rule (Eq.~\ref{eq:main-connection} plus endpoint restoration), comparing normal inference with direct edited prefill and the subsequent native cached replay. Both paths replay identical tokens. Prompt lengths range from 58 to 2005 tokens; the output lengths below are 512 and 1,024 predictions, including the first prediction from prefill (511 and 1,023 cached decoding calls). The model configuration and cache precision are unchanged.}

\rev{The timed prefill includes the actual edit assignments. Prefill-plus-decode is continuous synchronized wall time through the final decoding step. Reference capture, CPU schedule construction, tensor preparation and warmup precede this timing boundary. For a request that requires a new schedule, total latency also includes the reference pass and schedule preparation.}
\begin{table}[H]
\centering
\small
\setlength{\tabcolsep}{5pt}
\caption{\textbf{Prepared-plan latency: normal $\to$ connection.} Prefill is in milliseconds; prefill plus decode is in seconds.}
\label{tab:prepared-cost}
\begin{tabular}{@{}lrrr@{}}
\toprule
Model & Output predictions & Prefill (ms) & Prefill + decode (s)\\
\midrule
340M & 512 & 81.51 $\to$ 82.38 & 19.368 $\to$ 19.382\\
340M & 1024 & 82.58 $\to$ 83.98 & 38.989 $\to$ 38.869\\
1.3B & 512 & 90.76 $\to$ 91.14 & 19.902 $\to$ 19.866\\
1.3B & 1024 & 91.63 $\to$ 92.23 & 39.748 $\to$ 39.793\\
\bottomrule
\end{tabular}
\end{table}
\rev{For each fixed prompt template we take the median of three paired repetitions, then report the median across templates. All length conditions and repetitions are retained. These descriptive timings use eight prompt templates; repetitions are not treated as independent inputs.}
\endgroup
\FloatBarrier

\clearpage

\newcommand{\tokenmorphologyplot}[4]{%
\begin{subfigure}[t]{\textwidth}
    \centering
    \includegraphics[
        width=\textwidth,
        height=0.18\textheight,
        keepaspectratio
    ]{#1/#2.pdf}
    \vspace{-3mm}
    \caption{#3}
    \label{#4}
\end{subfigure}
}

\begin{figure*}[p]
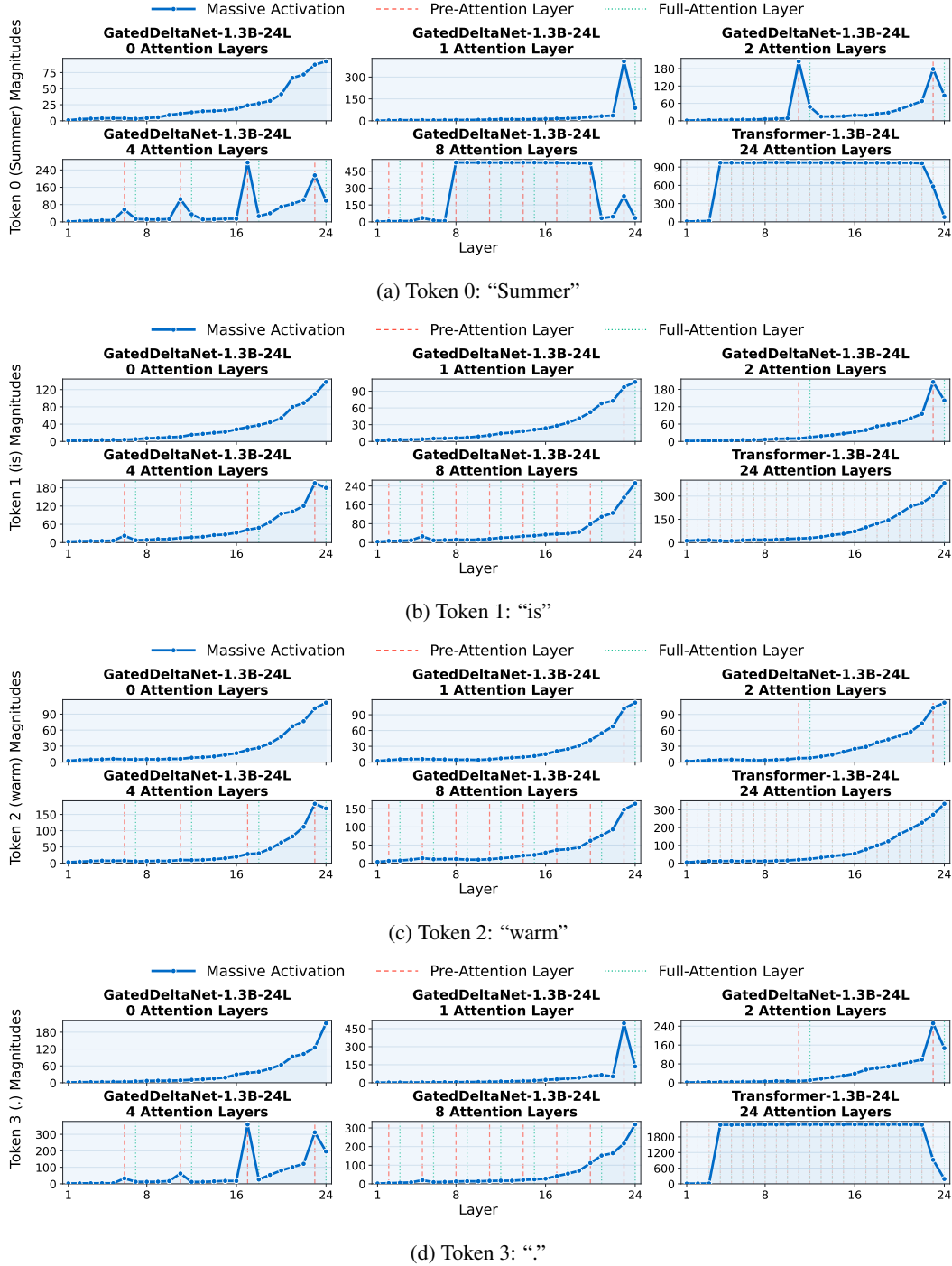

    \centering

    \tokenmorphologyplot
    {figures/appendix/map/token/gdn}
    {token_000_summer}
    {Token 0: ``Summer''}
    {fig:gdn-token-summer}

    \vspace{0.15em}

    \tokenmorphologyplot
    {figures/appendix/map/token/gdn}
    {token_001_is}
    {Token 1: ``is''}
    {fig:gdn-token-is-first}

    \vspace{0.15em}

    \tokenmorphologyplot
    {figures/appendix/map/token/gdn}
    {token_002_warm}
    {Token 2: ``warm''}
    {fig:gdn-token-warm}

    \vspace{0.15em}

    \tokenmorphologyplot
    {figures/appendix/map/token/gdn}
    {token_003_period}
    {Token 3: ``.''}
    {fig:gdn-token-period-first}

    \caption{
        \textbf{MA dynamics across token positions in GDN.}
        Each panel traces the maximum absolute hidden-state activation of one token in \textit{``Summer is warm. Winter is cold.''} across pure linear attention, the $24{:}1$, $12{:}1$, $6{:}1$, and $3{:}1$ hybrid configurations, and full attention.
        The attention-sink positions, ``Summer'' and the first period, exhibit pronounced PAS and increasingly connected ISP as full attention becomes denser, whereas non-sink positions exhibit neither morphology.
        This figure extends the first-token analysis in \S\ref{sec:architecture-dependence}; detailed token-wise discussion is provided in Appendix~\ref{app:map-token-dynamics}.
    }
    \label{fig:gdn-token-dynamics}
\end{figure*}

\begin{figure*}[p]
    \ContinuedFloat
    \centering

    \tokenmorphologyplot
    {figures/appendix/map/token/gdn}
    {token_004_winter}
    {Token 4: ``Winter''}
    {fig:gdn-token-winter}

    \vspace{0.15em}

    \tokenmorphologyplot
    {figures/appendix/map/token/gdn}
    {token_005_is}
    {Token 5: ``is''}
    {fig:gdn-token-is-second}

    \vspace{0.15em}

    \tokenmorphologyplot
    {figures/appendix/map/token/gdn}
    {token_006_cold}
    {Token 6: ``cold''}
    {fig:gdn-token-cold}

    \vspace{0.15em}

    \tokenmorphologyplot
    {figures/appendix/map/token/gdn}
    {token_007_period}
    {Token 7: ``.''}
    {fig:gdn-token-period-second}

    \caption[]{
        \textbf{MA dynamics across token positions in GDN (continued).}
    }
\end{figure*}
\clearpage

\newcommand{\mapdomainplot}[3]{%
\begin{subfigure}[t]{\textwidth}
    \centering
    \includegraphics[
        width=\textwidth,
        height=0.22\textheight,
        keepaspectratio
    ]{#1}
    \vspace{-4mm}
    \caption{#2}
    \label{#3}
\end{subfigure}
}

\begin{figure*}[p]
    \centering

    \mapdomainplot
    {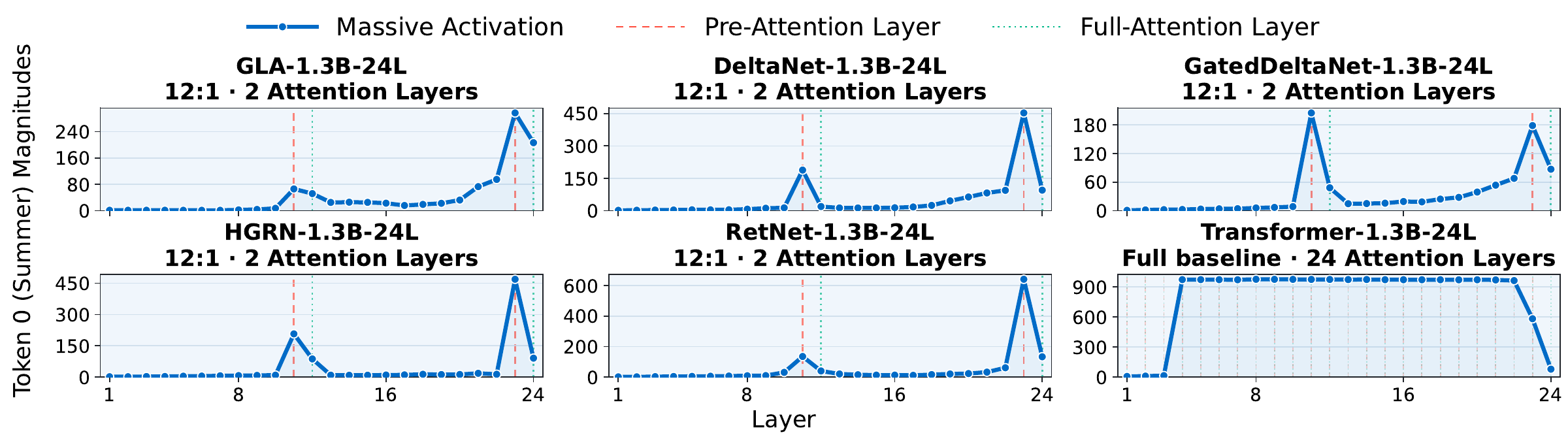}
    {Running example: \textit{``Summer is warm. Winter is cold.''}}
    {fig:map-domain-summer}

    \vspace{0.2em}

    \mapdomainplot
    {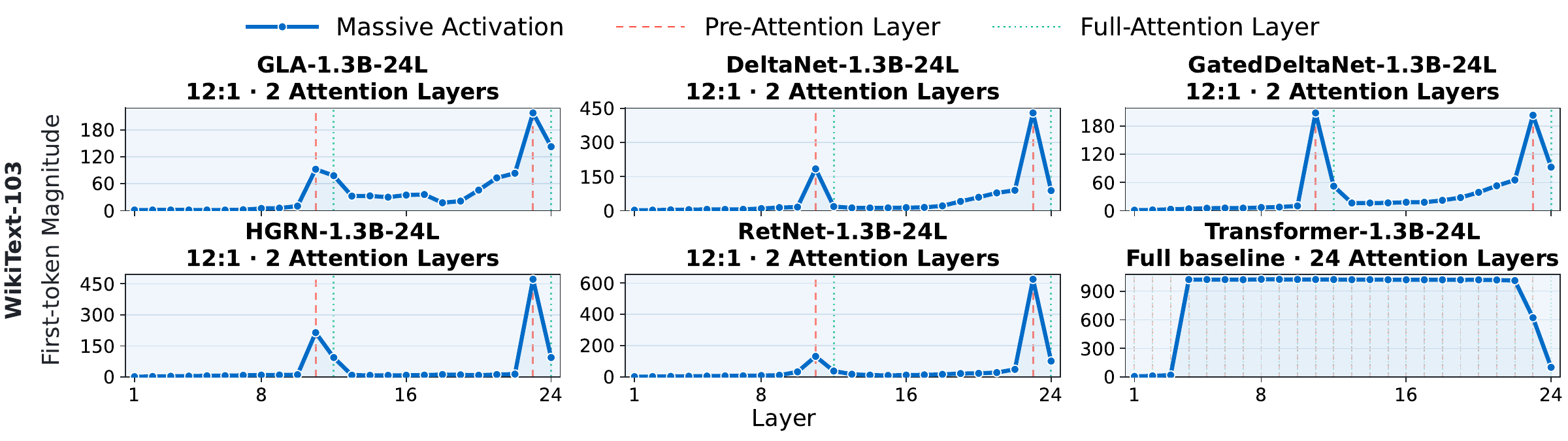}
    {General prose from WikiText-103.}
    {fig:map-domain-wikitext}

    \vspace{0.2em}

    \mapdomainplot
    {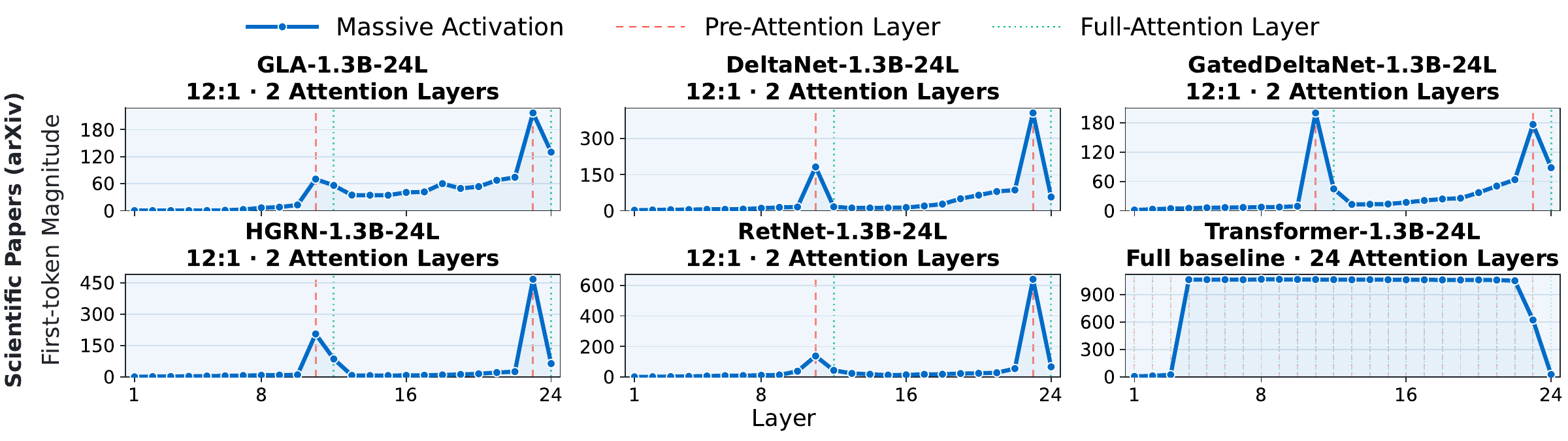}
    {Scientific writing from Scientific Papers.}
    {fig:map-domain-scientific-papers}
    
    \caption{
        \textbf{MA dynamics across input domains in the M-A-P model suite.}
        Each panel traces the maximum absolute hidden-state activation of the first token across depth for five 1.3B HLA architectures under a fixed $12{:}1$ hybridization ratio, with a full attention Transformer included as a reference.
        Across the running example and five representative domain-specific inputs, PAS consistently emerge immediately before full attention layers.
        This figure extends the cross-domain analysis in \S\ref{sec:architecture-dependence}; detailed discussion is provided in Appendix~\ref{app:map-domain-dynamics}.
    }
\label{fig:map-cross-domain-dynamics}
\end{figure*}

\begin{figure*}[p]
    \ContinuedFloat
    \centering

    \mapdomainplot
    {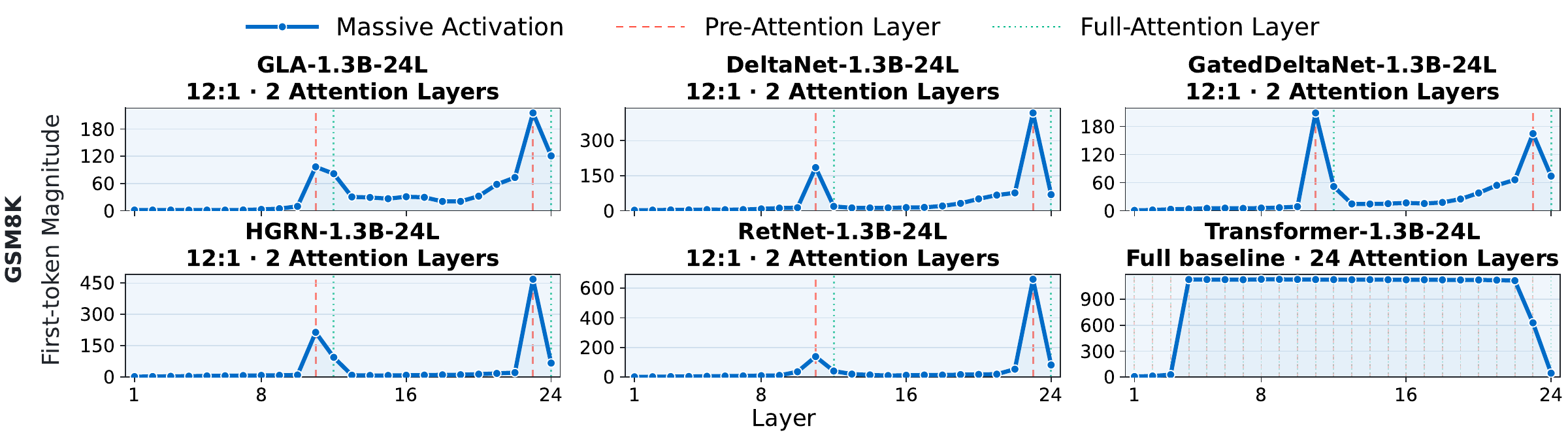}
    {Mathematical reasoning from GSM8K.}
    {fig:map-domain-gsm8k}

    \vspace{0.2em}

    \mapdomainplot
    {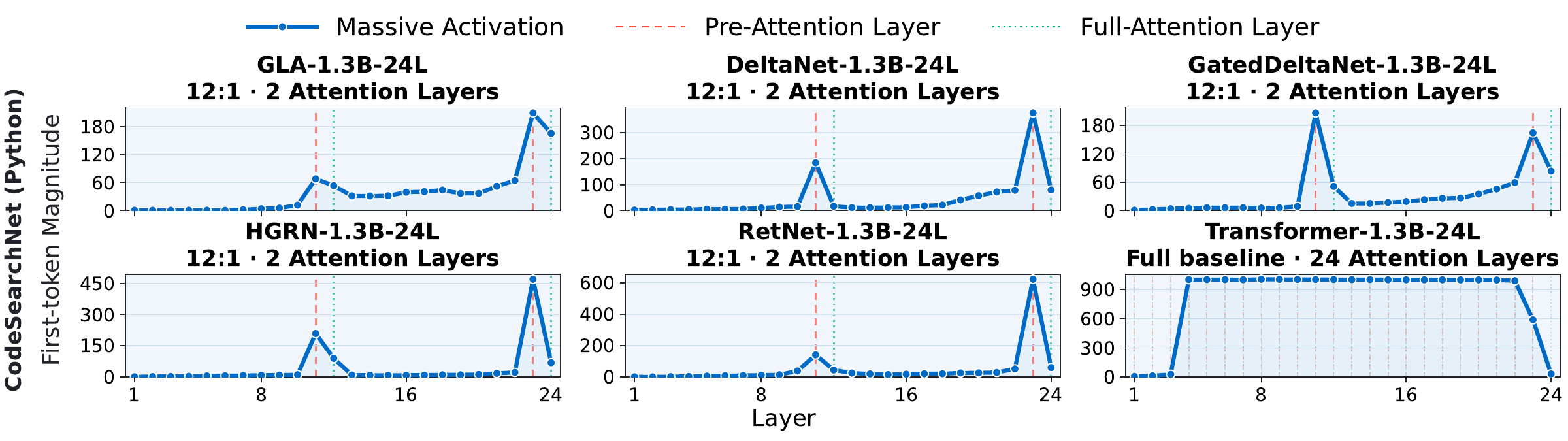}
    {Python code from CodeSearchNet.}
    {fig:map-domain-codesearchnet}

    \vspace{0.2em}

    \mapdomainplot
    {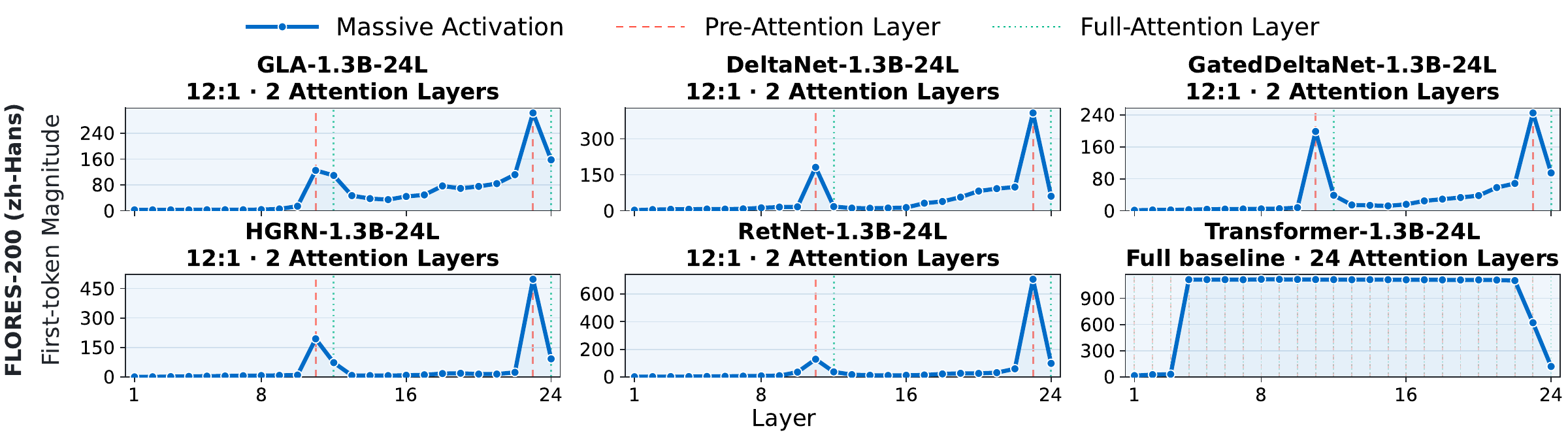}
    {Multilingual text from FLORES-200.}
    {fig:map-domain-flores}

    \caption[]{
        \textbf{MA dynamics across input domains in the M-A-P model suite (continued).}
    }
\end{figure*}
\clearpage

\newcommand{\ratiomorphologyplot}[4]{%
\begin{subfigure}[t]{\textwidth}
    \centering
    \includegraphics[
        width=\textwidth,
        height=0.18\textheight,
        keepaspectratio
    ]{#1/#2.pdf}
    \vspace{-3mm}
    \caption{#3}
    \label{#4}
\end{subfigure}
}

\begin{figure*}[p]
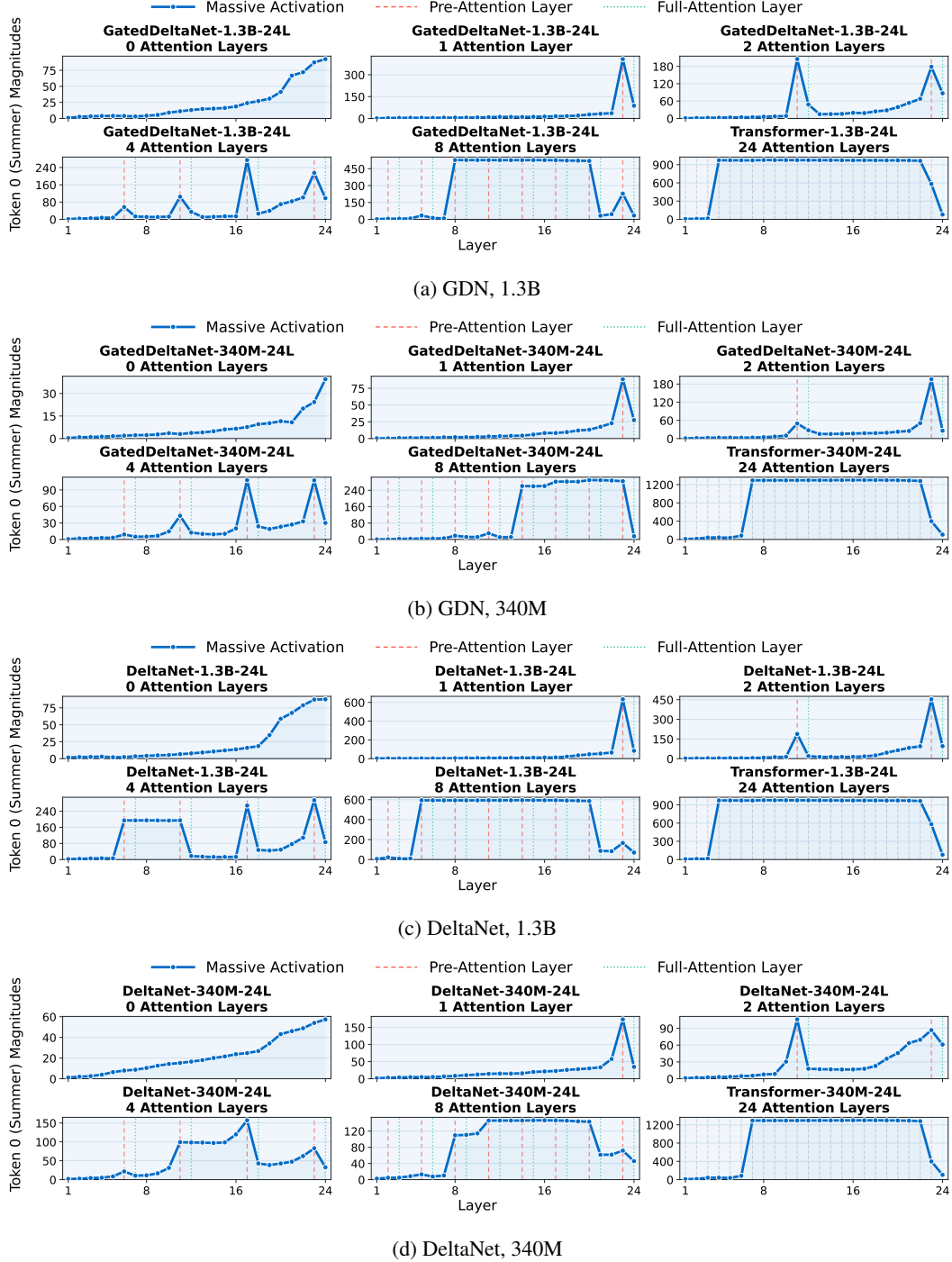

    \centering

    \ratiomorphologyplot
    {figures/appendix/map/la/1.3B}
    {GatedDeltaNet}
    {GDN, 1.3B}
    {fig:ratio-gdn-1p3b}

    \vspace{0.15em}

    \ratiomorphologyplot
    {figures/appendix/map/la/340M}
    {GatedDeltaNet}
    {GDN, 340M}
    {fig:ratio-gdn-340m}

    \vspace{0.15em}

    \ratiomorphologyplot
    {figures/appendix/map/la/1.3B}
    {DeltaNet}
    {DeltaNet, 1.3B}
    {fig:ratio-deltanet-1p3b}

    \vspace{0.15em}

    \ratiomorphologyplot
    {figures/appendix/map/la/340M}
    {DeltaNet}
    {DeltaNet, 340M}
    {fig:ratio-deltanet-340m}

    \caption{
        \textbf{MA dynamics across hybridization ratios in the M-A-P model suite.}
        Each panel traces the maximum absolute hidden-state activation of the first token across pure linear attention, the $24{:}1$, $12{:}1$, $6{:}1$, and $3{:}1$ hybrid configurations, and full attention.
        Across architectures and model scales, isolated PAS become progressively connected through ISP as full attention becomes denser.
        Further discussion appears in \S\ref{sec:ratio-dependence} and Appendix~\ref{app:map-ma-dynamics}.
    }
    \label{fig:ratio-morphologies}
\end{figure*}

\begin{figure*}[p]
    \ContinuedFloat
    \centering

    \ratiomorphologyplot
    {figures/appendix/map/la/1.3B}
    {GLA}
    {GLA, 1.3B}
    {fig:ratio-gla-1p3b}

    \vspace{0.15em}

    \ratiomorphologyplot
    {figures/appendix/map/la/340M}
    {GLA}
    {GLA, 340M}
    {fig:ratio-gla-340m}

    \vspace{0.15em}

    \ratiomorphologyplot
    {figures/appendix/map/la/1.3B}
    {HGRN}
    {HGRN, 1.3B}
    {fig:ratio-hgrn-1p3b}

    \vspace{0.15em}

    \ratiomorphologyplot
    {figures/appendix/map/la/340M}
    {HGRN}
    {HGRN, 340M}
    {fig:ratio-hgrn-340m}

    \caption[]{
        \textbf{MA dynamics across hybridization ratios in the M-A-P model suite (continued).}
    }
\end{figure*}

\begin{figure*}[p]
    \ContinuedFloat
    \centering

    \ratiomorphologyplot
    {figures/appendix/map/la/1.3B}
    {RetNet}
    {RetNet, 1.3B}
    {fig:ratio-retnet-1p3b}

    \vspace{0.4em}

    \ratiomorphologyplot
    {figures/appendix/map/la/340M}
    {RetNet}
    {RetNet, 340M}
    {fig:ratio-retnet-340m}

    \caption[]{
        \textbf{MA dynamics across hybridization ratios in the M-A-P model suite (continued).}
    }
\end{figure*}

\clearpage

\newcommand{\largemodelpanel}[4]{%
\begin{subfigure}[t]{\textwidth}
    \centering
    \includegraphics[
        width=\textwidth,
        height=#4\textheight,
        keepaspectratio
    ]{#1}
    \vspace{-3mm}
    \caption{#2}
    \label{#3}
\end{subfigure}
}

\begin{figure*}[p]
    \centering

    \largemodelpanel
    {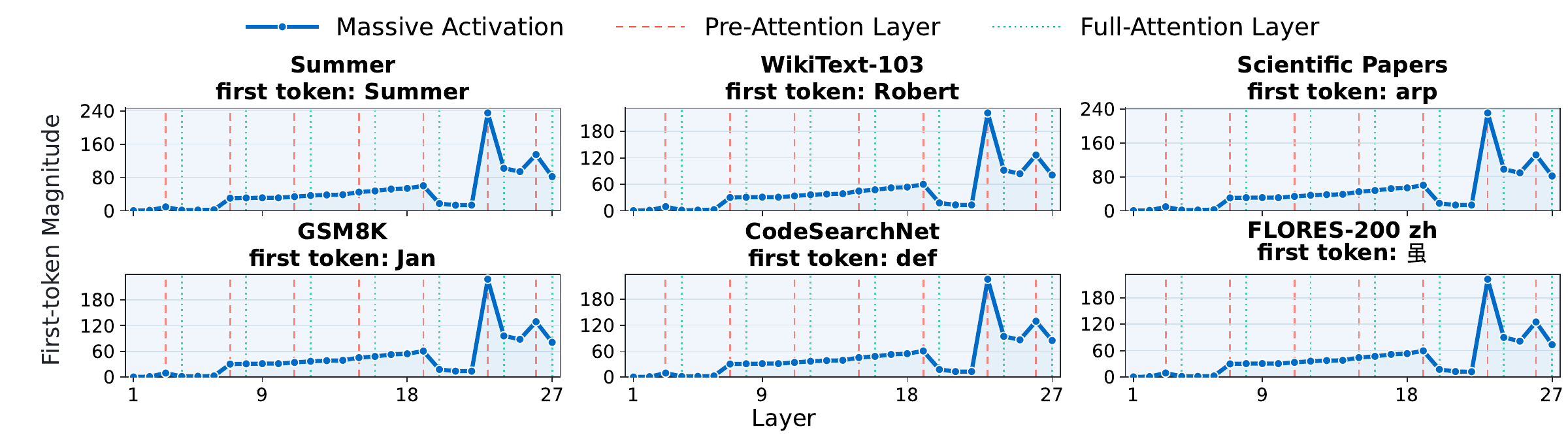}
    {Kimi-Linear-48B-A3B-Base}
    {fig:kimi-linear-base}
    {0.28}

    \vspace{0.3em}

    \largemodelpanel
    {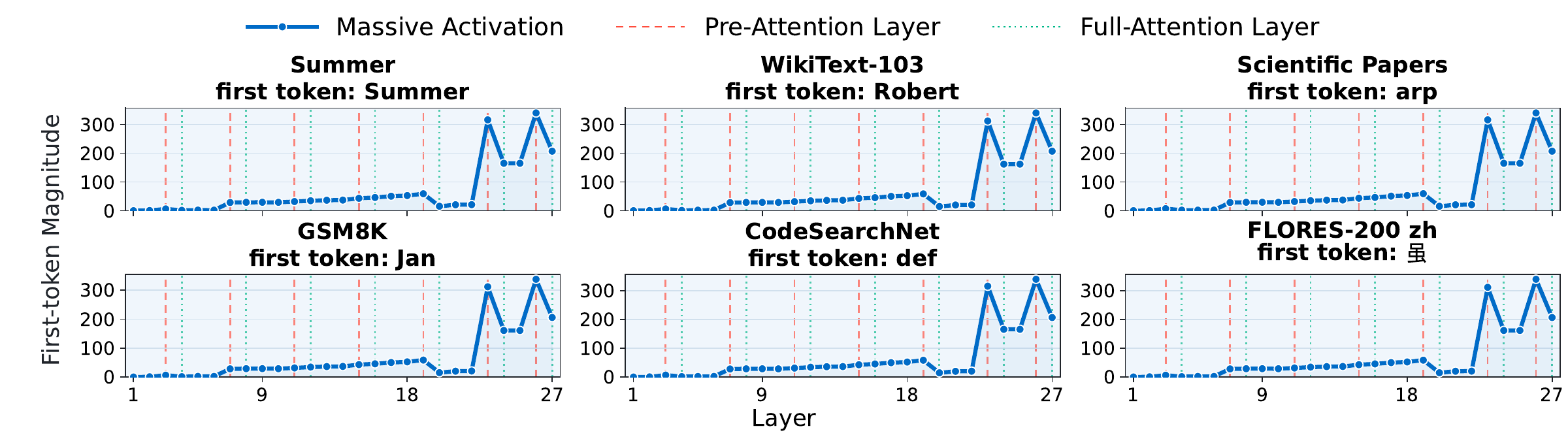}
    {Kimi-Linear-48B-A3B-Instruct}
    {fig:kimi-linear-instruct}
    {0.28}

    \caption{
        \textbf{MA dynamics in large-scale pretrained hybrid models.}
        Each checkpoint panel presents first-token activation trajectories for the running example and five domain-specific inputs.
        Results are grouped by family to compare matched base/instruction-tuned checkpoints, parameter scales, and linear attention versus state-space backbones.
        PAS and ISP remain aligned with the hybrid layer arrangement across these settings, while their absolute magnitudes vary across checkpoints and inputs.
        Detailed cross-model analysis is provided in Appendix~\ref{app:large-scale-ma-dynamics}.
    }
    \label{fig:large-scale-hybrid-ma-dynamics}
\end{figure*}

\begin{figure*}[p]
    \ContinuedFloat
    \centering

    \largemodelpanel
    {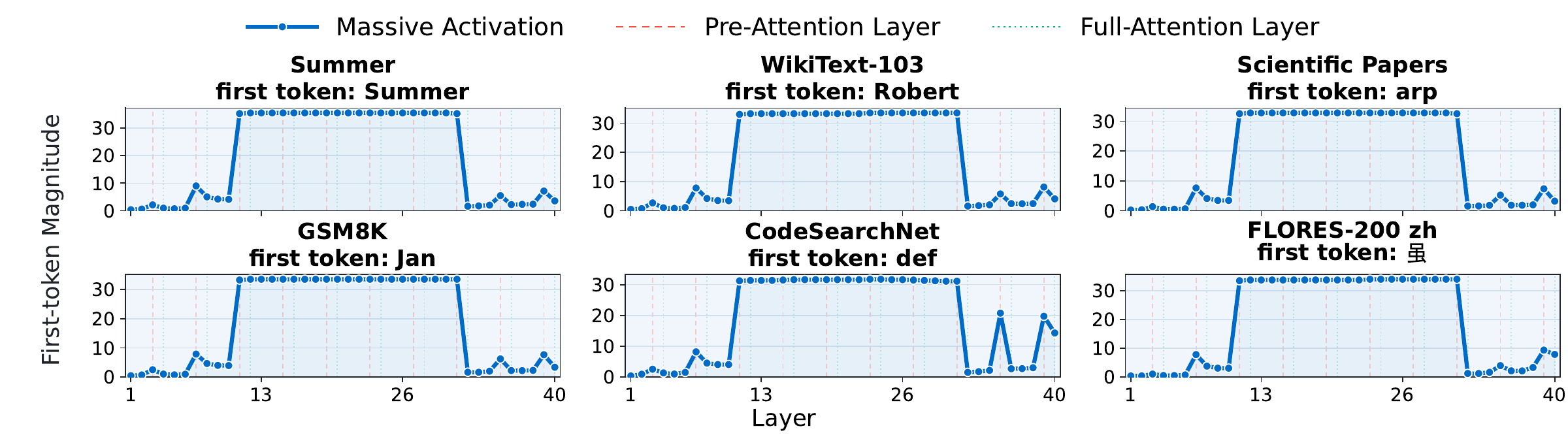}
    {Qwen3.5-35B-A3B-Base}
    {fig:qwen35-35b-base}
    {0.28}

    \vspace{0.3em}

    \largemodelpanel
    {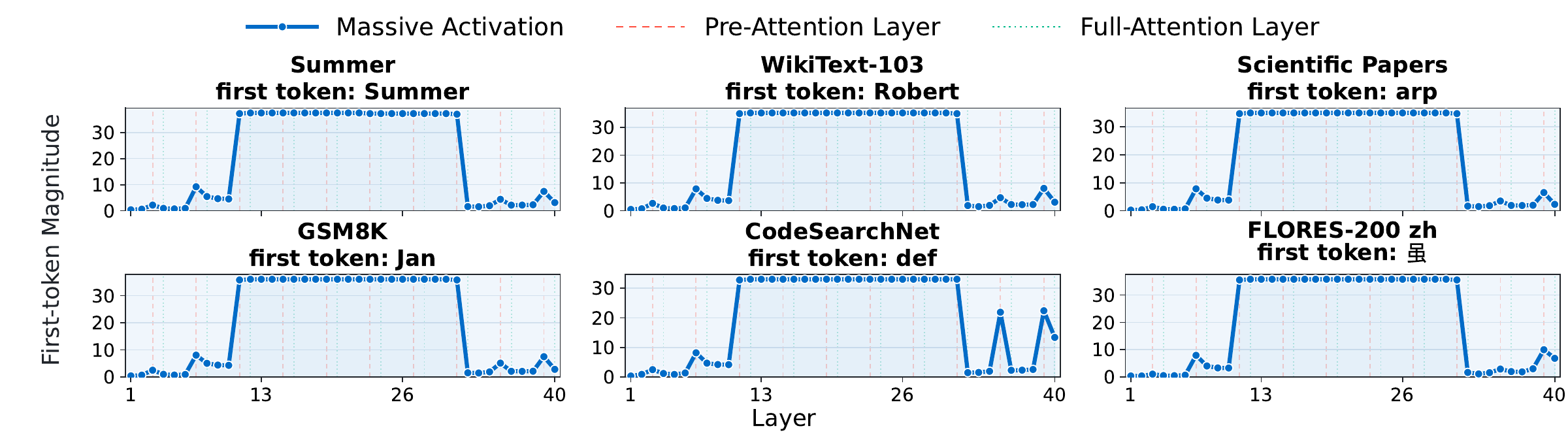}
    {Qwen3.5-35B-A3B}
    {fig:qwen35-35b}
    {0.28}

    \caption[]{
        \textbf{MA dynamics in large-scale pretrained hybrid models (continued).}
        The matched Qwen3.5-35B Base and instruction-tuned checkpoints exhibit similar PAS locations and plateau spans, despite differences in absolute activation magnitude.
    }
\end{figure*}

\begin{figure*}[p]
    \ContinuedFloat
    \centering

    \largemodelpanel
    {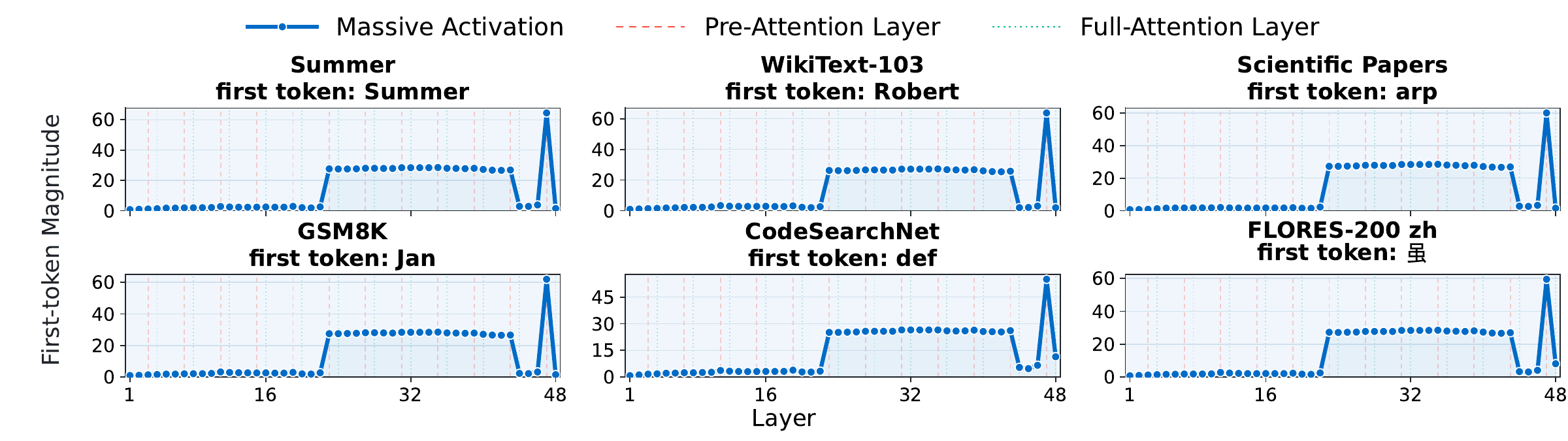}
    {Qwen3.5-122B-A10B}
    {fig:qwen35-122b}
    {0.28}

    \vspace{0.3em}

    \largemodelpanel
    {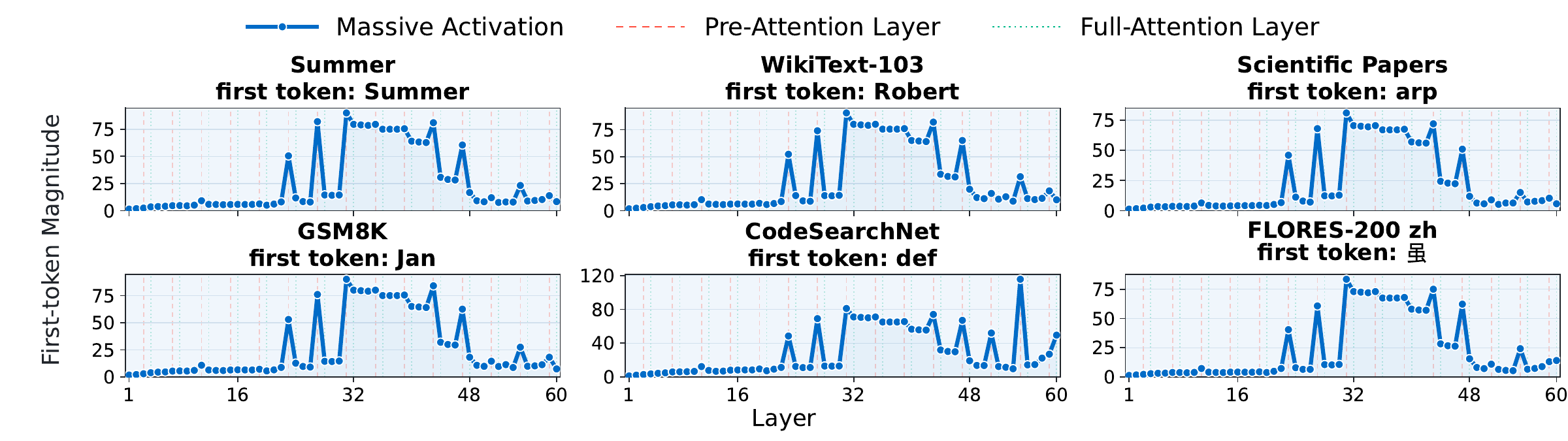}
    {Qwen3.5-397B-A17B}
    {fig:qwen35-397b}
    {0.28}

    \caption[]{
        \textbf{MA dynamics in large-scale pretrained hybrid models (continued).}
        Qwen3.5 checkpoints at the 122B and 397B scales preserve the architecture-aligned PAS--ISP organization, although their activation magnitudes and plateau profiles differ.
    }
\end{figure*}

\begin{figure*}[p]
    \ContinuedFloat
    \centering

    \largemodelpanel
    {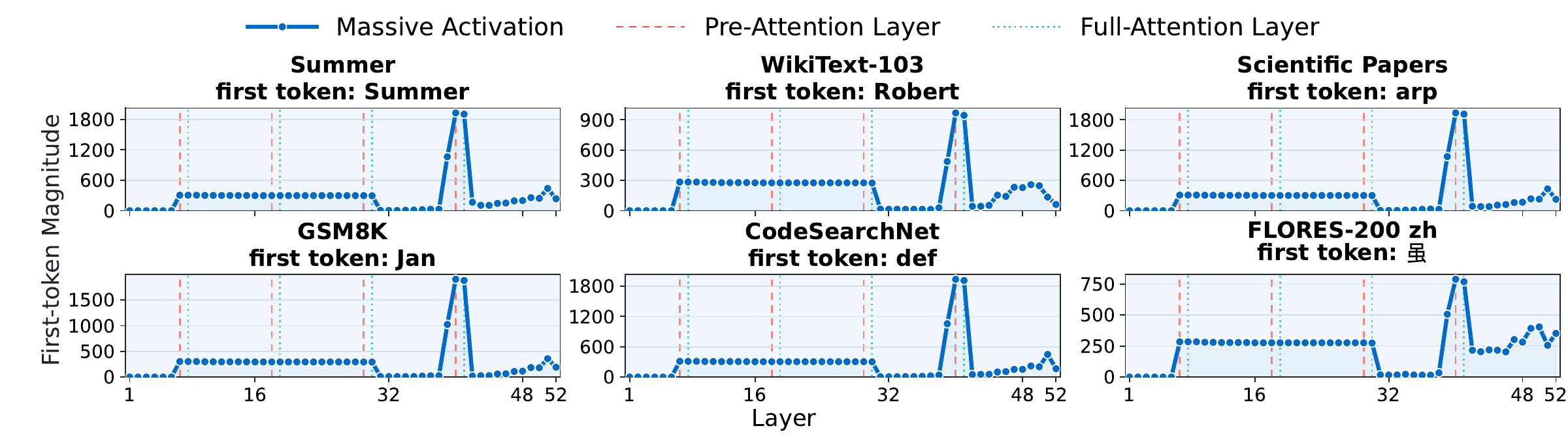}
    {Nemotron-H-8B-Base-8K}
    {fig:nemotron-h-8b}
    {0.21}

    \vspace{0.2em}

    \largemodelpanel
    {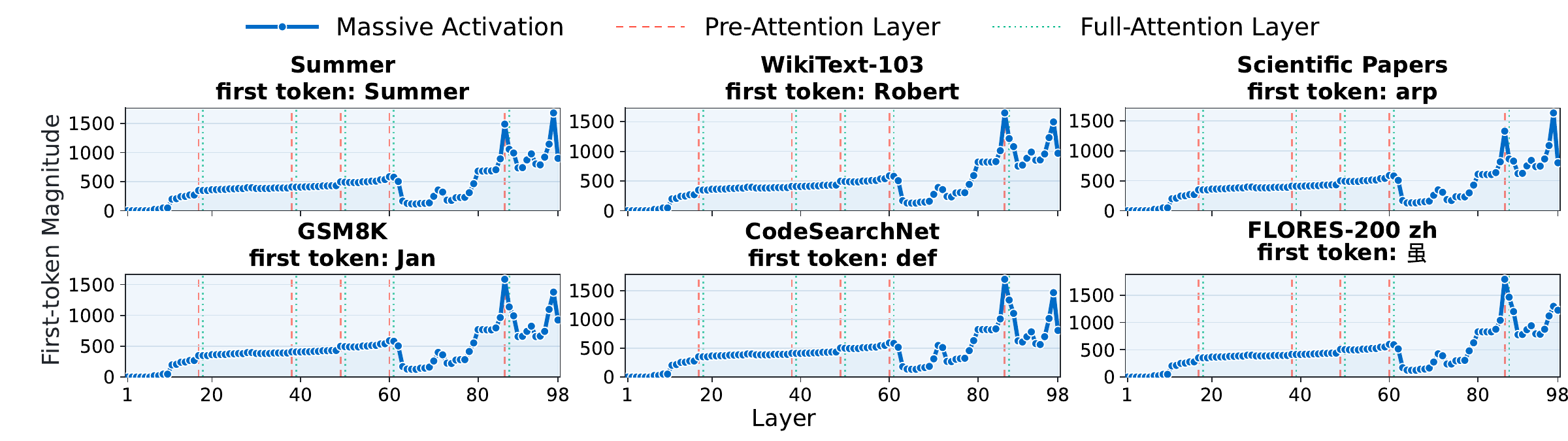}
    {Nemotron-H-47B-Base-8K}
    {fig:nemotron-h-47b}
    {0.21}

    \vspace{0.2em}

    \largemodelpanel
    {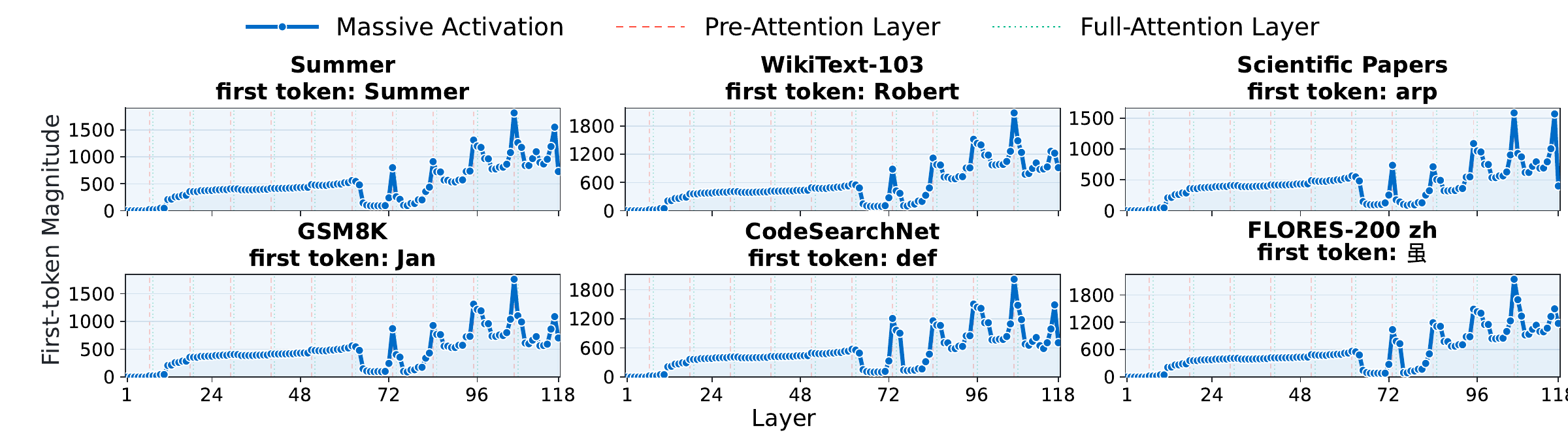}
    {Nemotron-H-56B-Base-8K}
    {fig:nemotron-h-56b}
    {0.21}

    \caption[]{
        \textbf{MA dynamics in large-scale pretrained hybrid models (continued).}
        Nemotron-H checkpoints at the 8B, 47B, and 56B scales exhibit PAS- and ISP-like activation patterns aligned with the placement of full attention layers, extending the observed organization to state-space hybrids.
    }
\end{figure*}

\begin{figure*}[p]
    \ContinuedFloat
    \centering

    \largemodelpanel
    {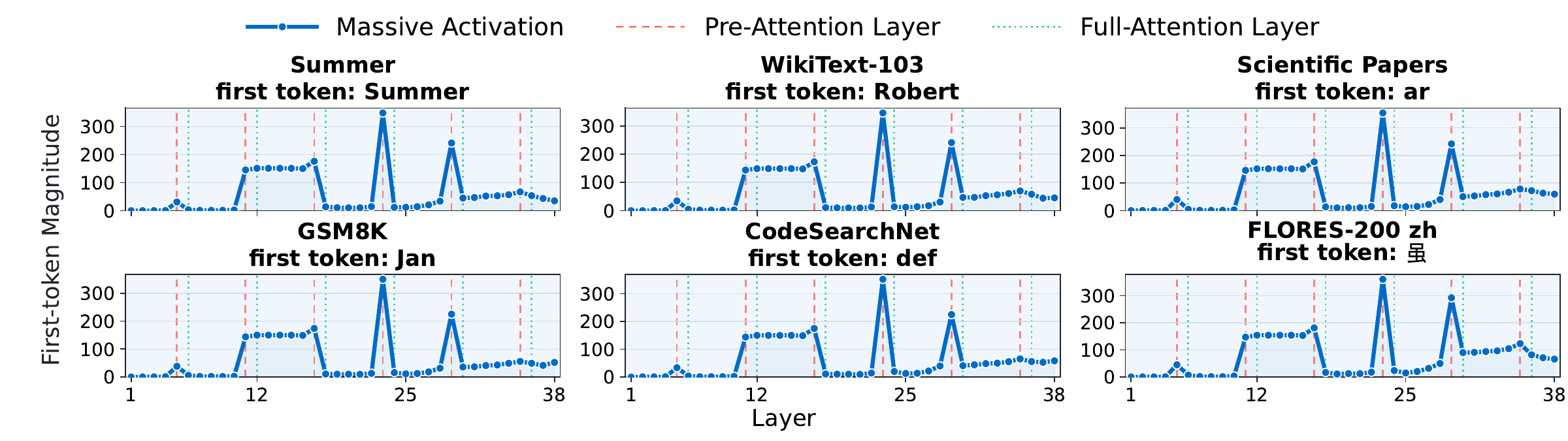}
    {Zamba2-1.2B}
    {fig:zamba2-1p2b}
    {0.21}

    \vspace{0.2em}

    \largemodelpanel
    {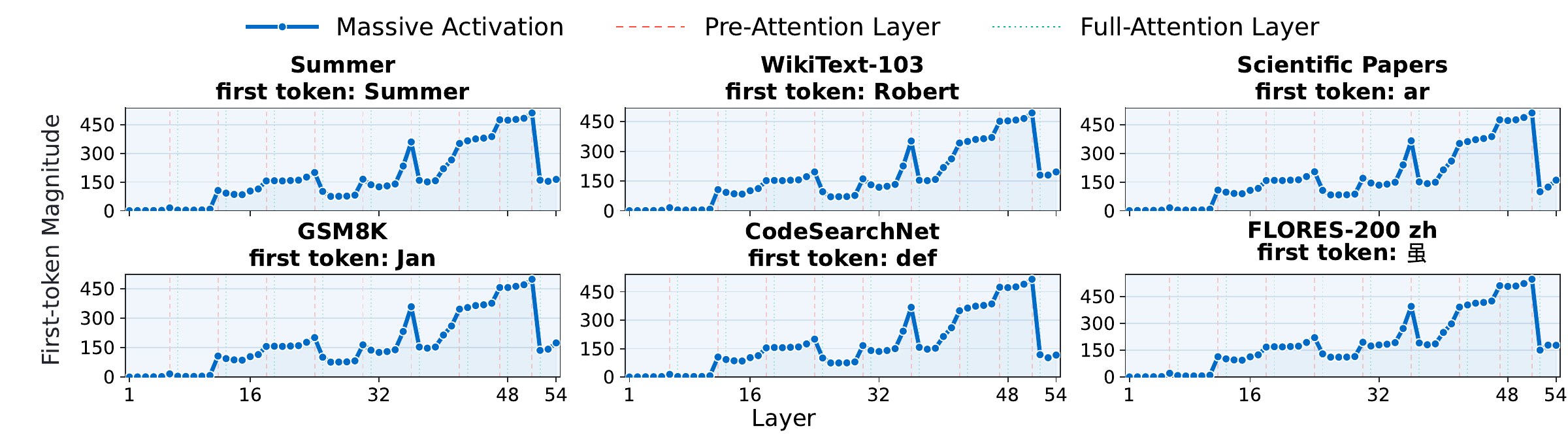}
    {Zamba2-2.7B}
    {fig:zamba2-2p7b}
    {0.21}

    \vspace{0.2em}

    \largemodelpanel
    {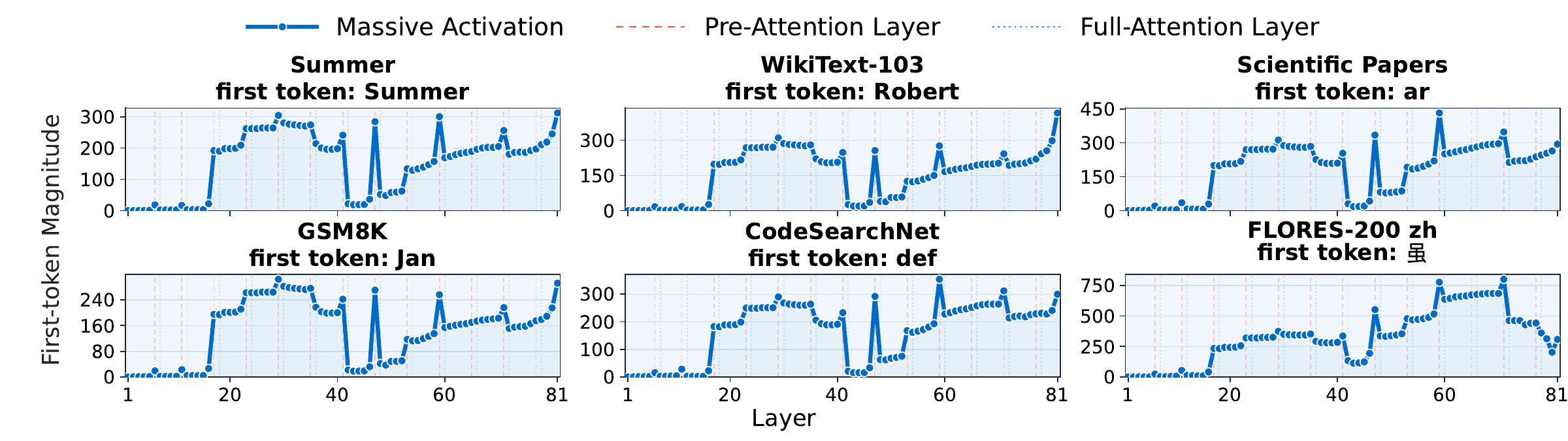}
    {Zamba2-7B}
    {fig:zamba2-7b}
    {0.21}

    \caption[]{
        \textbf{MA dynamics in large-scale pretrained hybrid models (continued).}
        Zamba2 checkpoints from 1.2B to 7B parameters further demonstrate that architecture-aligned spike and plateau morphologies recur across model scales within this state-space hybrid family.
    }
\end{figure*}

\clearpage

\clearpage

\newcommand{\maplifecyclefigure}[3]{%
\begin{figure*}[p]
    \centering

    \begin{subfigure}[t]{\textwidth}
        \centering
        \includegraphics[
            width=0.92\linewidth,
            height=0.17\textheight,
            keepaspectratio
        ]{
            figures/appendix/pas_lifecycle/appendix/#1/hybrid_24_1.pdf
        }
        \vspace{-4mm}
        \caption{#2, $24{:}1$.}
    \end{subfigure}

    \vspace{0.1em}

    \begin{subfigure}[t]{\textwidth}
        \centering
        \includegraphics[
            width=0.92\linewidth,
            height=0.17\textheight,
            keepaspectratio
        ]{
            figures/appendix/pas_lifecycle/appendix/#1/hybrid_12_1.pdf
        }
        \vspace{-4mm}
        \caption{#2, $12{:}1$.}
    \end{subfigure}

    \vspace{0.1em}

    \begin{subfigure}[t]{\textwidth}
        \centering
        \includegraphics[
            width=0.92\linewidth,
            height=0.17\textheight,
            keepaspectratio
        ]{
            figures/appendix/pas_lifecycle/appendix/#1/hybrid_6_1.pdf
        }
        \vspace{-4mm}
        \caption{#2, $6{:}1$.}
    \end{subfigure}

    \vspace{0.1em}

    \begin{subfigure}[t]{\textwidth}
        \centering
        \includegraphics[
            width=0.92\linewidth,
            height=0.17\textheight,
            keepaspectratio
        ]{
            figures/appendix/pas_lifecycle/appendix/#1/hybrid_3_1.pdf
        }
        \vspace{-4mm}
        \caption{#2, $3{:}1$.}
    \end{subfigure}

    \caption{
        \textbf{Layerwise systematic-outlier patterns across
        hybridization ratios in 1.3B #2 models from the M-A-P suite.}
        This figure extends the hybridization-ratio analysis in
        \S\ref{sec:ratio-dependence}; additional results and discussion
        are provided in Appendices~\ref{app:map-ma-dynamics}
        and~\ref{app:layerwise-lifecycle}.
    }
    \label{fig:lifecycle-#3-ratios}
\end{figure*}
\clearpage
}

\maplifecyclefigure
{GatedDeltaNet}
{GDN}
{gdn}

\maplifecyclefigure
{DeltaNet}
{DeltaNet}
{deltanet}

\maplifecyclefigure
{GLA}
{GLA}
{gla}

\maplifecyclefigure
{HGRN}
{HGRN}
{hgrn}

\maplifecyclefigure
{RetNet}
{RetNet}
{retnet}

\newcommand{\opensourcelifecyclepanel}[3]{%
\begin{subfigure}[t]{\textwidth}
    \centering
    \includegraphics[
        width=0.96\linewidth,
        height=0.22\textheight,
        keepaspectratio
    ]{#1}
    \vspace{-4mm}
    \caption{#2}
    \label{#3}
\end{subfigure}
}

\begin{figure*}[p]
    \centering

    \opensourcelifecyclepanel
    {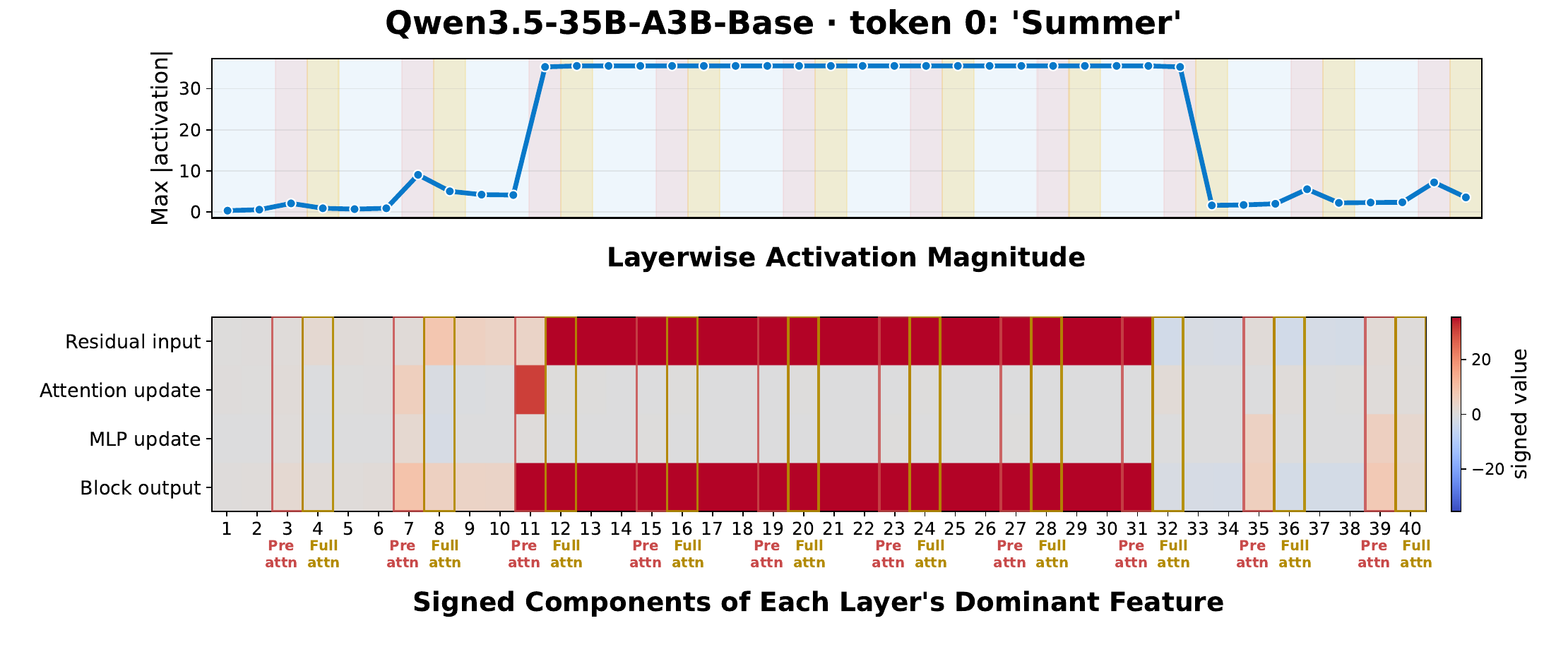}
    {Qwen3.5-35B-A3B-Base}
    {fig:lifecycle-qwen35-base}

    \vspace{0.2em}

    \opensourcelifecyclepanel
    {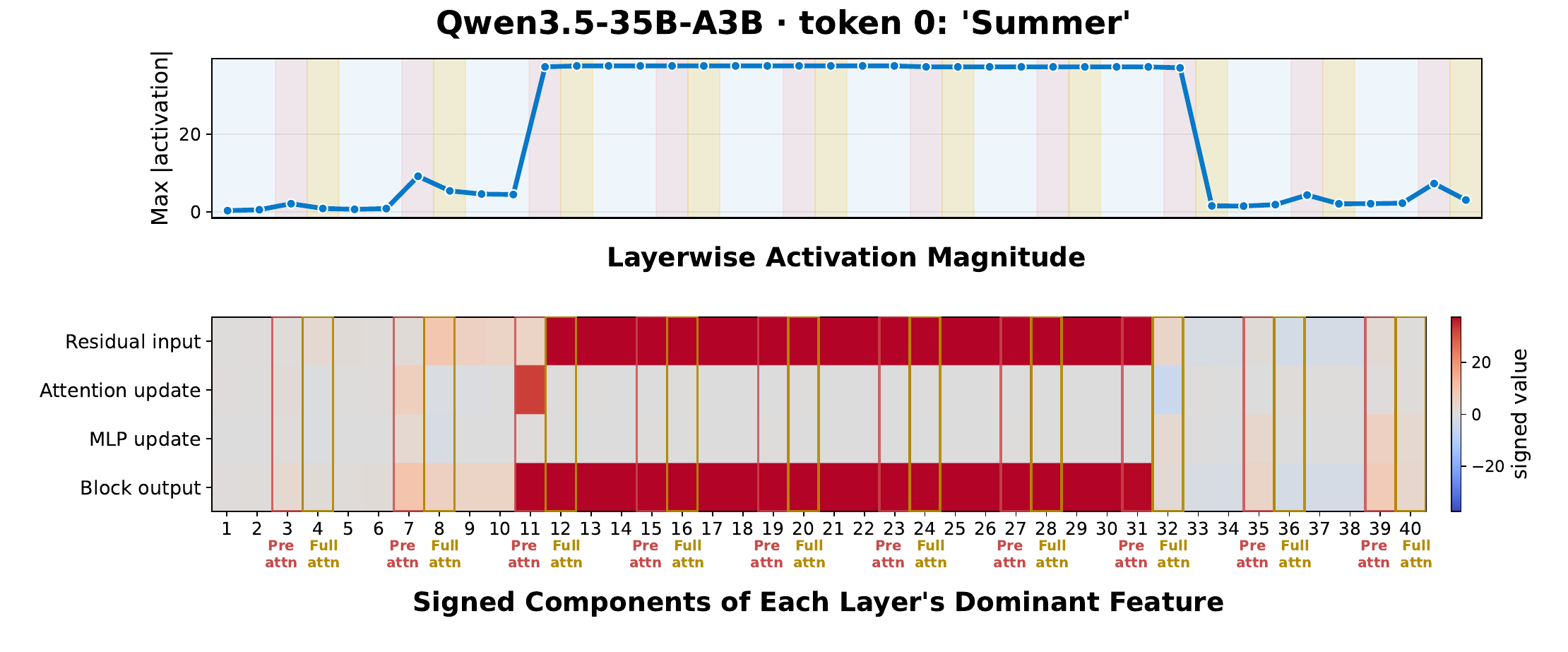}
    {Qwen3.5-35B-A3B}
    {fig:lifecycle-qwen35}

    \vspace{0.2em}

    \opensourcelifecyclepanel
    {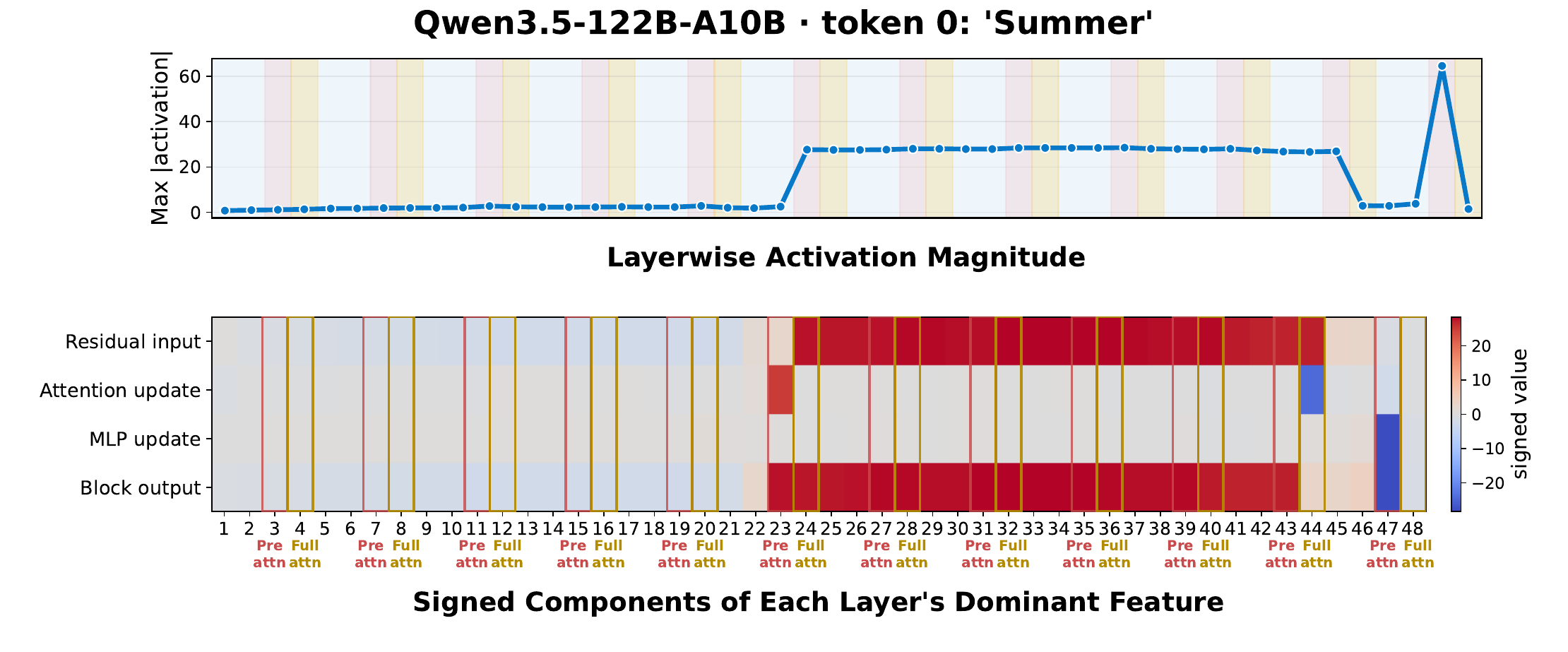}
    {Qwen3.5-122B-A10B}
    {fig:lifecycle-qwen122}

    \caption{
            \textbf{Layerwise systematic-outlier patterns in large-scale Qwen3.5 linear attention hybrid models.}
        Each panel traces the first-token MA trajectory for
        \textit{``Summer is warm. Winter is cold.''} together with its
        signed module-level decomposition.
        Across the evaluated scales and post-training stages, the dominant MA
        transitions remain systematically aligned with full attention
        placement.
        PAS and ISP persist despite the native output gating in Qwen3.5.
        This figure extends \S\ref{sec:large-scale-generalization};
        additional results are provided in
        Appendices~\ref{app:large-scale-ma-dynamics}
        and~\ref{app:layerwise-lifecycle}.
    }
    \label{fig:lifecycle-qwen-models}
\end{figure*}
\clearpage

\begin{figure*}[p]
    \centering

    \opensourcelifecyclepanel
    {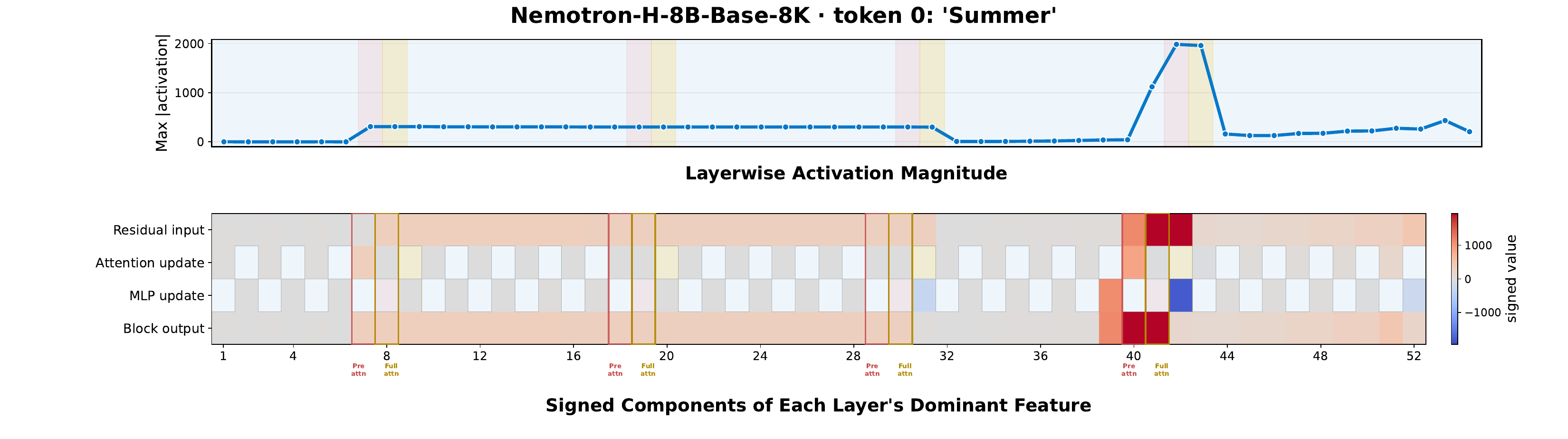}
    {Nemotron-H-8B-Base-8K}
    {fig:lifecycle-nemotron8}

    \vspace{0.2em}

    \opensourcelifecyclepanel
    {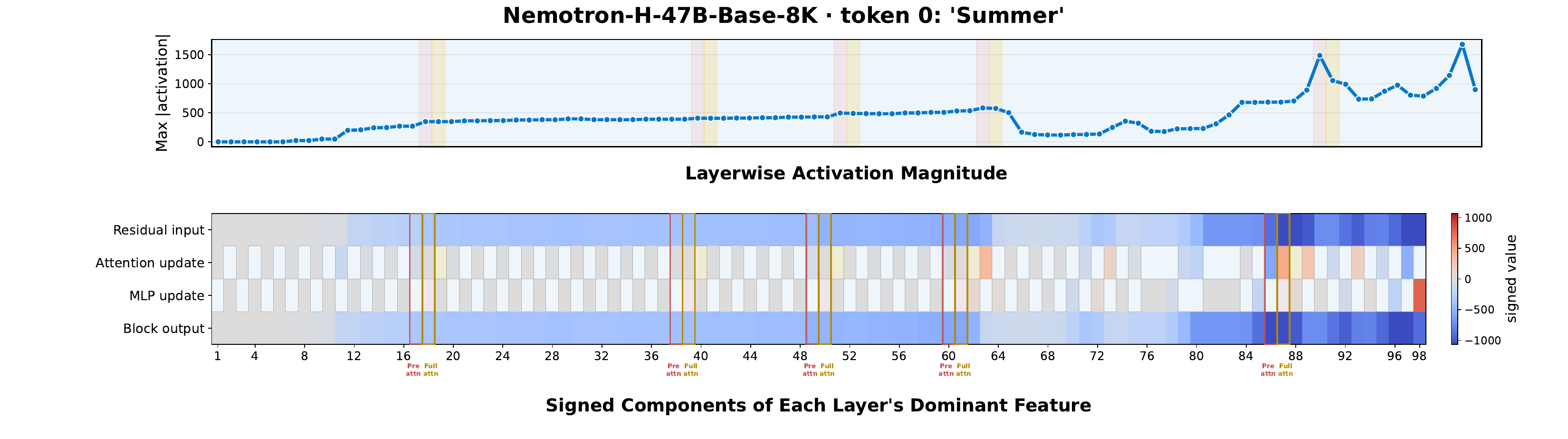}
    {Nemotron-H-47B-Base-8K}
    {fig:lifecycle-nemotron47}

    \vspace{0.2em}

    \opensourcelifecyclepanel
    {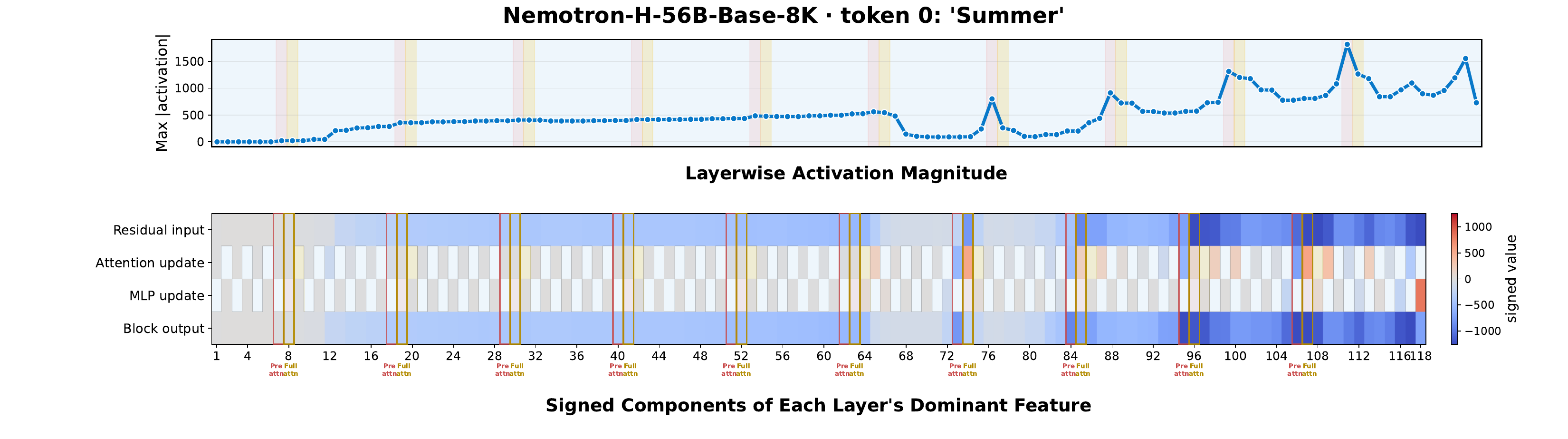}
    {Nemotron-H-56B-Base-8K}
    {fig:lifecycle-nemotron56}

    \caption{
        \textbf{Layerwise systematic-outlier patterns in large-scale Nemotron-H state-space hybrids.}
        Each panel traces the first-token MA trajectory for
        \textit{``Summer is warm. Winter is cold.''} together with its signed
        module-level decomposition.
        Across the evaluated checkpoints and layer schedules, dominant outlier
        amplification and opposing updates recur near full attention boundaries,
        extending the observed architecture-aligned pattern beyond linear
        attention backbones to hybrids interleaving Mamba-2 and full attention.
        This figure complements the analysis in
        \S\ref{sec:large-scale-generalization}; additional results across
        input domains are provided in
        Appendix~\ref{app:large-scale-ma-dynamics}.
    }
    \label{fig:lifecycle-nemotron-models}
\end{figure*}
\clearpage

\end{document}